\documentclass[11pt,twoside]{article}
\usepackage{graphicx}
\usepackage{multirow}
\usepackage{fancyhdr}
\usepackage{amsmath}
\usepackage{amssymb}
\usepackage{subfigure}
\usepackage{color}

\usepackage[textwidth=16cm,textheight=22cm,top=3cm,bottom=3cm]{geometry}

\usepackage{algorithm}
\usepackage[noend]{algpseudocode}
\makeatletter
\def\BState{\State\hskip-\ALG@thistlm}
\makeatother

\newtheorem{theorem}{Theorem}
\newtheorem{remark}{Remark}

\newcommand{\drop}[1]{}
\newcommand{\no}{\noindent}
\newcommand{\fer}[1]{(\ref{#1})}
\newcommand{\qtext}[1]{\quad\text{#1}}

\newcommand{\bx}{\mathbf{x}}
\newcommand{\by}{\mathbf{y}}

\newcommand{\bq}{\mathbf{q}}
\newcommand{\cK}{\mathcal{K}}

\newcommand{\vfi}{\varphi}

\newcommand{\R}{\mathbb{R}}

\def\O{\Omega}

\newcommand{\abs}[1]{| #1 |}
\newcommand{\nor}[1]{\| #1 \|}

\DeclareMathOperator{\NF}{NF}

\DeclareMathOperator{\cum}{cum}

\DeclareMathOperator{\F}{F}

\def\CC{{C\nolinebreak[4]\hspace{-.05em}\raisebox{.4ex}{\tiny\bf ++}}}

\begin{document}
% \title{ On a fast bilateral filtering formulation using functional rearrangements}
% 
% \author{Gonzalo Galiano         \and
%         Juli\'an Velasco 
% }
% 
% \institute{G.Galiano \at
%               Dpt. of Mathematics, Universidad de Oviedo,
%  c/ Calvo Sotelo, 33007-Oviedo, Spain \\
%               Tel.: +34-985103343\\
%               Fax: +34-985103354\\
%               \email{galiano@uniovi.es}           
%            \and
%            J. Velasco \at
%               Dpt. of Mathematics, Universidad de Oviedo,
%  c/ Calvo Sotelo, 33007-Oviedo, Spain

\title{On a fast bilateral filtering formulation using functional rearrangements
\thanks{Supported by Spanish MCI Project MTM2013-43671-P.}}

\author{Gonzalo Galiano  \thanks{Dpt. of Mathematics, Universidad de Oviedo,
 c/ Calvo Sotelo, 33007-Oviedo, Spain ({\tt galiano@uniovi.es, julian@uniovi.es})}
    \and Juli\'an Velasco\footnotemark[2] }
%         \thanks{VOn a fast bilateral filtering formulation using functional rearrangementsarious Affiliations, 
%         supported by various foundation grants.}}
\date{}

\pagestyle{fancy}
%\fancypagestyle{plain}
\fancyhead{}
\fancyhead[LE]{G. Galiano and J. Velasco} 
\fancyhead[RO]{On a fast bilateral filtering formulation using functional rearrangements}
%\pagestyle{myheadings}
%\lhead{G. Galiano and J. Velasco} 
% \chead{TEXTO}
%\rhead{Neighborhood filters and decreasing rearrangement}
%\markboth{G. Galiano and J. Velasco}{Neighborhood filters and decreasing rearrangement}
%\markleft{G. Galiano and J. Velasco}
%\markright{G. Galiano and J. Velasco}{Neighborhood filters and decreasing rearrangement}
\thispagestyle{plain}

\maketitle

\begin{abstract}
We introduce an exact reformulation of a broad class of neighborhood filters, among which the bilateral filters, in terms of two functional rearrangements: the decreasing and the relative rearrangements.

Independently of the \emph{image} spatial dimension (one-dimensional signal, image, volume of images, etc.), we reformulate 
these filters as integral operators defined in a one-dimensional space
corresponding to the level sets measures.

We prove the equivalence between the usual pixel-based version and the rearranged version of the
filter. When restricted to the discrete setting, our reformulation of bilateral filters extends previous results for the so-called
fast bilateral filtering. We, in addition, prove that the solution of the discrete setting, understood as constant-wise interpolators, 
converges to the solution of the continuous setting.

Finally, we numerically illustrate  computational aspects concerning quality approximation and execution time provided by the rearranged formulation. 

\no\emph{Keywords: }{Neighborhood filters, bilateral filter, decreasing rearrangement, relative rearrangement, denoising.}
%\subclass{68U10}
\end{abstract}

To appear in Journal of Mathematical Imaging and Vision, DOI  10.1007/s10851-015-0583-y

%%%%%%%%%%%%%%%%%%%%%%%%%%%%%%%%%%%%%%%%%%%%%%%%%%%%%%
\section{Introduction}

Let $\O\subset\R^d$ $(d\geq 1)$ be an open and bounded set,
$u\in L^\infty(\O)$ be an intensity image, and consider the family of filters to which we shall refer 
broadly as to \emph{bilateral filters},   defined by
\begin{align}
 \label{def.NL}
   \F u(\bx)= 
   \frac{1}{C(\bx)}\int_\O \cK_h(u(\bx)-u(\by)) w_\rho (\abs{\bx-\by})  u(\by)d\by,
 \end{align}
 where $h$ and $\rho$ are positive constants, and 
 \begin{equation*}
C(\bx)=\int_\O \cK_h(u(\bx)-u(\by)) w_\rho (\abs{\bx-\by}) d\by  
 \end{equation*}
is a normalization factor. 
 
Functions $\cK_h(\xi)=\cK(\xi/h)$ and $w_\rho$ are the \emph{range} kernel and the \emph{spatial} kernel of the filter, 
respectively, making reference to their type of interaction with the image domain. A usual choice for $\cK$ is the Gaussian $\cK(\xi)=\exp(-\xi^2)$, while different choices of $w_\rho$ give rise to several well known neighborhood filters, e.g.,  
\begin{itemize}
 \item The Neighborhood filter, see \cite{Buades2005}, for $w_\rho\equiv 1$.
 \item The Yaroslavsky filter \cite{Yaroslavsky1985,Yaroslavsky2003}, 
 for $w_\rho (\abs{\bx-\by})\equiv \chi_{B_\rho(\bx)}(\by)$, the characteristic function of a ball centered at $\bx$ of radios $\rho$.
 \item The SUSAN \cite{Smith1997} or Bilateral filters \cite{Tomasi1998}, 
 for $w_\rho (s)=\exp(-(s/\rho)^2)$.
 \end{itemize}
These filters have been introduced in the last decades
as  alternatives to local methods such as those expressed in terms 
of nonlinear diffusion partial differential equations (PDE), among which the pioneering approaches of Perona and Malik \cite{Perona1990}, \'Alvarez, Lions and Morel \cite{Alvarez1992} and Rudin, Osher and Fatemi \cite{Rudin1992} are fundamental. 
We refer the reader to \cite{Buades2010} for a review and comparison of these methods.

\subsection{Prior work}

Neighborhood filters have been mathematically analyzed
from different points of view. For instance, Barash \cite{Barash2002}, Elad \cite{Elad2002},  Barash et al. \cite{Barash2004}, and Buades et al. \cite{Buades2006}
investigate the asymptotic relationship between the Yaroslavsky filter and the Perona-Malik equation. Gilboa et al. \cite{Gilboa2008} study certain applications of 
nonlocal operators to image processing. In \cite{Peyre2008}, Peyr\'e establishes a relationship
between nonlocal filtering schemes and thresholding in adapted
orthogonal basis.
In a more recent paper, Singer et al. \cite{Singer2009}
interpret the Neighborhood filter as a stochastic diffusion process,
explaining in this way the attenuation of high frequencies in the processed images.

From the computational point of view, until the reformulation given by Porikli \cite{Porikli2008}, their actual implementation was of limited use due to the high computational demand of the direct space-range discretization. Only window-sliding optimization, like that introduced by Weiss \cite{Weiss2006} to avoid redundant kernel calculations, or filter
approximations, like the introduced by Paris and Durand \cite{Paris2006}, were of computational use. In \cite{Paris2006}, the space and range domains are merged into a single domain where the bilateral filter may be expressed as a linear convolution, followed by 
two simple nonlinearities. This allowed the authors to derive simple down-sampling criteria which were the key for
filtering acceleration.

However, in \cite{Porikli2008}, the author introduced a new \emph{exact} discrete formulation of the bilateral filter 
for spatial box kernel (Yarsolavsky filter) using the local histograms of the image, $h_\bx =h|_{B_\rho(\bx)}$, where $B_\rho(\bx)$ is the box of radios $\rho$ centered at pixel $\bx$, arriving to the formula
\begin{equation}
\label{porikli}
 \F u(\bx) = \frac{1}{C(\bx)}\sum_{i=1}^n q_i h_\bx (q_i) \cK_h(u(\bx)-q_i),
\end{equation}
where the range of summation is over the quantized values of the image, $q_1,\ldots,q_n$, instead of 
over the pixel spatial range. In addition, a zig-zag pixel scanning technique was used so that the local 
histogram is actualized only in the borders of the spatial kernel box.

Formula \fer{porikli} is an exact formulation of the box filter where all the terms but the local histogram may be computed separately in constant time, and it is therefore referred to as a \emph{constant time $O(1)$ method}.

Unfortunately, the use of local histograms is only valid for constant-wise spatial kernels, and subsequent 
applications of the new formulation to general spatial kernels is, with the exception of polynomial  and trigonometric polynomial kernels,
only approximated. Thus, in \cite{Porikli2008} polynomial approximation was used to deal with the usual spatial
Gaussian kernel. This idea was improved in \cite{Chaudury2011} by using trigonometric expansions.

In \cite{Yang2009}, Yang et al. introduced a new $O(1)$ method capable of handling arbitrary spatial and range kernels, as an extension of the ideas of  Durand et al \cite{Durand2002}. They use the so-called 
Principle Bilateral Filtered Image Component $J_k$, given by, for 
$u(\bx)=q_k$,
\begin{equation*}
 J_{q_k} (\bx)= \frac{\sum_{\by\in N(\bx)} \cK(q_k-u(\by)) w_\rho(\abs{\bx-\by}) u(\by)}{\sum_{\by\in N(\bx)} \cK(q_k-u(\by)) w_\rho(\abs{\bx-\by}) },
\end{equation*}
where $N(\bx)$ is some neighborhood of $\bx$. Then, the bilateral filter may be expressed as $ \F u(\bx) = J_{u(\bx)} (\bx)$.
In practice, only a subset of the range values is considered, and the final filtered image is produced by linear interpolation. In this situation this filter is, thus, an approximation to the bilateral filter.  The same authors have recently extended and optimized \cite{Yang2014} the method of Paris et al. \cite{Paris2006} by 
solving cost volume aggregation problems.

Other approaches include that of Kass et al. \cite{Kass2010}, who used a smoothed histogram to accelerate median filtering and proposed mode-based filters.
Adams et al. \cite{Adams2009} proposed to use Gaussian KD-trees for efficient
high-dimensional Gaussian filtering, integrating this method with Paris et al. method \cite{Paris2006} for fast bilateral filtering. They later proposed
to use permutohedral lattice \cite{Adams2010} for
bilateral filtering, which is faster than Gaussian KD-trees
for relatively lower dimensionality. However, both Gaussian KD-trees and permutohedral
lattice are relatively less efficient when applied to
intensity images. Ram et al. \cite{Ram2013} used a smooth patches reordering 
to image denoising and inpainting which, when the patches are shrunk to pixels, 
contain similar ideas to the decreasing rearrangement approach.
A review of some of these 
methods may be found in \cite{Milanfar2012}.

\subsection{Preceding applications of functional rearrangement}

In \cite{gv2013}, we heuristically introduced a denoising algorithm based in the Neighborhood filter for denoising images produced as time-frequency representations of one-dimensional signals, for which the large computational effort made useless other denoising methods based 
on PDE's. The main observation was that the Neighborhood filter can be computed using only the level sets of the image, instead of the pixel space, producing  a fast denoising algorithm. 

In \cite{Galiano2014}, this idea was rendered to a rigorous mathematical formulation through the notion of \emph{decreasing rearrangement} of a function. In short, the decreasing rearrangement of $u$, denoted by $u_*$,  is defined as the inverse of the \emph{distribution function} $m_u(q) = \abs{\{\bx \in\O : u(\bx) >q\}}$, see Section 2 for the precise definition and some of its properties.

Realizing that the structure of level sets of $u$ is invariant through the Neighborhood filter operation as well as through the decreasing rearrangement of 
$u$ allowed us to rewrite \fer{def.NL}, for the homogeneous kernel $w_\rho\equiv 1$, in terms of the one-dimensional integral expression
\begin{align}
\label{def.NFin}
 \NF_*  u(\bx)  = \frac{\int_{0}^{\abs{\O}}  \cK_h(u(\bx)-u_*(s)) u_*(s) ds}{\int_{0}^{\abs{\O}}  \cK_h(u(\bx)-u_*(s)) ds},
\end{align}
which is computed jointly for all the pixels in each level set $\{\bx:u(\bx)=q\}$.

Perhaps, the most important consequence of using the rearrangement was, apart from 
the large dimensional reduction, the reinterpretation of the Neighborhood filter as a 
\emph{local} algorithm. Indeed, notice that since $u_*$ is decreasing, we have 
\[
\abs{u_*(t)-u_*(s)}<h \qtext{if}\quad \abs{t-s}<c(h),
\] 
for some continuous function $c$. Thus, the range kernel of the pixel-based filter \fer{def.NL}, whose support in the pixel space may be disconnected, is transformed into a local kernel in its rearranged version \fer{def.NFin}.

Thanks to this we proved, among others, the following properties for the most usual nonlinear iterative variant of the Neighborhood filter:
\begin{itemize}
 \item The asymptotic behavior of $\NF_*$ as a shock filter of the type introduced by \'Alvarez et al. \cite{Alvarez1994}, combined with a contrast loss effect.
 
 \item The contrast change character of the Neighborhood filter, i.e. the existence of a continuous and
 increasing function $g:\R\to\R$ such that \[\NF u(\bx)\equiv \NF_* u(\bx) =g(u(\bx)).\]
\end{itemize}

\subsection{Objective of the article}

In this article we extend the use of functional rearranging techniques, as introduced in \cite{Galiano2014}, to bilateral filters which, in the discrete case and for the spatial box 
window, coincides with the 
bilateral filter reformulation given by Porikli \cite{Porikli2008}.

We provide a rigorous mathematical ground  which allows to interpret the discrete 
formulas of \cite{Porikli2008}, and their extension to general spatial and range kernels, as the discrete counterpart of continuous pixel-space formulations. 

The formulas we obtain for general bilateral filters allows to reinterpret these
filters from the range space point of view, as a natural extension to Porikli's 
discrete model. In addition, these formulas may be computationally competitive 
when a high degree of approximation to the pixel-based formulation is required.

This work may be considered as a non trivial extension of the rearranging techniques introduced in \cite{Galiano2014} for the simpler Neighborhood filter. Thus, even if the spatial kernel $w_\rho$ is non-homogeneous, we may still use our approach by introducing the \emph{relative rearrangement} of the spatial kernel with respect to the image, see Section 2 for definitions. 

Using this tool, we may express the general bilateral filter \fer{def.NL}
in terms of the one-dimensional integral expression
\begin{align}
\label{def.ubar0}
 F_* u(\bx)  = \frac{\int_{0}^{\abs{\O}}  \cK_h(u(\bx)-u_*(s))  w_\rho(\abs{\bx-\cdot})_{*u} (s) u_*(s) ds}{\int_{0}^{\abs{\O}}  \cK_h(u(\bx)-u_*(s)) w_\rho(\abs{\bx-\cdot})_{*u} (s)ds},
\end{align}
where $v_{*u}$ denotes the relative rearrangement of $v$ with respect to $u$.

To gain some insight into formula \fer{def.ubar0}, let us consider a constant-wise interpolation of a given image, $u$,
quantized in $n$ levels labeled by $q_i$, with $\max (u)=q_1 >\ldots > q_n=0$. That is $u(\bx)=\sum_{i=1}^n q_i\chi_{E_i} (\bx),$
where $E_i$ are the level sets of $u$, 
 \begin{equation*}
E_i=\{\bx \in \O : u(\bx)=q_i \},\quad i=1,\ldots,n.
\end{equation*}
Similarly, let $w_\rho$ be a constant-wise interpolation of the spatial kernel   
quantized in $m$ levels, $r_j$, with $\max (w_\rho)=r_1 >\ldots > r_m=\min(w_\rho)\geq0$.
For each $\bx\in \O$, consider  the partition of $E_i$ given by 
$F_j^i(\bx)=\{\by \in E_i : w_\rho(\abs{\bx-\by})=r_j \}.$
Then, we show in Theorems~\ref{th.equivalence} and \ref{th.approximation} that for each $\bx\in E_k$, $k=1,\ldots,n,$
\begin{align}
%\label{def.NLRD}
 \F u(\bx) \equiv \F_* u(\bx)=\frac{\sum_{i=1}^n \cK_h(q_k-q_i)  W_{im}(\bx) q_i}{\sum_{i=1}^n \cK_h(q_k-q_i)W_{im}(\bx)},
\end{align}
where $W_{im}(\bx)=\sum_{j= 1}^m  r_j \abs{F_j^i(\bx)}$, with $\abs{\cdot}$ denoting set measure (number of pixels). That is, $\abs{F_j^i(\bx)}$ is the number of pixels 
in the $q_i$-level set of $u$ that belongs to the $r_j$-level set of $w_\rho$.

Observe that, like for the Neighborhood filter,  the range kernel is transformed into a local kernel, independent of the pixel location, and it is therefore computed once and for all, explaining the large gain in computational cost, as already observed in \cite{Porikli2008}. 

In fact, notice that in the particular case in which $w_\rho$ is taken as a box window we have that  $W_{im}(\bx)$ is just the local histogram, $h_\bx$, used in Porikli's formula, so \fer{porikli} and \fer{def.ubar0} coincide in this case. 
However, the rearranged formula  \fer{def.ubar0} is valid for any other shape of the spatial kernel. 

The results in this article may be summarized as follows:

1. The rearranged formulation \fer{def.ubar0} is exact and  has no restrictions on the spatial or range kernels shape (Theorem~\ref{th.equivalence}).  In fact, our results are not limited to the special functional form
of the spatial kernel, $w_\rho \equiv w_\rho(\abs{\bx - \by})$, that we considered along this article. 
One can easily modify our arguments to deal with more general kernels of the type  $w_\rho \equiv w_\rho(\bx,\by)$, thus including other 
well known filters like the median filter  or the cross-joint 
bilateral filter \cite{Eisemann2004,Petschnigg2004}.
We preferred to limit ourselves to bilateral type filters for the sake of clarity, and leave the more general case to future developments of the work.

2. When restricted to a spatial box kernel, the relative rearrangement term of formula \fer{def.ubar0}  is
just the local histogram, like in Porikli's discrete formulation. For more general spatial kernels, the relative 
rearrangement may be interpreted as an averaged local histogram evaluated on the kernel level sets, extending in this way the notion
of local histogram and the discrete formulation of \cite{Porikli2008}.

3. We establish the relationship between the discrete and the continuous image setting of the rearranged version of the bilateral filter,
proving the convergence of constantwise interpolators to their continuous counterpart (Theorems~\ref{th.convergence},~\ref{th.approximation}).  

\subsection{Outline}

The plan of the article is the following. In Section~2 we recall the notions 
of decreasing and relative rearrangements and motivate the deduction of the rearranged formula 
\fer{def.ubar0} under some restrictive regularity assumptions. 

In Section~3, we remove those assumptions and establish the equivalence between 
the usual pixel-based expression of the filter \fer{def.NL} and its rearranged formulation \fer{def.ubar0}. In addition, we provide a fully discrete algorithm to approximate by constant-wise functions the filter $F_* u(\bx)$ given by \fer{def.ubar0}, and thus the original equivalent filter $Fu(\bx)$ given by \fer{def.NL}. We also prove the convergence of this discretization to the solution of the continuous setting. In Section~4, we give the proofs of these results.

In Section~5 we illustrate the performance of the rearranged filter formulation  in comparison to the 
\emph{brute force} pixel-based implementation and to other state of the art filters:
the $O(1)$ method of Yang et al. \cite{Yang2009}, and the permutohedral lattice method of Adams et al. \cite{Adams2009}. 

Finally, in Section~6 we give the Summary.

\section{Neighborhood filters in terms of functional rearrangements}

\subsection{The decreasing rearrangement}

Let us denote by $\abs{E}$ the Lebesgue measure of any measurable set $E$.
For a Lebesgue measurable function $u:\O\to\R$, the function 
$q\in\R\to m_u(q) = \abs{\{\bx \in\O : u(\bx) >q\}}$ is called the \emph{distribution function} corresponding to $u$. 

Function $m_u$ is non-increasing and therefore admits a unique  generalized inverse, called the
\emph{decreasing rearrangement}. This inverse takes the usual pointwise meaning when 
the function $u$ has no flat regions, i.e. when $\abs{\{\bx \in\O : u(\bx) =q\}} =0$ for any $q\in\R$. In general, 
the decreasing rearrangement $u_*:[0,\abs{\O}]\to\R$ is given by:
\begin{equation*}
u_*(s) =\left\{
\begin{array}{ll}
 {\rm ess}\sup \{u(\bx): \bx \in \O \} & \qtext{if }s=0,\\
 \inf \{q \in \R : m_u(q) \leq s \}& \qtext{if } s\in (0,\abs{\O}),\\
 {\rm ess}\inf \{u(\bx): \bx \in \O \} & \qtext{if }s=\abs{\O}.
\end{array}\right.
 \end{equation*}
We shall also use the notation $\O_*=(0,\abs{\O})$. Notice that since $u_*$ is non-increasing in $\bar\O_*$, it is continuous but at most a countable subset of  
$\bar\O_*$. In particular, it is then right-continuous for all $t\in (0,\abs{\O}]$.

The notion of rearrangement of a function is classical and was introduced by Hardy, Littlewood and Polya \cite{Hardy1964}. Applications include the study of isoperimetric and variational inequalities \cite{Polya1951,Bandle1980,Mossino1984}, comparison of solutions of partial differential equations \cite{Talenti1976,Alvino1978,Vazquez1982,Diaz1995,Alvino1996}, and others.
We refer the reader to the textbook \cite{Lieb2001} for the basic definitions.

Two of the most remarkable properties of the decreasing rearrangement are the  equi-measurability property 
\begin{equation*}
%\label{prop.1}
 \int_\O f(u(\by))d\by = \int_0^{\abs{\O}} f(u_* (s))ds,
\end{equation*}
for any Borel function $f:\R\to\R_+$, and the contractivity
\begin{equation}
 \label{prop.2}
 \nor{u_*-v_*}_{L^p(\O_*)}\leq \nor{u-v}_{L^p(\O)},
\end{equation}
for $u,v\in L^p(\O)$, $p\in[1,\infty]$.

\subsection{Motivation}

Apart from the pure  mathematical interest, 
the reformulation of  neighborhood filters in terms of functional rearrangements 
is  useful for computational purposes, specially when the spatial kernel, $w$, is homogeneous, that is $w_\rho\equiv 1$. In this case, it may be proven
\cite{Galiano2014} that the level sets of $u$ are invariant through the filter, i.e.  $u(\bx)=u(\by)$ implies  $F(u)(\bx)=F(u)(\by)$, and thus, it is sufficient 
to compute the filter only for each (quantized) level set, instead of for each  pixel, meaning a huge gain of computational effort.

For non-homogeneous kernels the advantages of the filter rearranged version are 
kernel-dependent, and in any case, the gain is never comparable to that of homogeneous kernels. The main reason is that the non-homogeneity of the spatial kernel breaks, in 
general, the invariance of level sets through the filter application.

In the following lines, we provide a heuristic derivation of the Bilateral filter rearranged version, 
as first noticed in \cite{Galiano2014}. We shall prove in Theorem~\ref{th.equivalence}  that the resulting formula is valid in a very general setting.

Under suitable regularity assumptions, the coarea formula states 
\begin{equation*}
%\label{coarea}
 \int_\O g(\by)\abs{\nabla u(\by)} d\by =\int_{-\infty}^{\infty} \int_{u=t} g(\by) d\Gamma(\by) dt.
\end{equation*}
Taking $ g(\by)=\cK_h(u(\bx)-u(\by)) w_\rho (\abs{\bx-\by}) u(\by)/\abs{\nabla u(\by)}$,
and using  $u(\bx) \in [0,Q]$ for all $\bx\in\O$ we get, for the numerator of the filter \fer{def.NL} ,
\begin{align*}
 I(\bx) & :=\int_\O \cK_h(u(\bx)-u(\by)) w_\rho (\abs{\bx-\by}) u(\by)d\by \\
  & =\int_{0}^{Q}  \cK_h(u(\bx)-t)  t 
  \int_{u=t}  \frac{w_\rho (\abs{\bx-\by})}{\abs{\nabla u(\by)}} d\Gamma(\by) dt.
\end{align*}
Assuming that $u_*$ is strictly decreasing and introducing the change of variable $t=u_{*}(s)$  we find
\begin{align}
 I(\bx) & =-\int_{0}^{\abs{\O}}  \cK_h(u(\bx)-u_*(s))u_*(s) \frac{d u_* (s)}{ds} \nonumber \\
  & \times \int_{u=u_*(s)}  \frac{w_\rho (\abs{\bx-\by})}{\abs{\nabla u(\by)}} d\Gamma(\by) ds \nonumber\\
 & =\int_{0}^{\abs{\O}}  \cK_h(u(\bx)-u_*(s)) u_*(s) w_\rho(\abs{\bx-\cdot})_{*u} (s)ds.
 \label{chco}
\end{align}
Here, the notation $v_{*u}$ stands for the \emph{relative rearrangement of $v$ with respect to $u$} which, 
under regularity conditions, may be expressed as 
\begin{equation}
\label{rrr}
 v_{*u}(s)  = \frac{\displaystyle\int_{u=u_*(s)}  \frac{v(\by)}{\abs{\nabla u(\by)}} d\Gamma(\by)}{ 
 \displaystyle\int_{u=u_*(s)}  \frac{1}{\abs{\nabla u(\by)}} d\Gamma(\by)},
\end{equation}
see the next subsection for details.
Transforming the denominator of the filter, $C(\bx)$, in a similar way allows us to deduce 
\begin{align}
\label{def.NLR}
 \F u(\bx)  = \frac{\int_{0}^{\abs{\O}}  \cK_h(u(\bx)-u_*(s))  w_\rho(\abs{\bx-\cdot})_{*u} (s) u_*(s)ds}{\int_{0}^{\abs{\O}}  \cK_h(u(\bx)-u_*(s)) w_\rho(\abs{\bx-\cdot})_{*u} (s)ds}.
\end{align}

\subsection{The relative rearrangement}

The relative rearrangement was introduced by Mossino and Temam \cite{Mossino1981} as the directional derivative of the decreasing rearrangement. Thus, if for $u\in L^1(\O)$ and $v\in L^p(\O)$, with $p\in [1,\infty]$, we consider the function
\begin{equation*}
 \vfi(s)=\int_{u>u_*(s)} v(\bx)d\bx + \int_0^{s-\abs{u>u_*(s)}} \big(v|_{u=u_*(s)}\big)_* (\sigma) d\sigma,
\end{equation*}
then the \emph{relative rearrangement of $v$ with respect to $u$}, $v_{*u}$, is defined as
\begin{equation*}
%\label{def.rr1}
 v_{*u}:=\frac{d}{ds} \vfi \in L^p(\O_*).
\end{equation*}
This identity may also be understood as the weak $L^p(\O_*)$ directional derivative 
(weak* $L^\infty(\O_*)$, if $p=\infty$),
\begin{equation}
\label{def.rr}
 v_{*u}= \lim_{t\to0}\frac{(u+tv)_* - u*}{t}. 
\end{equation}
Under the additional assumptions $u\in W^{1,1}(\O)$ and $\abs{\{\by\in\O :\nabla u(\by)=0 \}}=0$, i.e. the non-existence of flat regions of $u$, the identity \fer{rrr} is well defined and coincides with \fer{def.rr}. In this case, the relative rearrangement 
represents an averaging procedure of the values of $v$ on the level sets of $u$ labeled by 
the superlevel sets measures, $s$.

When formula \fer{rrr} does not apply, that is, in flat regions of $u$, we may resort to 
identity \fer{def.rr} to interpret the relative rearrangement  as the 
decreasing rearrangement of $v$ restricted to such flat regions of $u$.

After the seminal work of Mossino and Temam \cite{Mossino1981}, 
the relative rearrangement was further studied by Mossino and Rakotoson \cite{Mossino1986}
and applied to several types of problems, among which those related to variable exponent spaces and functional properties  \cite{Fiorenza2007,Fiorenza2009,Rakotoson2010,Rakotoson2012}, or nonlocal formulations of plasma physics problems \cite{Diaz1996,Diaz1998}.
In the rest of the article, we shall make an extensive use of the results given in the monograph on the relative rearrangement by Rakotoson \cite{Rakotoson2008}.

\subsection{Example: Rearrangement of constant-wise functions}

   \begin{figure*}[t]
\centering
\subfigure[Functions $u$ (blue) and $v$ (red)]
{\includegraphics[width=5cm,height=5cm]{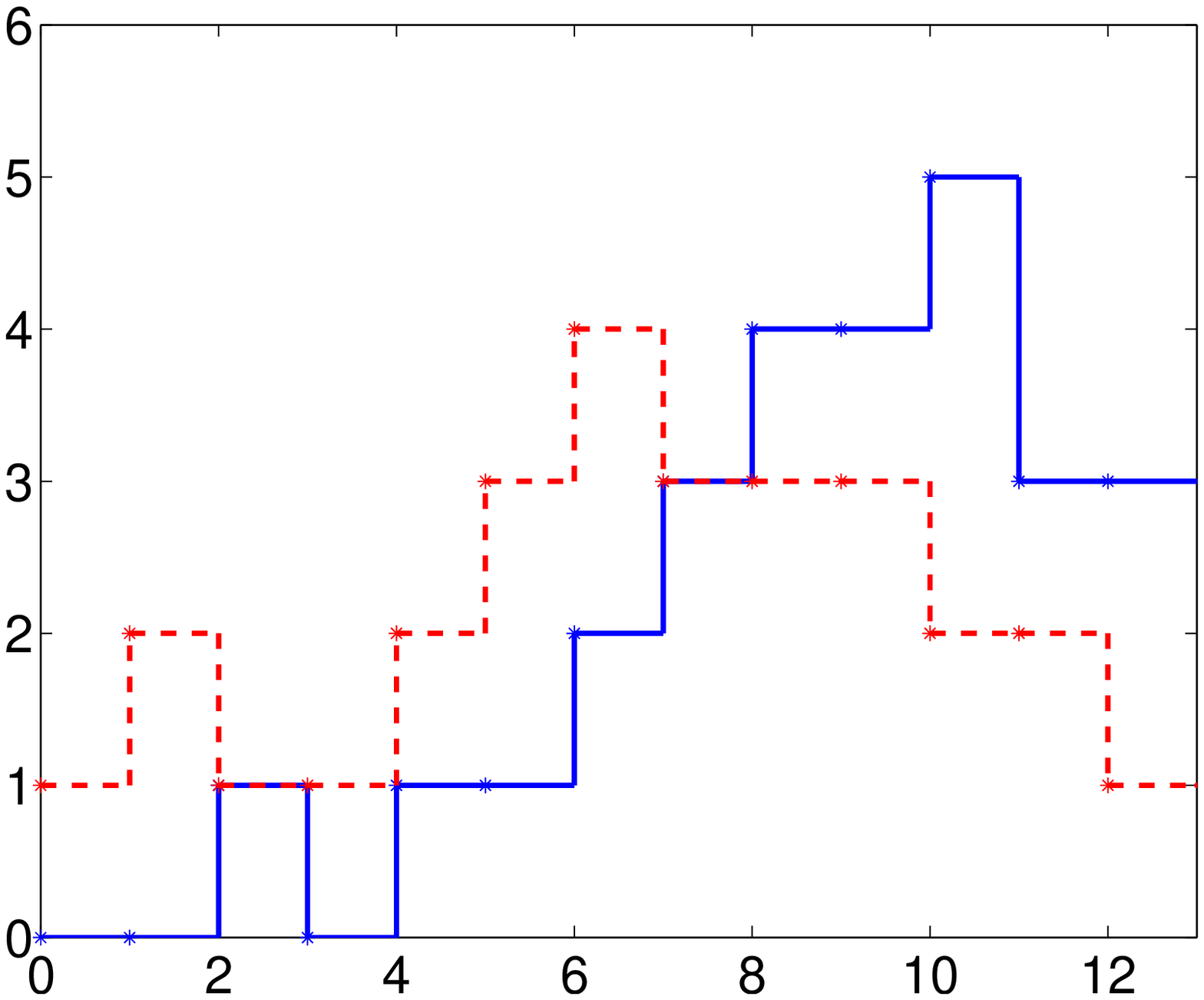}}
% \hspace{1cm}
\subfigure[Decreasing rearrangement of $u$, $u_*$]
 {\includegraphics[width=5cm,height=5cm]{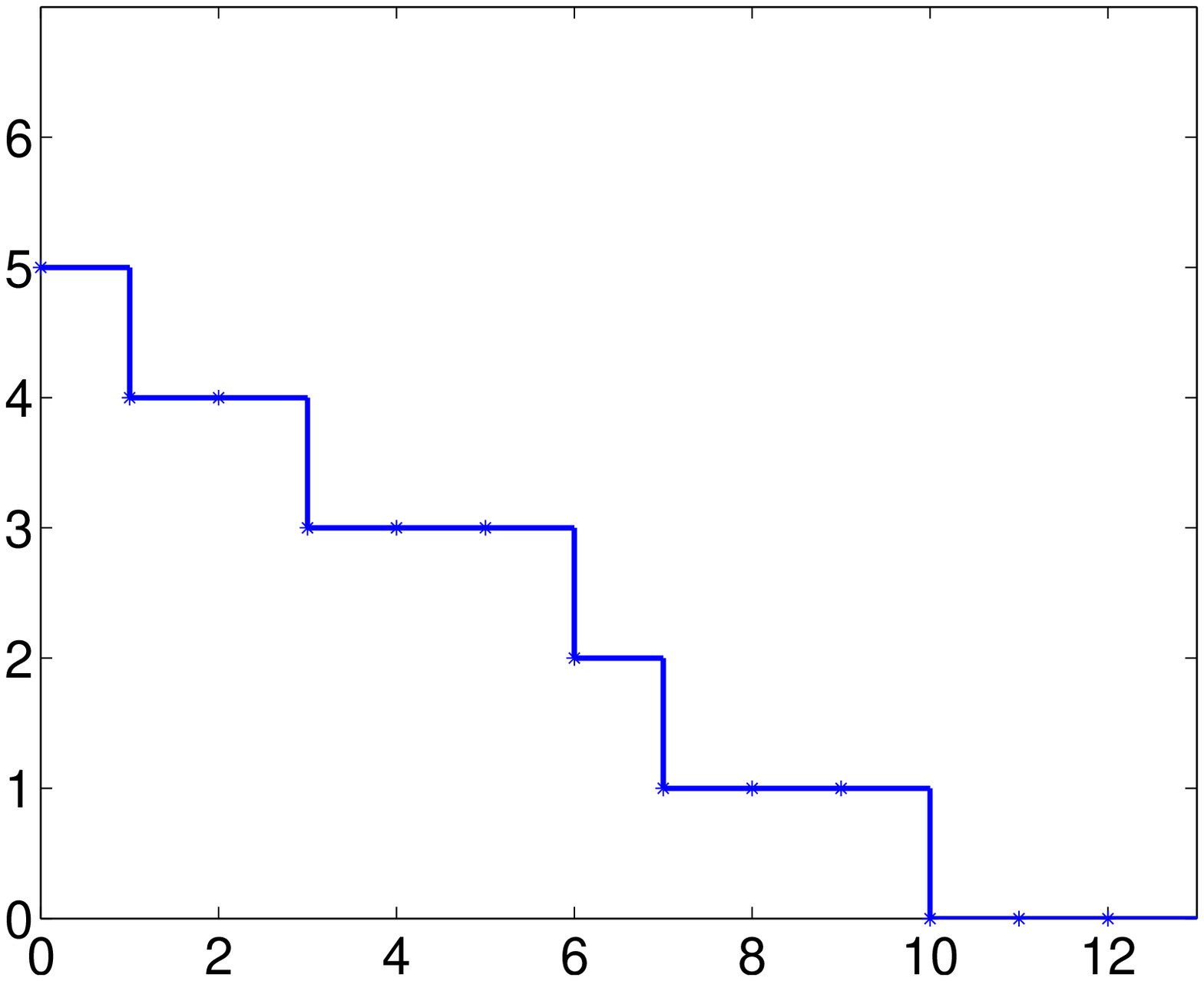}}\\
 \subfigure[$u_*$ (blue) and transported $v$ (red) ]
 {\includegraphics[width=5cm,height=5cm]{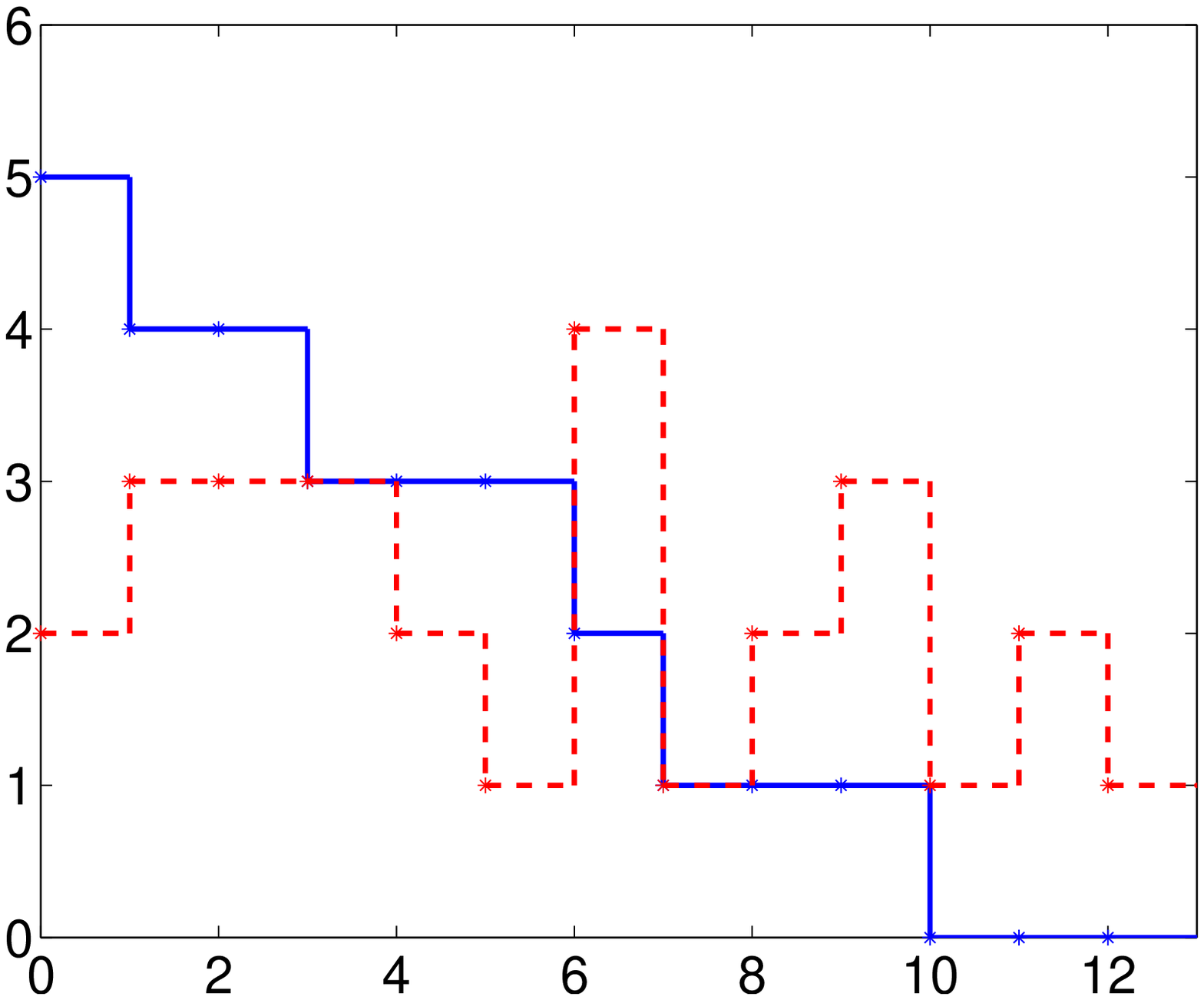}}
 \subfigure[The relative rearrangement $v_{*u}$]
 {\includegraphics[width=5cm,height=5cm]{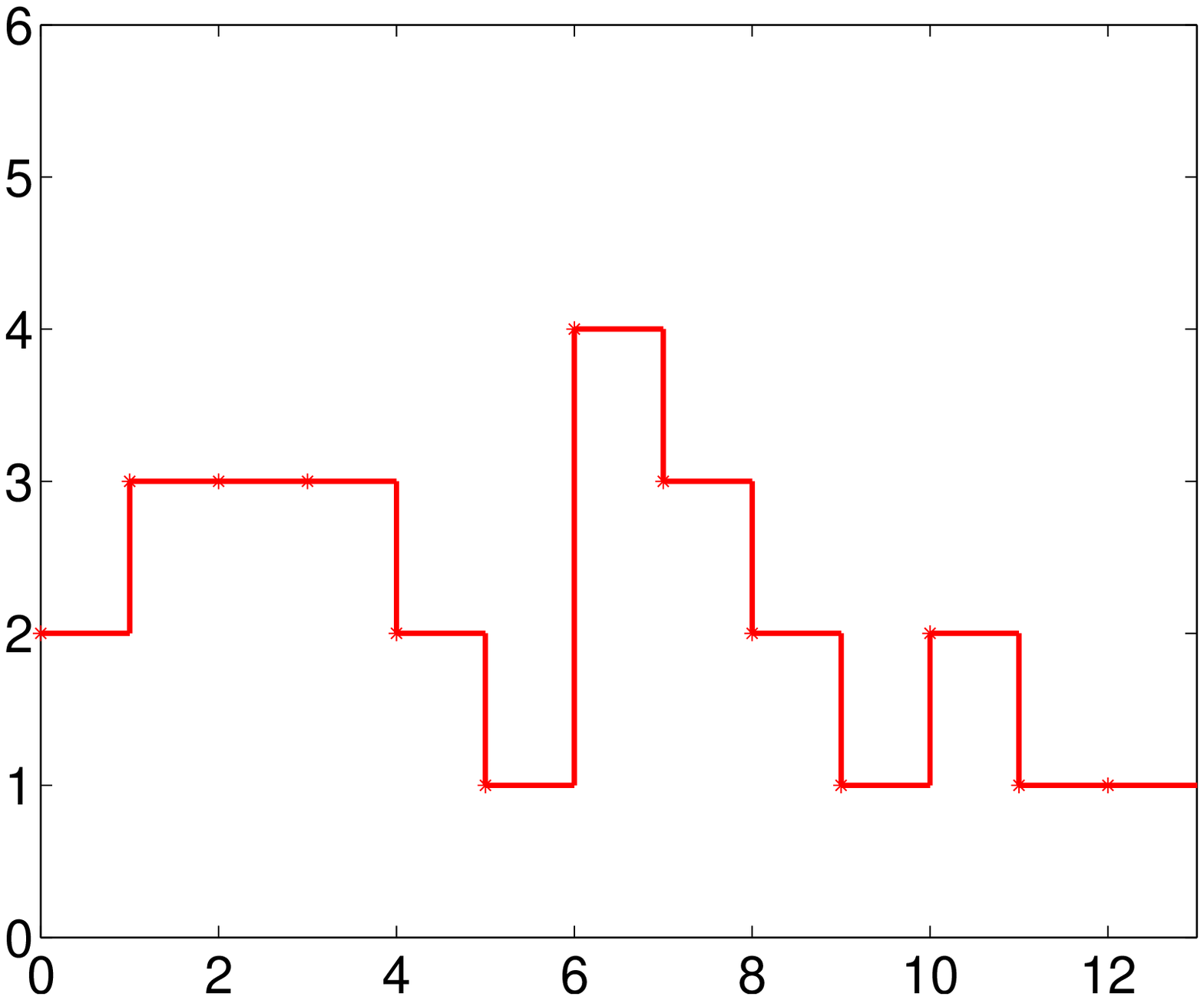}}
 \caption{{\small Example of construction of the relative rearrangement. (c) 
 shows $u_*$ and $v$ transported as by the displacement of the level sets of $u$ in the construction of $u_*$. (d) shows the decreasing rearrangement of $v$ restricted
 to the level sets of $u$, that is $v_{*u}$.
 }} 
\label{rem.fig}
\end{figure*}

Consider the constant-wise functions $u,v:[0,13]\to\R$ given in Fig.~\ref{rem.fig}~(a).
Writing $\max(u)=5=q_1 >\ldots > q_6=0=\min(u)$, we may express $u$ as $u(x)=\sum_{i=1}^6 q_i\chi_{E_i} (x)$, where $E_i$ are the level sets of $u$, $E_1=(10,11]$, $E_2=(8,10]$, etc.
Then, the decreasing rearrangement of $u$ is constant-wise too, and given by 
\begin{equation*}
 u_*(s)=\sum_{i=1}^n q_i\chi_{I_i}(s), 
\end{equation*}
with $I_i=[a_{i-1},a_i)$ for $i=1,\ldots,6$, and with $a_0=0$, $a_1=\abs{E_1}=1$, $a_2=\abs{E_1}+\abs{E_2}=3$,$\ldots$,$a_6=\sum_{i=1}^6 \abs{E_i}=\abs{\O}=13$.
The corresponding plot is shown in Fig.~\ref{rem.fig}~(b).

In Fig.~\ref{rem.fig}~(c) we show the graphs $\{E_i, v(E_i)\}_{i=1}^6$ transported as it was done 
in the step before to construct $u_*$. For instance, the highest level set of $u$, $E_1=(10,11]$, was transported to $[0,1]$; $E_2=(8,10]$ to  $(1,3]$, etc. Thus 
$\{E_1, v(E_1)\}=\{(10,11], \{2\} \}$ is transported to $\{[0,1], \{2\}\}$; 
$\{E_2, v(E_2)\}=\{(8,10], \{3\} \}$ is transported to $\{[1,3], \{3\}\}$, etc.

Finally, to obtain the decreasing rearrangement of $v$ with respect to $u$, $v_{*u}$, 
we rearrange decreasingly the restriction of $v$ to the level sets of $u$, $E_i$, as shown in  Fig.~\ref{rem.fig}~(d).

\section{Main results}

In this section we provide analytical results showing that, given an image,  
$u$, a spatial kernel, $w_\rho$, and a range kernel, $\cK_h$, all bounded in $L^\infty$,
we may approximate the filtered image $v=\F u$ by a sequence of constant-wise 
functions, $v_n$, which are obtained by filtering a constant-wise approximation 
of $u$, $u_n$, through a discrete filter, $\F_{m}$, involving a constant-wise approximation 
of $w_\rho$, $w_{_\rho m }$. 

In addition, and fundamentally, we show that the filter and its approximations may be obtained using the rearranged version \fer{def.NLR}, and give explicit formulas 
for their computation.

The main assumption we implicitly made for the heuristic deduction of formula \fer{def.NLR} is 
the condition $\abs{\{\by\in\O :\nabla u(\by)=0 \}}=0$, i.e. the non-existence of flat regions of $u$, 
which gives sense on one hand to formula \fer{rrr}, and on the other hand,  
allows us to obtain the strictly decreasing behavior of $u_*$, which justifies the change of variable in \fer{chco}.

Our first result is that formulas \fer{def.NL} and \fer{def.NLR} for the pixel-based filter and its rearranged version are equivalent under  weaker hypothesis. Due to the nature of our application, we keep the assumption on the boundedness of
$u$ and $w$ in $L^\infty$, although these can be also weakened. Theorems~\ref{th.equivalence} and \ref{th.convergence} are proven in Section~\ref{sec.proofs}. We use the notation $\R_+=\{t\in \R: t\geq 0\}$.

\begin{theorem}
\label{th.equivalence}
 Let $\O\subset\R^d$ be an open and bounded set, $d\geq 1$, $\cK\in L^\infty(\R,\R_+)$ 
 and $w_\rho\in L^\infty(\R_+,\R_+)$. Assume that $u\in L^\infty(\O)$ is, without loss of generality, non-negative. Consider $\F u(\bx)$ given by \fer{def.NL} and 
 \begin{align}
\label{def.ubar}
 \F_* u(\bx)  = \frac{\int_{0}^{\abs{\O}}  \cK_h(u(\bx)-u_*(s))  w_\rho(\abs{\bx-\cdot})_{*u} (s) u_*(s)ds}{\int_{0}^{\abs{\O}}  \cK_h(u(\bx)-u_*(s)) w_\rho(\abs{\bx-\cdot})_{*u} (s)ds}.
\end{align}
Then $\F_* u(\bx) = \F u(\bx)$ for a.e. $\bx \in \O$.
\end{theorem}

The next step towards the definition of a fully discrete algorithm to approximate the filter $\F_*$ given by \fer{def.ubar}, and thus the original equivalent filter $\F$ given by \fer{def.NL}, is showing that we may approximate $u$ and $w_\rho$ by constant-wise functions $u_n$ and $w_{_\rho m}$ such that,
as $m,n\to\infty$,
\begin{equation*}
%\label{th.conv3}
 \F_{*}^m u_n (\bx)\to \F_*u (\bx)
\end{equation*}
where $\F_{*}^m$ is the discrete version of $\F_*$, 
\begin{align}
\label{def.fhm}
 F_{*}^m u(\bx)  = \frac{\int_{0}^{\abs{\O}}  \cK_h(u(\bx)-u_*(s))  w_{_\rho m}(\abs{\bx-\cdot})_{*u} (s) u_*(s)ds}{\int_{0}^{\abs{\O}}  \cK_h(u(\bx)-u_*(s)) w_{_\rho m}(\abs{\bx-\cdot})_{*u} (s)ds}.
\end{align}

\begin{theorem}
\label{th.convergence}
Assume the conditions of Theorem~\ref{th.equivalence} and suppose that $u$ has a finite number of flat regions. Then, there exist sequences of constant-wise functions $u_n,~w_{_\rho m}$, with a finite number of flat regions, such that $u_n\to u$ strongly in $L^\infty(\O)$, $w_{_\rho m}\to w_\rho$ strongly in $L^\infty(\R_+)$,  and 
\begin{equation}
\label{th.conv3}
 F^m_{*} u_n \to F_*u \qtext{a.e. in }\O\qtext{and strongly in } L^\infty(\O).
\end{equation}
\end{theorem}
\begin{remark}
 Theorem~\ref{th.convergence} may be extended to the case in which $u$ has a countable 
 number of flat regions. However, in this case the approximation sequence of constant-wise functions $u_n$ has also a countable number of levels. Since our aim is providing a \emph{finite} discretization for numerical implementation,   such a sequence is not appropriate. 
\end{remark}

Theorem~\ref{th.convergence} gives utility to Theorem~\ref{th.approximation}, in which we produce the fully discrete formula actually used for the numerical implementation.

The main difficulty in computing formula \fer{def.fhm} is the determination of 
the relative rearrangement. However, in the case of constant-wise functions with 
a finite number of levels this computation is simplified thanks to identity \fer{def.rr}, 
which may be easily applied to this situation, as shown in \cite[Th.7.3.4]{Rakotoson2008}.

In few words, for the case of constant-wise functions $u$ and $v$, 
the relative rearrangement $v_{*u}$ may be computed as the decreasing rearrangement of $v$ 
restricted to the level sets of $u$.

%We drop the subindices $n$ and $m$ from these sequences for clarity.

\begin{theorem}
\label{th.approximation}
Let $u\in L^\infty(\O)$ be a constant-wise function quantized in $n$ levels labeled by $q_i$, with $\max (u)=q_1 >\ldots > q_n=0$. That is $u(\bx)=\sum_{i=1}^n q_i\chi_{E_i} (\bx),$
% \begin{equation*}
%  u(\bx)=\sum_{i=1}^n q_i\chi_{E_i} (\bx),
% \end{equation*}
where $E_i$ are the level sets of $u$, 
 \begin{equation*}
 %\label{def.levelsets}
E_i=\{\bx \in \O : u(\bx)=q_i \},\quad i=1,\ldots,n.
\end{equation*}
Similarly, let $w_\rho\in L^\infty(\R_+)$ be constant-wise and 
quantized in $m$ levels, $r_j$, with $\max (w_\rho)=r_1 >\ldots > r_m=\min(w_\rho)\geq0$.
For each $\bx\in \O$, consider  the partition of $E_i$ given by 
$F_j^i(\bx)=\{\by \in E_i : w_\rho(\abs{\bx-\by})=r_j \}.$
Then, for each $\bx\in E_k$, $k=1,\ldots,n$
\begin{align}
\label{def.NLRD}
 F_* u(\bx)=\frac{\sum_{i=1}^n \cK_h(q_k-q_i)  W_{im}(\bx)q_i}{\sum_{i=1}^n \cK_h(q_k-q_i)W_{im}(\bx)},
\end{align}
with $W_{im}(\bx)=\sum_{j= 1}^m  r_j \abs{F_j^i(\bx)}$.
\end{theorem}

\subsection{Examples}

As it clear from formula \fer{def.NLRD}, the main difficulty for its computation is the determination of the measures of the sets $F_{j}^i(\bx)$, which must be computed for each $\bx\in \O$. 

The formula provides the algorithm complexity. If $N$ is the number of pixels, then the complexity (without any further code optimization) is of the order $O(Nnm)$, where $n$ is the number of levels of the image and $m$ is the number of levels of the spatial kernel.
Let us examine some examples.

\no\emph{The Neighborhood filter. }In this case, $w_\rho\equiv 1$,   and therefore
$j=1$ and $F_j^i(\bx)=E_i$ is independent of $\bx$ for all $i=1,\ldots,n$. Thus, formula \fer{def.NLRD} is computed only on the level sets of $u$, that is, for all $\bx\in E_k$
\begin{align*}
%\label{def.NLRD}
 F^*_h u(\bx)=\frac{\sum_{i=1}^n  \cK_h(q_k-q_i)   \abs{E_i}q_i}{\sum_{i=1}^n  \cK_h(q_k-q_i)  \abs{E_i}}.
\end{align*}
In this case, the complexity is of order $O(n^2)$.

\no\emph{The Yaroslavsky filter. }
In this case, $w_\rho(\abs{\bx-\by})= \chi_{B_\rho(\bx)}(\by)$, and therefore
there are only two levels $r_1=1,~r_2=0$ of $w_\rho$ corresponding to the sets 
\begin{align*}
 %\label{def.levelsets}
& F_1^i(\bx)=\{\by \in E_i : \abs{\bx-\by)}<\rho \},\\ 
& F_2^i(\bx)=\{\by \in E_i : \abs{\bx-\by)}\geq\rho \},
\end{align*}
Thus, formula \fer{def.NLRD} reduces to: for each $\bx\in E_k$, $k=1,\ldots,n$
\begin{align*}
%\label{def.NLRD}
 F^*_h u(\bx)=\frac{\sum_{i=1}^n  \cK_h(q_k-q_i)   \abs{F_1^i(\bx)} q_i}{\sum_{i=1}^n  \cK_h(q_k-q_i)   \abs{F_1^i(\bx)}}
\end{align*}
The complexity is then of order $O(Nn)$.

\no\emph{The Bilateral filter. }In this case, $w_\rho(\abs{\bx-\by})=  \exp(-\abs{\bx-\by}^2/\rho^2)$ and therefore there is a continuous range of levels of $w_\rho$. However, for computational purposes the range of $w_\rho$ is quantized to some finite
number of levels, determined by the size of $\rho$. Thus, the full formula \fer{def.NLRD}
must be used in this case. The resulting complexity is of  order $O(Nnm)$.

The corresponding complexities for the pixel based formulation is, at least, of the order $O(N n\mu)$, where $\mu$ denote the discrete size of the  spatial kernel. However, meanwhile for the rearranged version the range kernel is computed only once, for the pixel-based version it must be computed for each pixel.

Observe that more optimized code and a further reduction of algorithm complexity may be reached by standard zig-zag 
techniques of window scanning, see \cite{Porikli2008}.

\section{Proofs}\label{sec.proofs}

\no\textbf{Proof of Theorem~\ref{th.equivalence}. } Let $f\in L^\infty(\R)$ and $b\in L^\infty(\O)$.
We start showing
\begin{equation}
\label{expr.0}
 \int_\O f(u(\by))b(\by)d\by = \int_0^{\abs{\O}} f(u_*(s))b_{*u}(s)ds.
\end{equation}
Consider the sets of flat regions of $u$ and $u_*$, 
\begin{equation}
 \label{def.P}
 P=\bigcup_{i\in D} P_i,\quad P_i=\{\by\in\O : u(\by)=q_i\},
\end{equation}
and $P_*=\cup_{i\in D} P^*_i$, with , $P^*_i=\{s\in\O_* : u_*(s)=q_i\}$, where the subindices set $D$ is, at most, countable.
According to \cite[Lemma 2.5.2]{Rakotoson2008}, we have 
\begin{align}
\label{expr.1}
 \int_0^{\abs{\O}} f(u_*(s))b_{*u}(s)= & \int_{\O\backslash P} f(u_* (m_u(u(\by))))b(\by)d \by \\
 & +\sum_{i\in D}\int_{P_i} M_{b_i}(h_i)(\by)b(\by)d\by,
\end{align}
where $b_i=b|_{P_i}$, $h_i(s)=f(u_*(s_i'+s))$ for $s\in [s_i',s_i''):=P_i^*$,  and
\begin{equation*}
M_{b_i}(h_i)(\by)=\left\{
\begin{array}{ll}
 h_i(m_{b_i}(b_i(\by))) & \text{if } \by\in P_i\backslash Q^i,\\
 \displaystyle\frac{1}{\abs{Q_j}}\int_{\sigma_j'}^{\sigma_j''} h_i(s)ds, & \text{if } \by\in Q^i_j,\\
\end{array}
\right.
\end{equation*}
where $Q^i=\cup_{j\in D'}Q_j^i$, with $Q_j^i$ the flat regions of $b_i$ and $ [\sigma_j',\sigma_j''):=Q_i^{j*}$. In \fer{expr.1},  since the functions $u_*$ and 
$m_u$ are strictly decreasing and inverse of each other in the set $\O\backslash P$, we obtain
\begin{align*}
%\label{expr.1}
 \int_{\O\backslash P} f(u_* (m_u(u(\by))))b(\by)d \by=
 \int_{\O\backslash P} f(u(\by))b(\by)d \by.
\end{align*}
In the flat regions of $u$ and $u_*$ we have, on one hand,
\begin{align*}
 \int_{P} f(u(\by))b(\by)d \by & = \sum_{i\in D}\int_{P_i} f(u(\by))b(\by)d \by\\
 & =
 \sum_{i\in D}f(q_i)\int_{P_i} b(\by)d \by.
\end{align*}
And, on the other hand, since $h_i(s)=f(u_*(s_i'+s))=f(q_i)$ for $s\in P_i^*$, 
\begin{align*}
 \sum_{i\in D} & \int_{P_i} M_{b_i}(h_i)(\by)b(\by)d\by \\
 = & \sum_{i\in D}  \left(\int_{P_i\backslash Q^i}   h_i(m_{b_i}(b_i(\by))) b(\by) d\by  \right. \\
&+ \left.\sum_{j\in D'} \frac{1}{\abs{Q_j}}\int_{\sigma_j'}^{\sigma_j''} h_i(s)ds \int_{ Q_j^i} b(\by)d\by\right) \\
= &  \sum_{i\in D}   \left(f(q_i) \int_{P\backslash Q^i}  b(\by) d\by+ f(q_i) \sum_{j\in D'} \int_{Q_j^i} b(\by)d\by\right) \\
= &   \sum_{i\in D}f(q_i)\int_{P_i} b(\by)d \by.
\end{align*}
Therefore, both sides of \fer{expr.0} are equal. 

Finally, for fixed $\bx\in \O$ set $b(\by) = w_\rho(\bx,\by)$ and first, 
$f(t)=\cK_h(u(\bx)-t)t$, for $t\geq0$  to obtain, using the identity \fer{expr.0}, the equality between the numerators of \fer{def.NL} and \fer{def.ubar}, and second,  $f(t)=K_h(u(\bx)-t)$ to obtain the equality between the denominators of those expressions.
$\Box$

\begin{remark}
 As deduced in the proof, identity \fer{expr.0} follows from  \cite[Lemma 2.5.2]{Rakotoson2008}. In fact, a little more than \fer{expr.0} may be obtained.
 Let $f,b\in L^\infty(\O)$. Then, if $f$ is constant in the flat regions of $u$, that is
 $f|_{P_i}=f_i=const$, then 
\begin{align}
\label{expr.10}
 \int_0^{\abs{\O}} f(s)b_{*u}(s)ds= & \int_\O f(m_u(u(\by)))b(\by)d\by \nonumber \\
 & +\sum_{i\in D}f(q_i)\int_{P_i} b(\by)d \by.
\end{align}
\end{remark}

\no\textbf{Proof of Theorem~\ref{th.convergence}. }We split the proof in several steps.

\no\emph{Step 1. }We use the construction of the sequence of constant-wise functions $u_n$ given in \cite[Th. 7.2.1]{Rakotoson2008}. In our case, the construction is simpler because $u$ has a finite number of flat regions, implying that $u_n$ has a finite number of levels.

In any case, this construction is such that $u_n\to u$ a.e. in $\O$ and strongly in $L^\infty(\O)$, and $w_{*u_n}\to w_{*u}$ weakly* in $L^\infty(\O_*)$. Besides, 
due to the strong continuity of the decreasing rearrangement \fer{prop.2}, we also have 
$(u_n)_*\to u_*$ strongly in $L^\infty(\O_*)$. Therefore, we readily see first that 
\begin{equation*}
 F_h^*u_n(\bx) \to F_h^* u(\bx) \qtext{for a.e. }\bx\in \O,
\end{equation*}
and then, due to the dominated convergence theorem,
\begin{equation}
\label{conv.1}
 F_h^*u_n \to F_h^* u \qtext{strongly in } L^\infty(\O).
\end{equation}

\no\emph{Step 2. }We consider a sequence of constant-wise functions with a finite number of levels, $w_m$, such that $w_m\to w$ strongly in $L^\infty(\O\times\O)$. Due to the contractivity property of the relative rearrangement, see \cite{Mossino1981}, we also have, for a.e. $\bx\in\O$, $w_m(\bx,\cdot)_{*v}\to w_\rho(\bx,\cdot)_{*v}$ strongly in $L^\infty(\O)$, for any $v\in L^\infty(\O)$. Thus, as $m\to \infty$,
\begin{align*}
%\label{def.ubar}
 F_{h,m}^* u_n(\bx)  =F_{h}^* u_n(\bx) \qtext{for a.e. }\bx\in \O,
\end{align*}
and, again, the dominated convergence theorem implies
\begin{equation}
\label{conv.2}
 F_{h,m}^*u_n \to F_h^* u_n \qtext{strongly in } L^\infty(\O).
\end{equation}

\no\emph{Step 3. }In view of \fer{conv.1} and \fer{conv.2}, we have
\begin{align*}
 \abs{F_{h,m}^*u_n -F_h^*u}\leq & \abs{F_{h,m}^*u_n -F_h^*u_n}\\
 & + \abs{F_{h}^*u_n -F_h^*u}\to0
\end{align*}
as $m\to\infty$ and $n\to\infty$, so \fer{th.conv3} follows.
$\Box$

\no\textbf{Proof of Theorem~\ref{th.approximation}. }Since $u$ is constant-wise, the decreasing rearrangement of $u$ is 
constant-wise too, and given by 
\begin{equation*}
 u_*(s)=\sum_{i=1}^n q_i\chi_{I_i}(s), 
\end{equation*}
with $I_i=[a_{i-1},a_i)$ for $i=1,\ldots,n$, and $a_0=0$, $a_1=\abs{E_1}$, $a_2=\abs{E_1}+\abs{E_2}$,$\ldots$,$a_n=\sum_{i=1}^n \abs{E_i}=\abs{\O}$. 
It is convenient to introduce here the cumulative sum of sets measures 
\begin{equation*}
 \cum(E_\circ,0)= 0,\qtext{and}\quad \cum(E_\circ,i)=\sum_{k=1}^i \abs{E_k},
\end{equation*}
for $i=1,...,n$, 
where the symbol $\circ$ denotes the summation variable. Thus, $a_i=\cum(E_\circ,i)$.

We consider, for fixed $\bx\in \O$ and $t>0$, the function 
$H(\by):=u(\by)+tw_\rho(\abs{\bx-\by})$. Since both $u$ and $w_\rho$  are constant-wise with a finite number of levels, we have, for $t$ small enough
\begin{equation*}
 q_{i+1}< q_i+tr_j< q_{i-1},
\end{equation*}
implying that each level set of $H$ is included in one and only one level set of $u$.
Thus, we have the orderings
\begin{itemize}
 \item For each $j$, if $i_1>i_2$   and $\by\in F_j^{i_1}(\bx),~ \bar \by \in F_j^{i_2}(\bx)$
 then  $H(\by)<H(\bar \by)$.
 
 \item For each $i$, if $j_1>j_2$   and $\by\in F_{j_1}^{i}(\bx),~ \bar \by \in F_{j_2}^{i}(\bx)$
 then  $H(\by)<H(\bar \by)$.
\end{itemize}
With these observations we may compute the decreasing rearrangement of $H$ as follows.
For instance, for $s\in I_1=[0,\abs{E_1})$ we have (omitting $\bx$ from the notation $F_i^j(\bx)$)
\begin{align*}
 H_*(s)=\left\{
 \begin{array}{ll}
  q_1+tr_1 & \text{if } s\in [0,\abs{F_1^1}),\\
  q_1+tr_2 & \text{if } s\in [\abs{F_1^1},\abs{F_1^1}+\abs{F_2^1}),\\
  \ldots & \ldots \\
  q_1+tr_m & \text{if } s\in [\cum(F^1_\circ,m-1),\cum(F^1_\circ,m)).
 \end{array}
\right.
\end{align*}
Notice that 
\begin{align*}
& [0,\abs{F_1^1})  =[\cum(F^1_\circ,0),\cum(F^1_\circ,1)),\\   
&    [\abs{F_1^1},\abs{F_1^1}+\abs{F_2^1})  = [\cum(F^1_\circ,1),\cum(F^1_\circ,2)),\\
& \ldots \\
&  [ \cum(F^1_{\circ},m-1), \cum(F^1_{\circ},m) )  = [ \cum(F^1_{\circ},m-1),\abs{E_1} ),
\end{align*}
so, in general, we may write for $i=1,\ldots,n$ and $j=1,\ldots,m$,
\begin{align*}
 H_*(s)= q_i+tr_j \qtext{if } s\in J_j^i(\bx),
\end{align*}
where $J_j^i(\bx):=\left[b_{j-1}^i(\bx),b_j^i(\bx)\right)$,
with $b_j^i(\bx):=a_{i-1}+\cum(F_\circ^i(\bx),j)$. 
Observe that $b_m^i(\bx)=a_i$. 
Finally, since $J_j^i(\bx)\subset E_i$ we have for $s\in J_j^i(\bx)$
\begin{align*}
 \frac{H_*(s)-u_*(s)}{t}=r_j,\qtext{implying}\quad w_\rho(\abs{\bx-\cdot})_{*u}(s) = r_j.
\end{align*}

We are now in disposition to compute formula \fer{def.ubar}. For $\bx \in E_k$,
\begin{align*}
 \int_{0}^{\abs{\O}} & \cK_h(u(\bx)-u_*(s)) u_*(s) w_\rho(\abs{\bx-\cdot})_{*u} (s)ds \\
 & = \sum_{i=1}^n \sum_{j= 1}^m \cK_h(q_k-q_i) q_i r_j \abs{J_j^i(\bx)} 
\end{align*}
In a similar way we obtain 
\begin{align*}
 C(\bx)= & \int_{0}^{\abs{\O}}  \cK_h(u(\bx)-u_*(s)) w_\rho(\abs{\bx-\cdot})_{*u} (s)ds \\
 = &  \sum_{i=1}^n \sum_{j= 1}^m \cK_h(q_k-q_i)  r_j \abs{J_j^i(\bx)}
\end{align*}
and therefore, using the definition of the sets $J_j^i(\bx)$ we deduce \fer{def.NLRD}.
$\Box$

\section{Experiments}
We conducted several  experiments on standard natural images to check the performance of the discrete formula \fer{def.NLRD} in comparison to well known state of the art denoising algorithms: Yang's et al. Real Time $O(1)$ Bilateral Filter\footnote{\CC~  code downloaded from \\ {\tt http://www.cs.cityu.edu.hk/$\sim$qiyang}} \cite{Yang2009}, 
and the Permutohedral Lattice Fast Filtering\footnote{\CC~ code downloaded from\\ {\tt http://graphics.stanford.edu/papers/permutohedral}} \cite{Adams2010}\label{pagina}. The aim of our experiments was to investigate the quality of the approximation of these algorithms to
\begin{enumerate}
 \item  the ground truth, and 
 
 \item the \emph{exact} pixel-based bilateral filter.
\end{enumerate}
We also compared execution times as delivered straightly from the available codes. Notice that execution time depends on code optimization and therefore a rigorous study of this aspect requires some kind of code normalization which was out of the scope of our study. 

In both experiments we used four intensity images of different sizes to compare execution times and peak signal to noise ratio (PSNR): \emph{Clock} ($256\times 256$), \emph{Boat} ($512\times 512$), \emph{Airport} ($1024\times 1024$), and \emph{Still life} ($2144\times 1424$). 
The first three images were taken from the data base of the 
Signal and Image Processing Institute,  University of Southern California, 
while the fourth was courtesy of A. Lubroth.

We applied the denoising filters to the test images corrupted with an additive Gaussian white noise of $\text{SNR}=10$, 
according to the noise measure $\text{SNR}=\sigma(u)/\sigma(\nu)$,
where $\sigma$ is the empirical standard deviation, $u$ is the original image, and $\nu$ is the noise. We repeated the experiments described in this section for higher levels of noise. Since the results were qualitatively similar, we omitted them for the sake of 
brevity.

We considered different spatial window sizes determined by $\rho$, with $\rho=4~,8,~16,~32$. For the Yaroslavsky filter, the spatial kernel is the characteristic function of a box of radius $\rho$, while for the bilateral filter a Gaussian function is taken in the experiments. 

The range size of the filter was taken, throughout all the experiments, as $h=\rho$ which, according to \cite{Buades2006}, is the regime in which the corresponding iterative filter behaves asymptotically as a Perona-Malik type filter. The shape of the range filter is always a Gaussian.

The discretization of the pixel-based and the rearranged version of both filters  was implemented in non-optimized \CC~ codes in which the main differences are those corresponding to formulas \fer{for.disg} and \fer{for.dis}. Execution time was measured 
 by means of function {\tt clock}. All the experiments were run on a standard laptop (Intel Core i7-2.80 GHz processor, 8GB RAM). 
 
\begin{algorithm}
\caption{Yaroslavsky filter (rearranged version)}\label{ps.yaros}
\begin{algorithmic}[1]
\State{ \textbf{* Algorithm according to formula \fer{for.dis}, with $q_k=k$*}}
\For {each pair $(i,k)$ of image quantization levels}
  \State 	$\cK_h(i,k)=exp(-((k-i)/h)^2)$
\EndFor
\State{ \textbf{(* Main loop *)} }
\For {each pixel, $\bx$, of the image}
  \State {$W_{i1} \gets 0$ }
  \If {$\bx$ is the first pixel}
  \For {each $\by \in B_\rho(\bx)$}
  \If {$u(\by)=i$}
    \State{ $W_{i1}(\bx) += 1$}
    \EndIf
  \EndFor
  \Else
    \For {each $\by \in B_\rho(\bx)$}
  \State {actualize $W_{i1}(\bx)$ adding/deleting new/old bins w.r.t. the previous box}
  \EndFor
  \EndIf
  \For {each quantization level, $i$ }
    \State {$numerator += \cK_h(i,u(\bx))*W_{i1}(\bx)*i$}
    \State {$denominator += \cK_h(i,u(\bx))*W_{i1}(\bx)$}
  \EndFor
  \State{ $uFiltered(\bx)= numerator/denominator$}
\EndFor
\end{algorithmic}
\end{algorithm}

\subsection{Yaroslavsky filter}
The performance of formula \fer{def.NLRD} for filter computation strongly depends on the functional form of the spatial kernel $w_\rho$. 
In the case in which $w_\rho$ is the characteristic function of a box of radius $\rho$, that is 
\begin{align}
 \label{def.NL2}
   \F u(\bx)= 
   \frac{1}{C(\bx)}\int_{B_\rho(\bx)} \cK_h (u(\bx)-u(\by)) u(\by)d\by,
 \end{align}
the number of level sets of the spatial kernel is just two, and the rearranged version requires only the computation of the measure of the sets $F_1^i(\bx)$ for each pixel $\bx$, that is, counting 
among all pixels in the $i-$level set of $u$ how many of them belong to the
spatial box.

 In the discretization of the pixel-based filter, 
 for each pixel $\bx$ we compute $u$ in the square neighborhood, $N(\bx)$, centered at $\bx$ of radius $2h$, containing  $(4h+1)^2$ pixels (close to the border, the image is extended by zero). Then,  we approximate \fer{def.NL2} as 
 \begin{align}
  \F u(\bx)\approx \frac{\sum_{\by\in N(\bx)} \cK_h (u(\bx)-u(\by)) u(\by)}{\sum_{\by\in N(\bx)} \cK_h (u(\bx)-u(\by))}.
  \label{for.disg}
 \end{align}
For the rearranged version we compute,  for each pixel $\bx\in E_k$, $k=1,\ldots,n$,
the number of pixels of the spatial box which belong to the different level sets of $u$,  
$ W_{i1}(\bx)=(W_{11}(\bx),\ldots,W_{n1}(\bx))$. That is, $W_{i1}(\bx)$ is the number of pixels of $E_i$ which are present in the box $N(\bx)$.
Then, we take 
 \begin{align}
  \F_* u(\bx)\approx \frac{ \sum_{i=1}^n  \cK_h (q_k-q_i)  W_{i1}(\bx) q_i}{ \sum_{i=1}^n \cK_h (q_k-q_i)W_{i1}(\bx)},
  \label{for.dis}
 \end{align}
where $\bq=(q_1,\ldots,q_n)$ are the quantization levels of $u$. Observe that since 
the rearranged version transforms the range kernel $\cK_h (u(\bx)-u(\by))$ 
into, $\cK_h (q_k-\bq)$, this term may be computed and stored outside the main loop running over all the pixels, see Algorithm~\ref{ps.yaros}. 

Thus, in \fer{for.dis}, only $M_r(\bx)$ and $q_k$ must be actualized for each  pixel, establishing an important difference with respect to the pixel-based version \fer{for.disg},
in which both the spatial and range kernels are actualized. In addition, following 
\cite{Porikli2008}, we perform the actualization of the measures $M_r(\bx)$ 
using previously computed information of neighbor boxes (zig-zag scanning technique).

\begin{algorithm}
\caption{Bilateral filter (rearranged version)}\label{ps.bilateral}
\begin{algorithmic}[1]
\State{ \textbf{* Algorithm according to formula \fer{for.bilr}, with $q_k=k$*}}
\For {each pair $(i,k)$ of image quantization levels}
  \State 	$\cK_h(i,k)=exp(-((k-i)/h)^2)$
\EndFor

\For {each $j$ of spatial kernel levels}
  \State { $r(j)=exp(-(j/\rho)^2)$}
\EndFor
\For {each pixel $\by$ in a generic Gaussian window}
  \If{ $exp(-(\by/\rho)^2)\approx r(j)$ }
        %\State 	$PixelLevel(\by)\gets$ its Gaussian level $j$
        \State 	$PixelLevel(\by)\gets$ $j$
  \EndIf
\EndFor
\State{ \textbf{(* Main loop *)} }
\For {each pixel, $\bx$, of the image}
  \State {$countBins \gets 0$ }
  \For {each pixel, $\by$, of Gauss. window centered at $\bx$}
    \State{ $countBins(u(\by),PixelLevel(\bx-\by)) += 1$}
  \EndFor
  \For {each quantization level, $i$ }
    \State {$W_{im}(\bx)\gets 0$}  
    \For {each spatial kernel level, $j$}
      \State {$W_{im}(\bx) += r(j)*countBins(i,j)$}
    \EndFor
    \State {$numerator += \cK_h(i,u(\bx))*W_{im}(\bx)*i$}
    \State {$denominator += \cK_h(i,u(\bx))*W_{im}(\bx)$}
  \EndFor
  \State{ $uFiltered(\bx)= numerator/denominator$}
\EndFor
\end{algorithmic}
\end{algorithm}
\subsection{Bilateral filter}

If the spatial kernel $w_\rho$ has a more complex profile than the characteristic function, e.g. the Gaussian function, then the rearranged version requires the computation of, for each pixel $\bx$,  the measure of the sets $F_j^i(\bx)$ for all the level setss, $1\leq j \leq m$, of some discretization of $w_\rho$. 
For instance, for $\rho=4~,8,~16,~32$, the corresponding number of the Gaussian function quantization levels arising from double precision arithmetic are $m=m(\rho)$,  with $m(4)=42$, $m(8)=135$, $m(16)=457$, and $m(32)=1621$. 

The approximations to the pixel based and the rearranged version are, respectively, 
 \begin{align}
  \F u(\bx)\approx \frac{\sum_{\by\in N(\bx)} \cK_h (u(\bx)-u(\by)) w_\rho(\abs{\bx-\by})u(\by)}{\sum_{\by\in N(\bx)} \cK_h (u(\bx)-u(\by)) w_\rho(\abs{\bx-\by})},
  \label{for.bil}
 \end{align}
and 
\begin{align}
\label{for.bilr}
 F_* u(\bx)\approx \frac{\sum_{i=1}^n \cK_h(q_k-q_i)  W_{im}(\bx)q_i}{\sum_{i=1}^n \cK_h(q_k-q_i)W_{im}(\bx)},
\end{align}
with $W_{im}(\bx)=\sum_{j= 1}^m  r_j \abs{F_j^i(\bx)}$,
where, for $\bx\in E_k$, $\abs{F_j^i(\bx)}$ is the number of pixels of $E_i$ which belong to the $j-$ level set of $w_\rho$.

The bilateral filter may be accelerated by manipulating the quantization levels of  the image, and of the spatial range. As shown in \cite{Yang2009} for the Yaroslavsky filter, the reduction of the image quantization levels leads to poor denoising results. 
However, we checked that a similar restriction applied to the spatial kernel reduces the execution time while conserving good denoising quality. In our experiments we tested this ansatz 
 with a fixed number of levels, $m=20$, for all the images and $\rho-$values.

\subsection{Other algorithms and discretization details}
The algorithm introduced by Yang et al. \cite{Yang2009} is the following. Let 
\[ 
\{\tilde q_1,\ldots,\tilde q_{\tilde n}\} \subset\{q_1,\ldots,q_n\}, \quad \tilde n \leq n,
\]
be a subset of the quantization levels of the image. Then, 
if $u(\bx)\in [\tilde q_k , \tilde q_{k+1}]$, Yang's filter is given by the image interpolation 
\begin{equation}
 \label{yang2}
 F_Y u(\bx) =(\tilde q_{k+1}- u(\bx))J_{\tilde q_k}(\bx) +
 ( u(\bx)-\tilde q_{k})J_{\tilde q_{k+1}}(\bx),
\end{equation}
where
\begin{equation}
 \label{yang1}
 J_{\tilde q_k}(\bx) = \frac{\sum_{\by\in N(\bx)} \cK_h (\tilde q_k -\by) w_\rho(\abs{\bx-\by}) u(\by)   }{\sum_{\by\in N(\bx)} \cK_h (\tilde q_k -\by) w_\rho(\abs{\bx-\by})}.
\end{equation}
Thus, if $\tilde n=n$ then $F_Y$ coincides with the exact discrete version of the bilateral filter, whereas if $\tilde n <n$ an approximation of the bilateral filter is obtained by  
interpolation on the quantization range kernel levels. Observe that since $\bx$ is no longer present in the range kernel of $J_{\tilde q_k}(\bx)$, both factors in \fer{yang1}
may be computed by fast approximated convolution algorithms, which is the main strength of the algorithm provided in \cite{Yang2009}

However, in the first experiment (Yaroslavsky filter), and in order to obtain \emph{exact} results when $\tilde n=n$, we implemented Yang's algorithm by using our rearranged version \fer{for.dis} applied to $J_{\tilde q_k}(\bx)$, explaining the increment of execution times when compared to the actual implementation of Yang et al.

As already mentioned, we also use for comparison purposes the fast filtering using the permutohedral lattice introduced by Adams et al. This algorithm is based on resampling  techniques and specially useful for high dimensional filtering. We refer to \cite{Adams2010} for a thoroughfull explanation of the method.

We use the following notation for the algorithms to be compared:
\begin{itemize}
 \item YPB: Yaroslavsky pixel-based version, formula \fer{for.disg}.
 
 \item YRR: rearranged Yaroslavsky version, formula \fer{for.dis}.
 
 \item EYang$\tilde n$: \emph{exact} Yang's algorithm, formula \fer{yang2}, computed with the rearranged strategy, for $\tilde n=256$, $64$ or $8$ interpolators.
 
 \item BPB: Bilateral pixel-based version, formula \fer{for.bil}.
 
 \item BRRM: rearranged bilateral version, formula \fer{for.bilr}, with the maximum number 
 of spatial kernel discretization levels, $m(4)=42$, $m(8)=135$,...
 
 \item BRR20: rearranged bilateral version, formula \fer{for.bilr}, with fixed number 
 of spatial kernel discretization levels, $m=20$.
 
 \item Y$\tilde n$: Yang's code provided by the authors, see footnote 1 in page \pageref{pagina}, for $\tilde n=256$ or $\tilde n = 8$ interpolators. 
 
 \item P: Permutohedral Lattice bilateral filter code provided  by the authors, see footnote 2 in page \pageref{pagina}.

\end{itemize}

\subsection{Results}

In the first experiment, we tested the approximation quality  of the rearranged 
and the exact Yang's versions of the Yaroslavsky filter to the pixel-based version. 
In Table~\ref{table1} we show the PSNR of YRR and EYang with respect to YPB. We recall 
that, according to \cite{Paris2006}, PSNR values above 40dB often correspond to almost invisible differences. The high PSNR values obtained for YRR and EYang256 show that these 
algorithms give practically the same results than YPB, up to rounding errors. EYang64
still gives very good approximation results, whereas these significantly diminishes for 
EYang8. 

In Table~\ref{table1} we also show the execution times. YRR is up to 400 times faster 
than YPB. The execution times of EYang compared to YRR are always higher, but this has to be taken cautiously since the code implementations were not optimized. In this experiment
we were more interested in approximation quality.

In the second experiment, we tested the approximation quality of the algorithms described in the previous subsection to the ground truth image and to the bilateral pixel-based filter (BPB).

In Table~\ref{table2} we show the corresponding PSNR's. We see that all the algorithms 
employed give similar results when compared to the ground truth image. Thus, if this were
the choice criterium, the faster, that is $Y8$, should be considered. 

However, when compared to the BPB, the PSNR's are quite different. As expected, the rearranged version
with the maximum number of quantization levels for the spatial kernel, BRRM, gives the same result, up to rounding errors, than the BPB. However, the Yang's algorithm with the maximum number of levels, Y256, although \emph{should} give exact results, it does not, revealing other sources of error beyond rounding errors. Even 
the BRR20 and the Yaroslavsky YRR give better approximations than Y256. In particular, BRR20 has always values of PSNR around 40dB. The Permutohedral algorithm gives similar results than Y256 and Y8. 

Let us mention two difficulties we found with the Yang 
code implementation. The first is related to low $h-$values, for which Y8 gives poor results due to some artifact formation. The second is related to large images, for which huge memory allocation requirements result in running time execution errors. 

In Table~\ref{table3} we collect the execution times obtained in this experiment. Only 
for the smaller image sizes and $h-$values Y8 has a competitor in YRR. For large images 
and high $h-$values the second fastest algorithm is, as expected, the permutohedral, although always taking around twice the time of Y8. Both BRRM and BRR20 give 
execution times considerably higher than the other algorithms, for our non-optimized 
codes. 

\begin{figure*}[ht] 
\centering 
{\includegraphics[width=3cm,height=3cm]{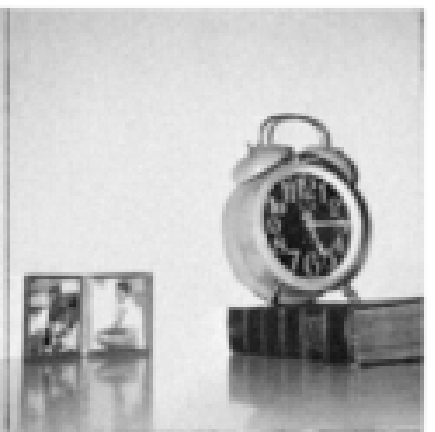}}
{\includegraphics[width=3cm,height=3cm]{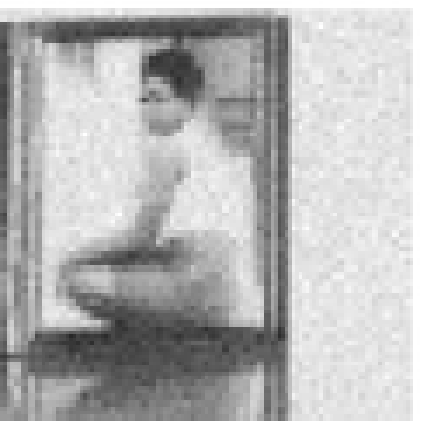}}
{\includegraphics[width=3cm,height=3cm]{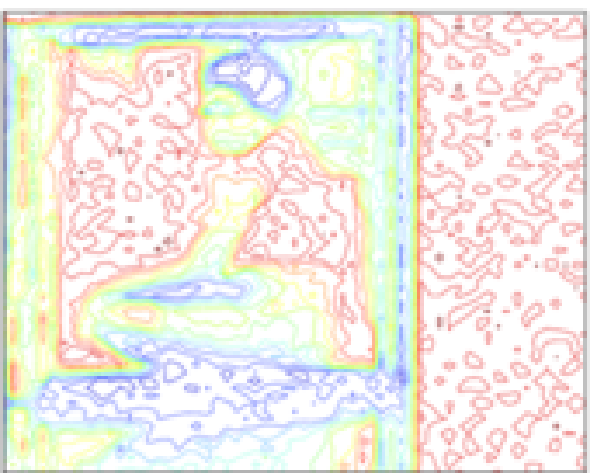}}
{\makebox[3cm]{}}\\
{\includegraphics[width=3cm,height=3cm]{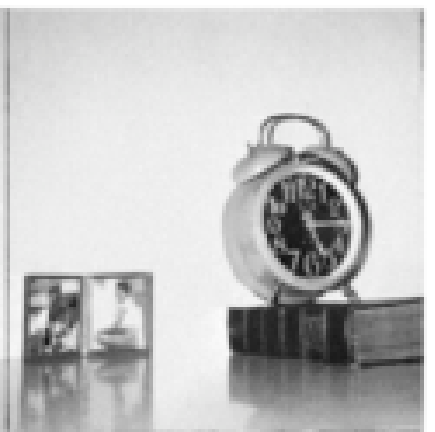}}
{\includegraphics[width=3cm,height=3cm]{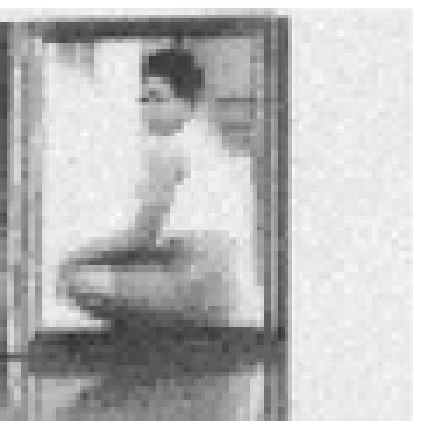}}
{\includegraphics[width=3cm,height=3cm]{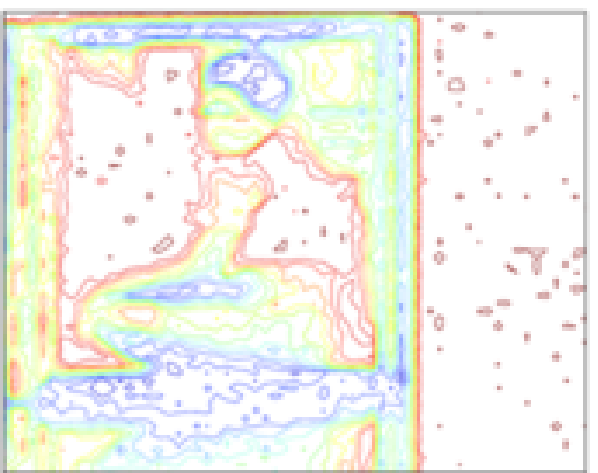}}
{\includegraphics[width=3cm,height=3cm]{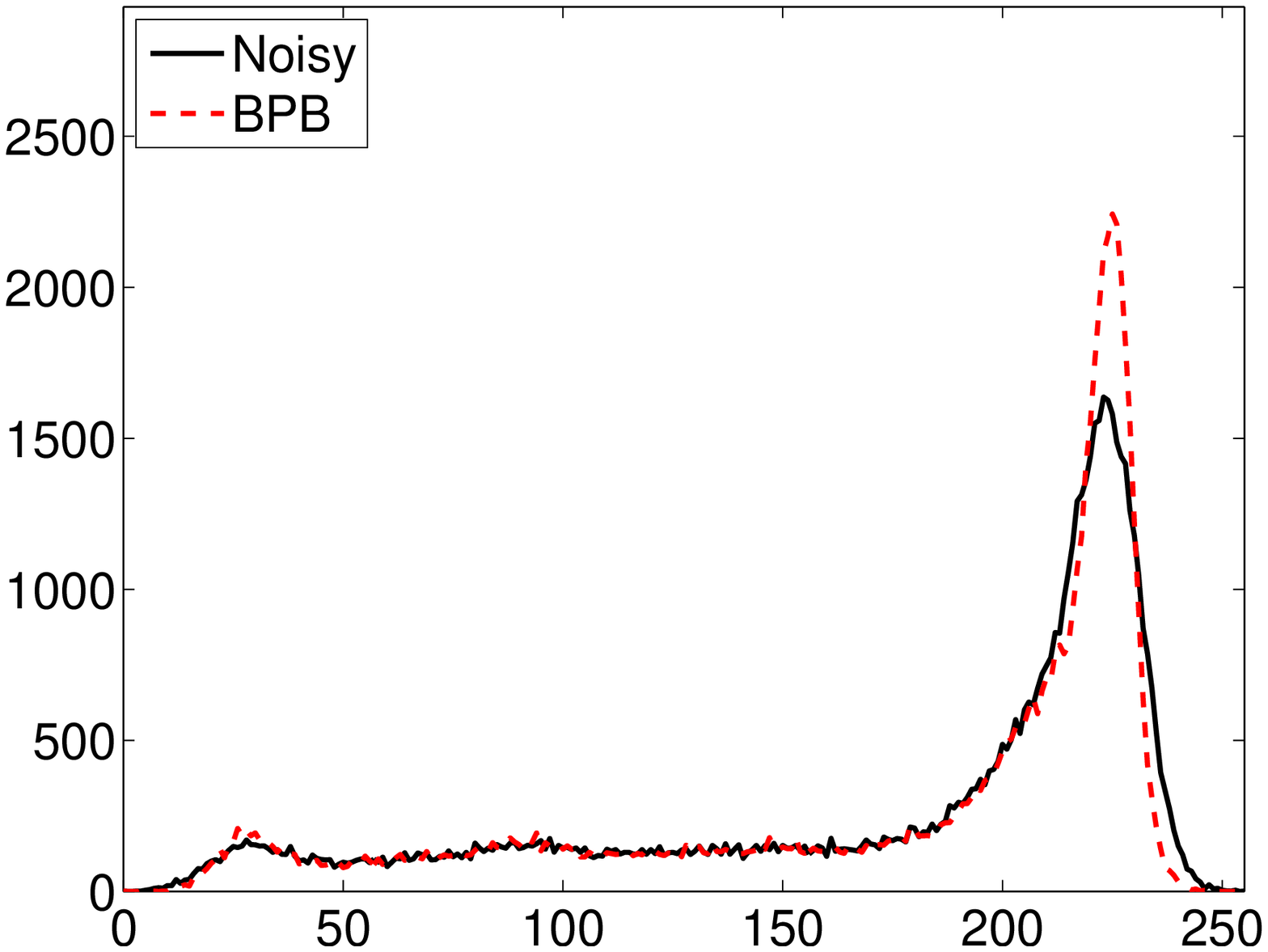}}\\
{\includegraphics[width=3cm,height=3cm]{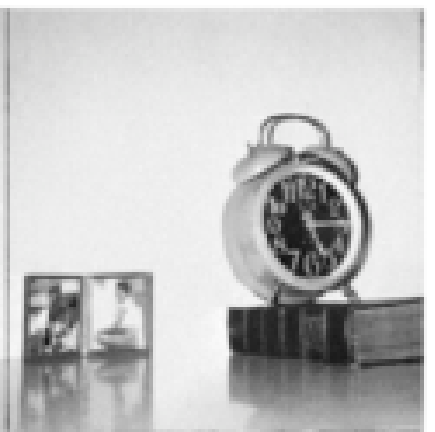}}
{\includegraphics[width=3cm,height=3cm]{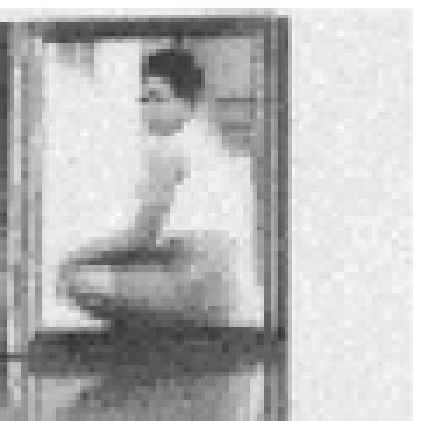}}
{\includegraphics[width=3cm,height=3cm]{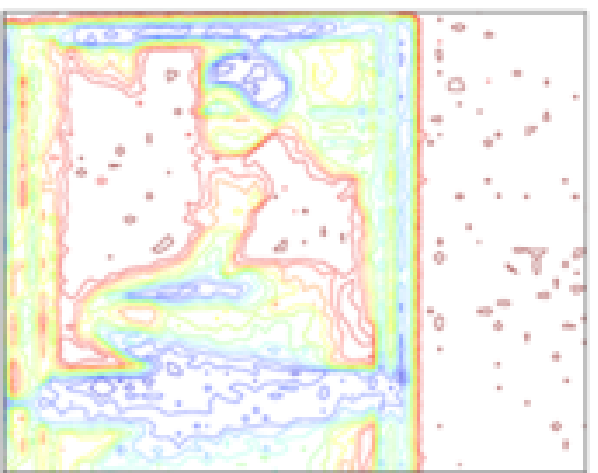}}
{\includegraphics[width=3cm,height=3cm]{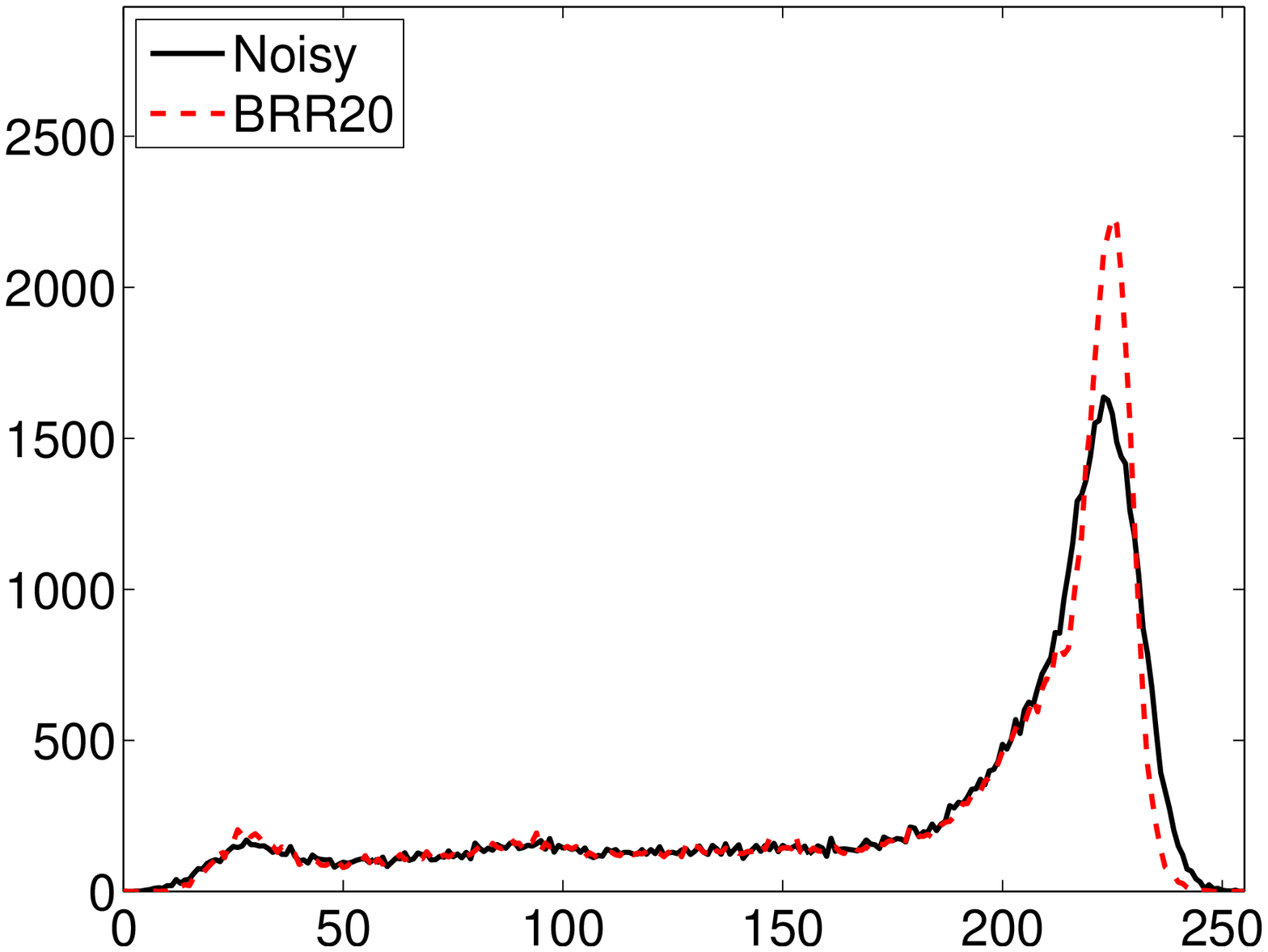}}\\
{\includegraphics[width=3cm,height=3cm]{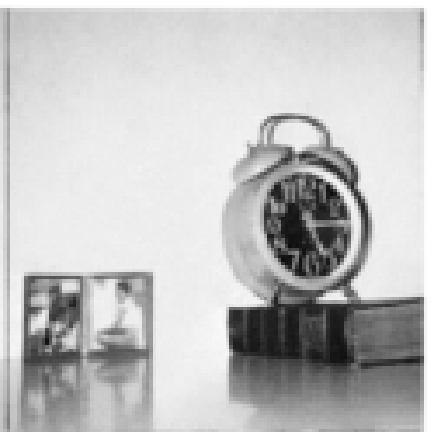}}
{\includegraphics[width=3cm,height=3cm]{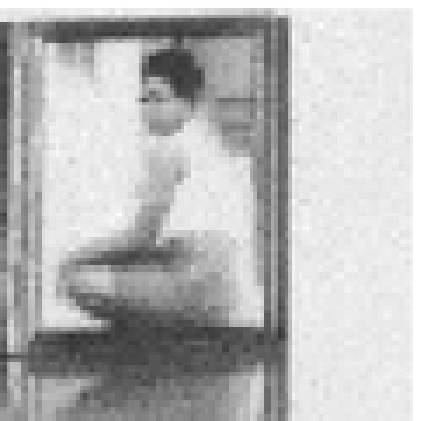}}
{\includegraphics[width=3cm,height=3cm]{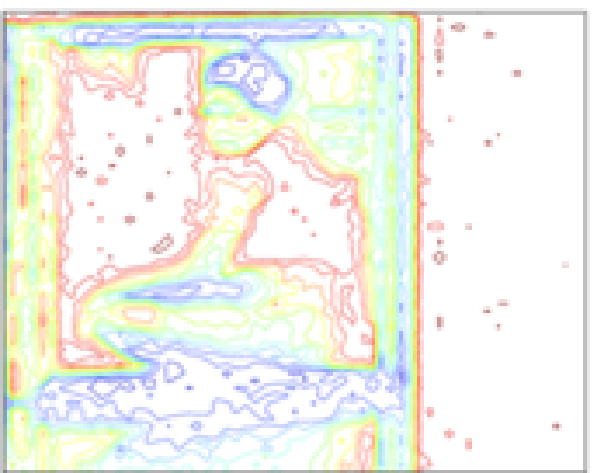}}
{\includegraphics[width=3cm,height=3cm]{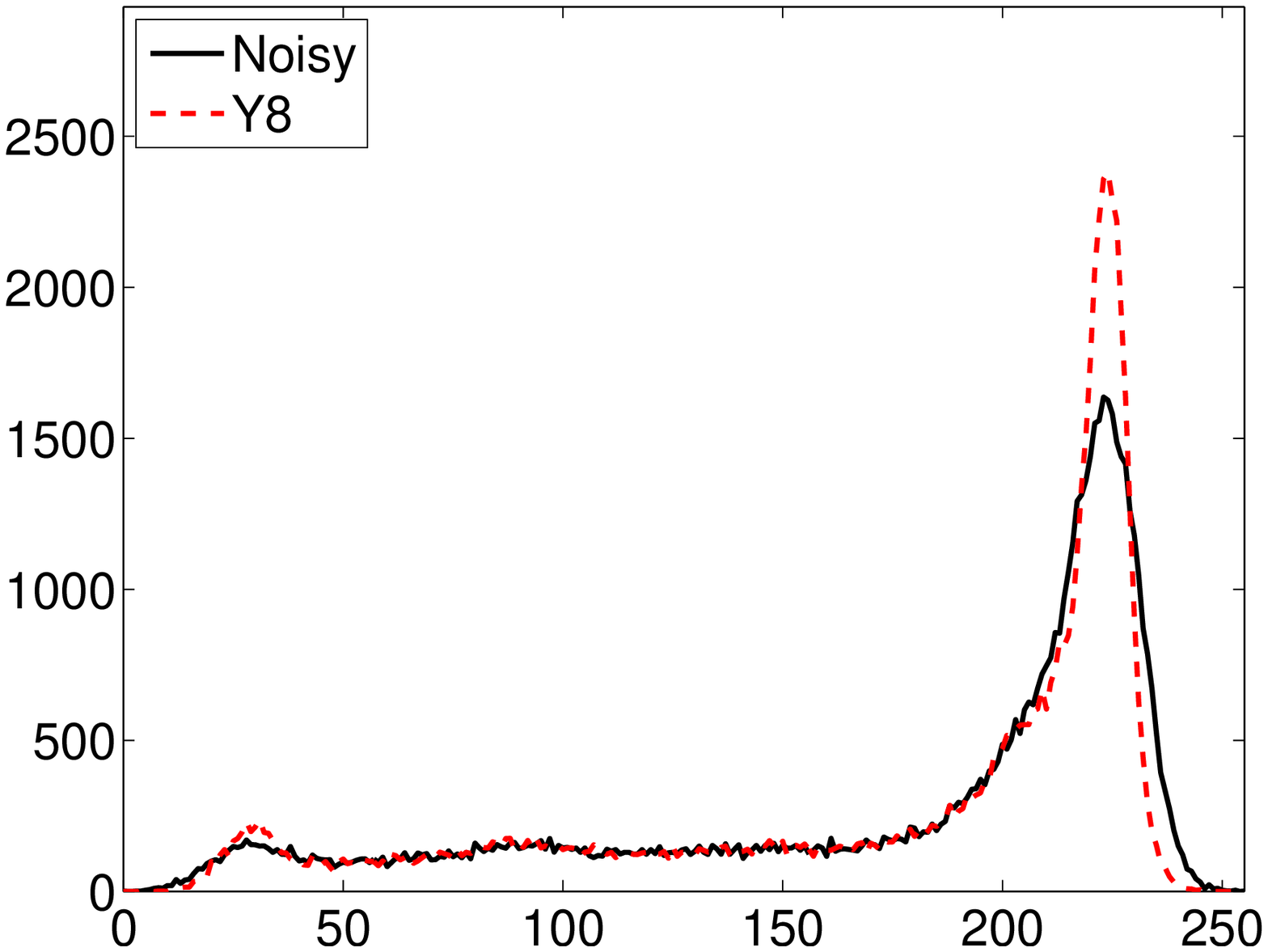}}\\
{\includegraphics[width=3cm,height=3cm]{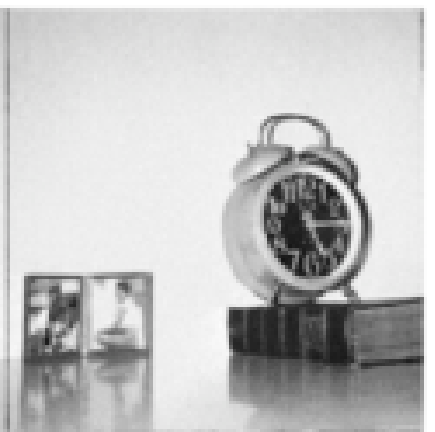}}
{\includegraphics[width=3cm,height=3cm]{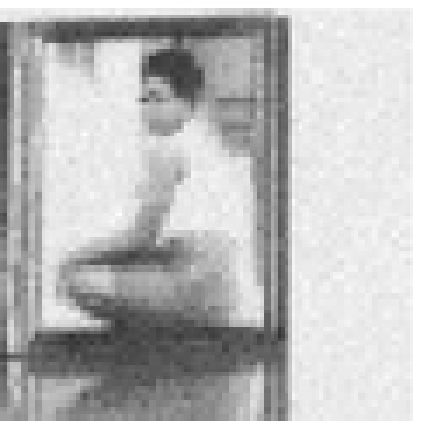}}
{\includegraphics[width=3cm,height=3cm]{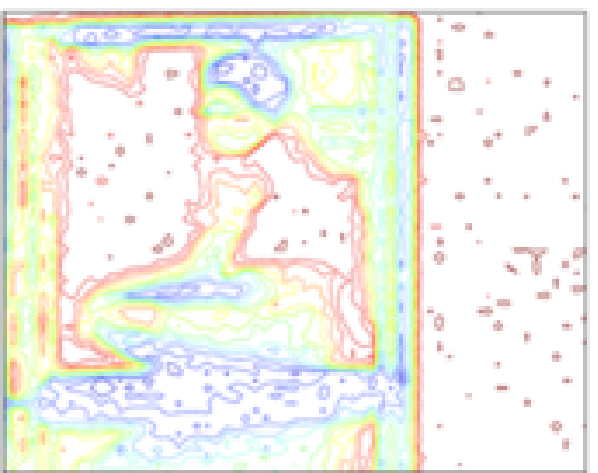}}
{\includegraphics[width=3cm,height=3cm]{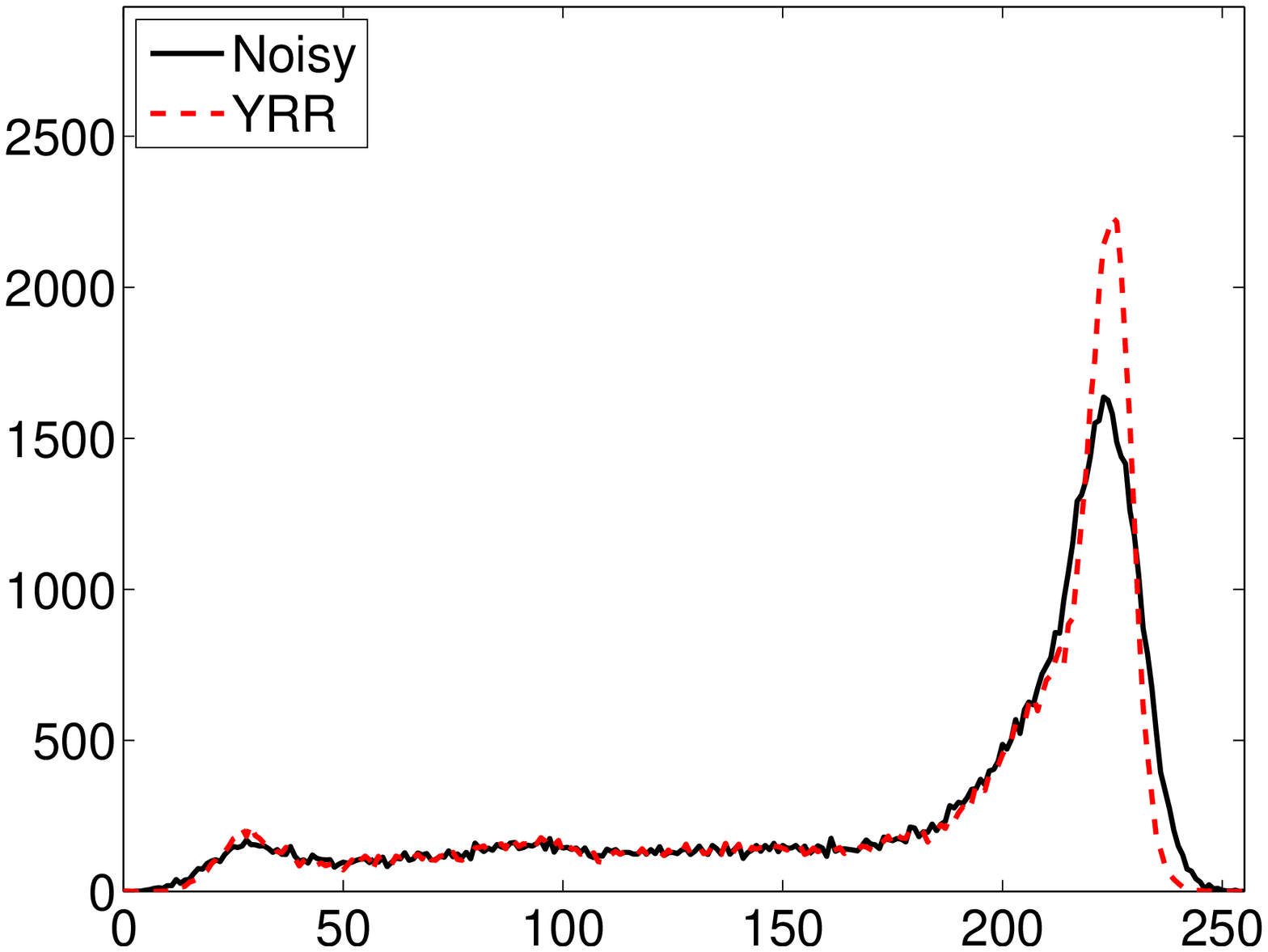}}\\
{\includegraphics[width=3cm,height=3cm]{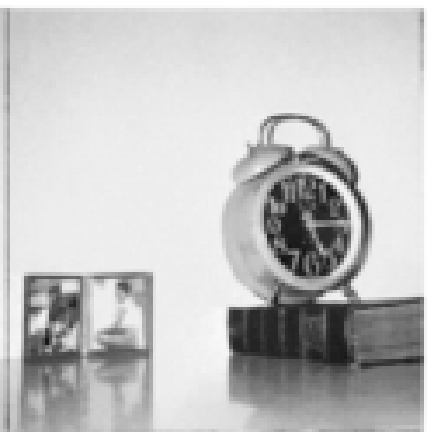}}
{\includegraphics[width=3cm,height=3cm]{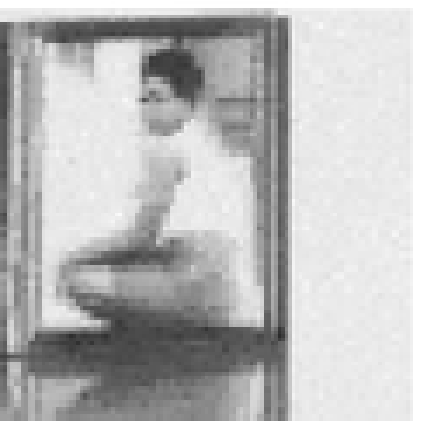}}
{\includegraphics[width=3cm,height=3cm]{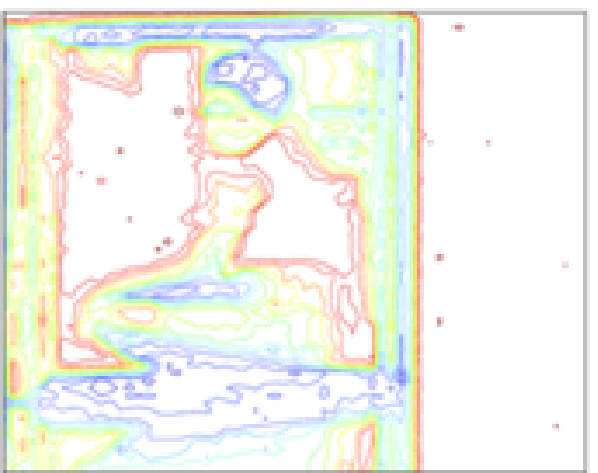}}
{\includegraphics[width=3cm,height=3cm]{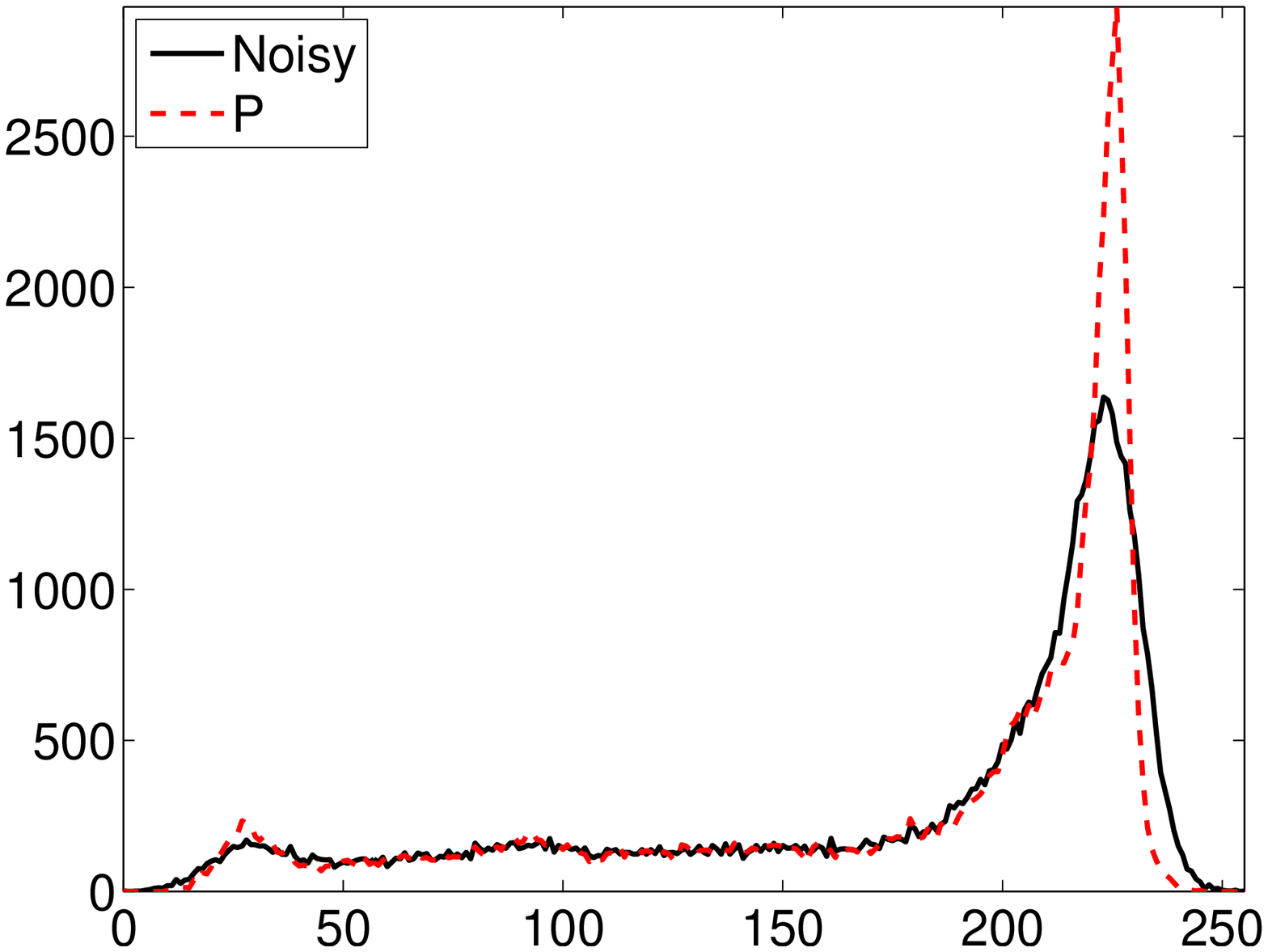}}\\
{\includegraphics[width=3cm,height=3cm]{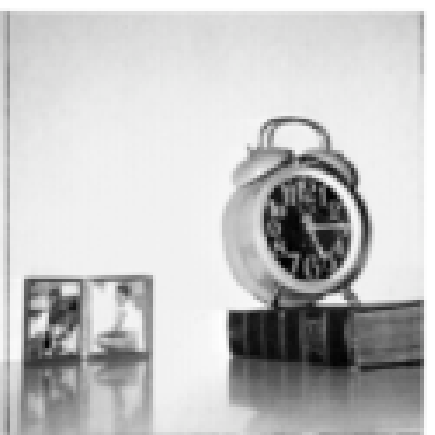}}
{\includegraphics[width=3cm,height=3cm]{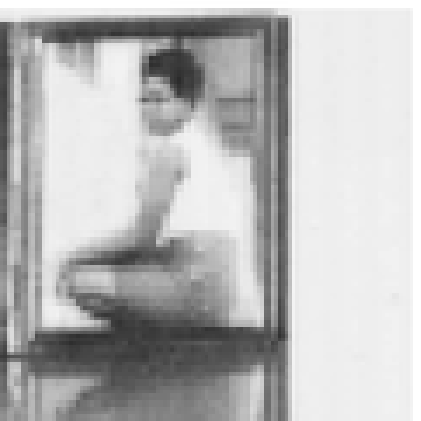}}
{\includegraphics[width=3cm,height=3cm]{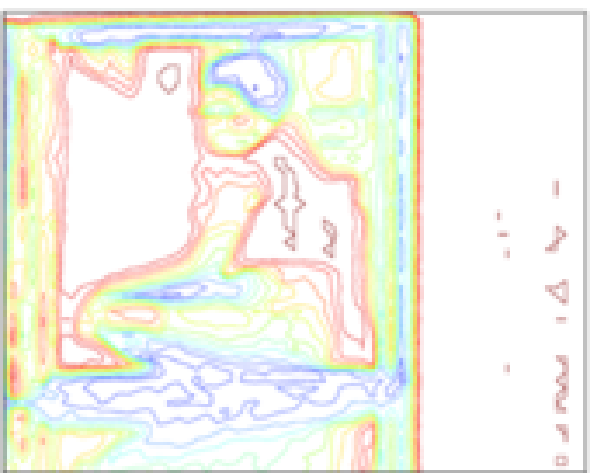}}
{\makebox[3cm]{}}\\
\caption{Image \emph{Clock}, $h=8$. The columns are, from left to right, the 
denoised image, a detail of the image, the contour plot of the detail, and the histogram of the denoised image. The rows give, from above to below, the noisy image, 
the results of BPB, BRR20, Y8, YRR and P, and in the last row, the ground truth clean image.  }
\label{fig_clock} 
\end{figure*}

\begin{figure*}[ht] 
\centering 
{\includegraphics[width=3cm,height=3cm]{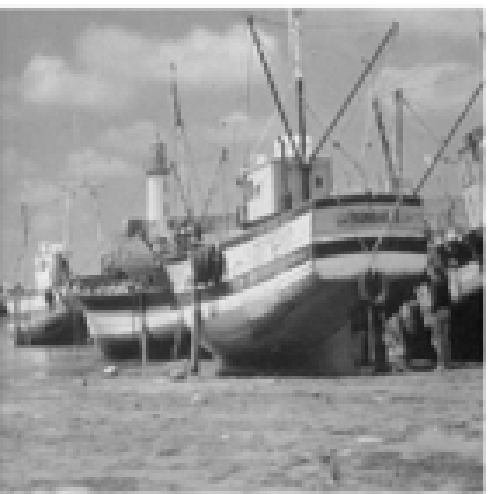}}
{\includegraphics[width=3cm,height=3cm]{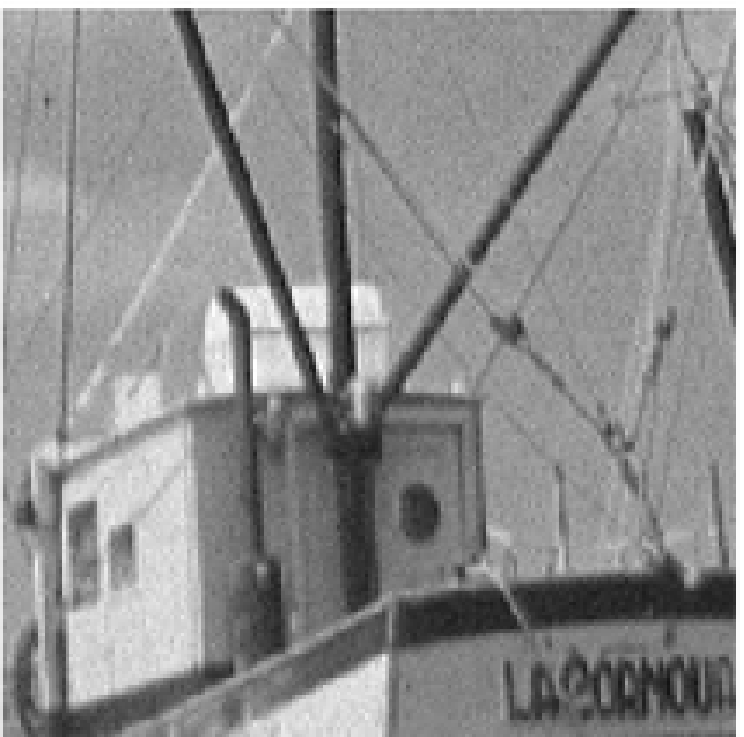}}
{\includegraphics[width=3cm,height=3cm]{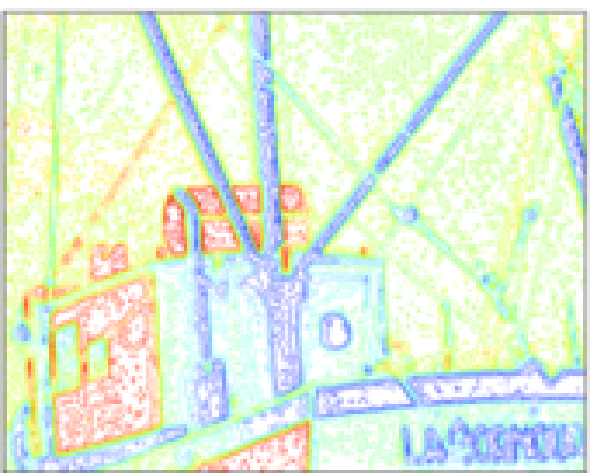}}
{\makebox[3cm]{}}\\
{\includegraphics[width=3cm,height=3cm]{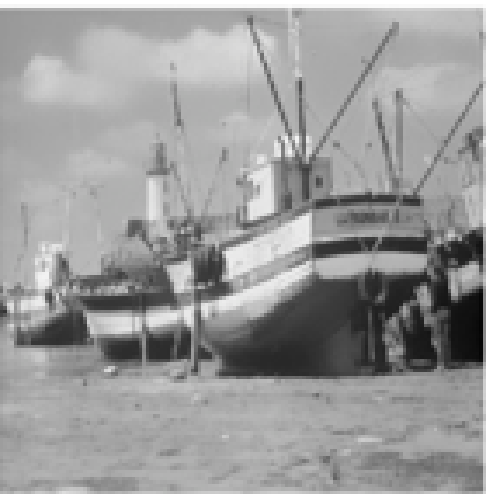}}
{\includegraphics[width=3cm,height=3cm]{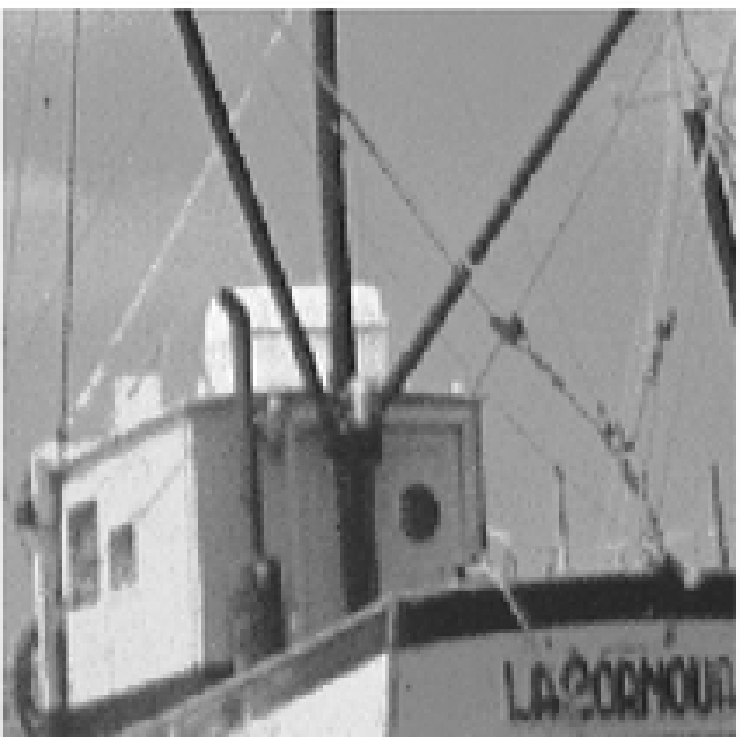}}
{\includegraphics[width=3cm,height=3cm]{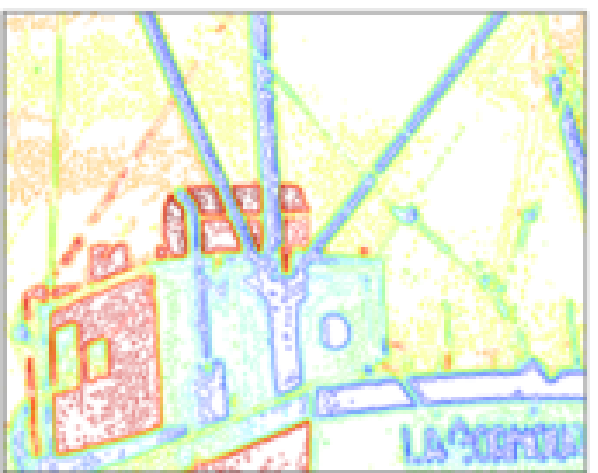}}
{\includegraphics[width=3cm,height=3cm]{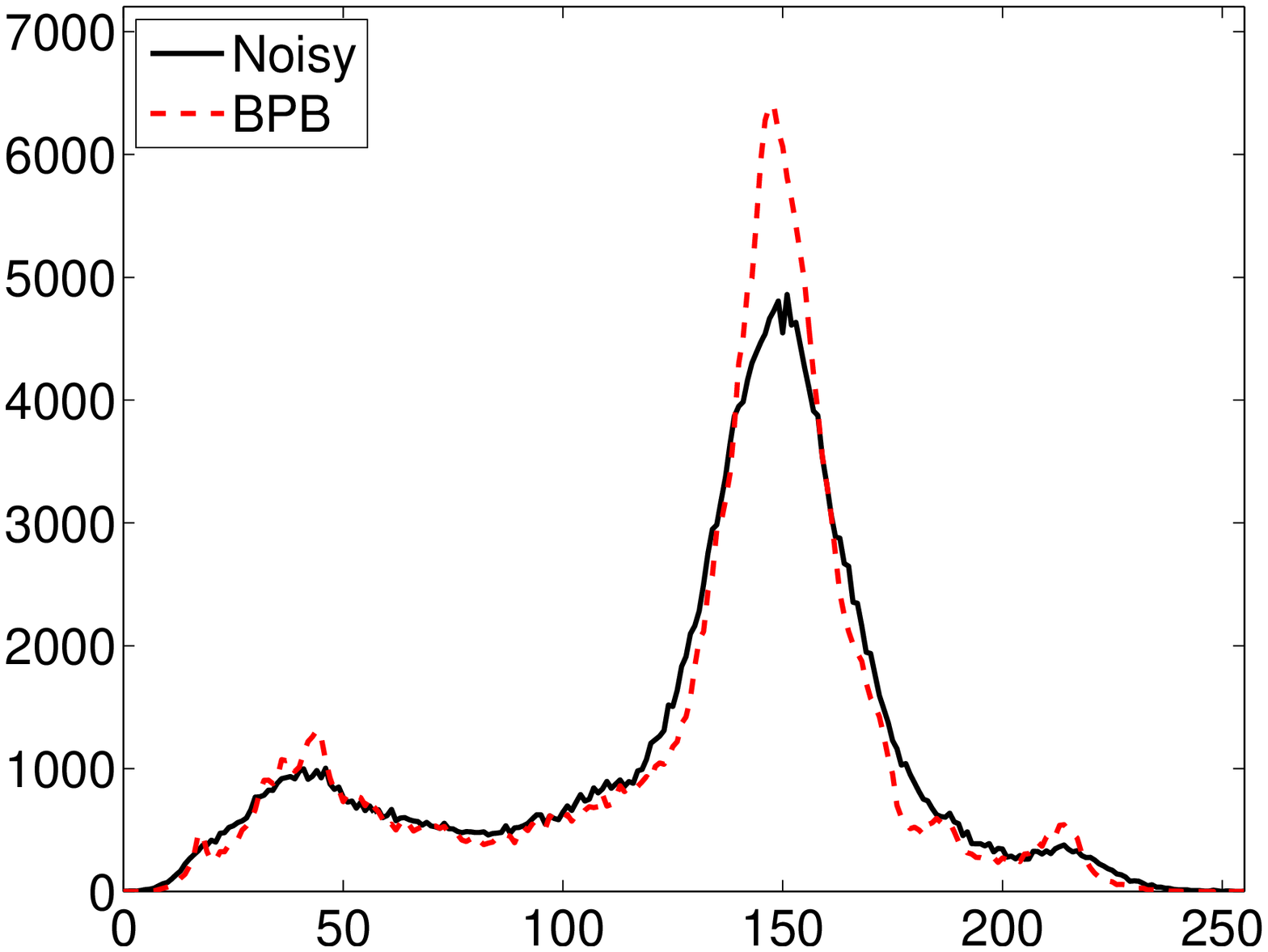}}\\
{\includegraphics[width=3cm,height=3cm]{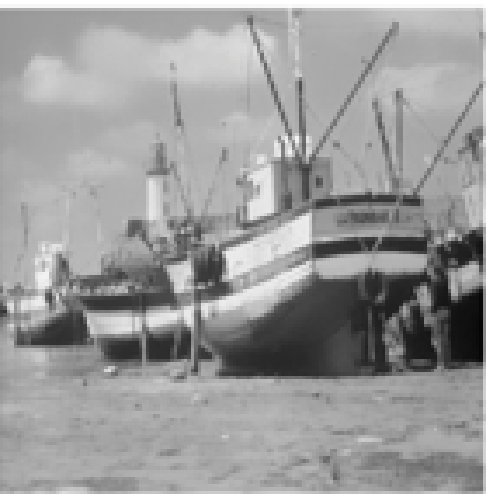}}
{\includegraphics[width=3cm,height=3cm]{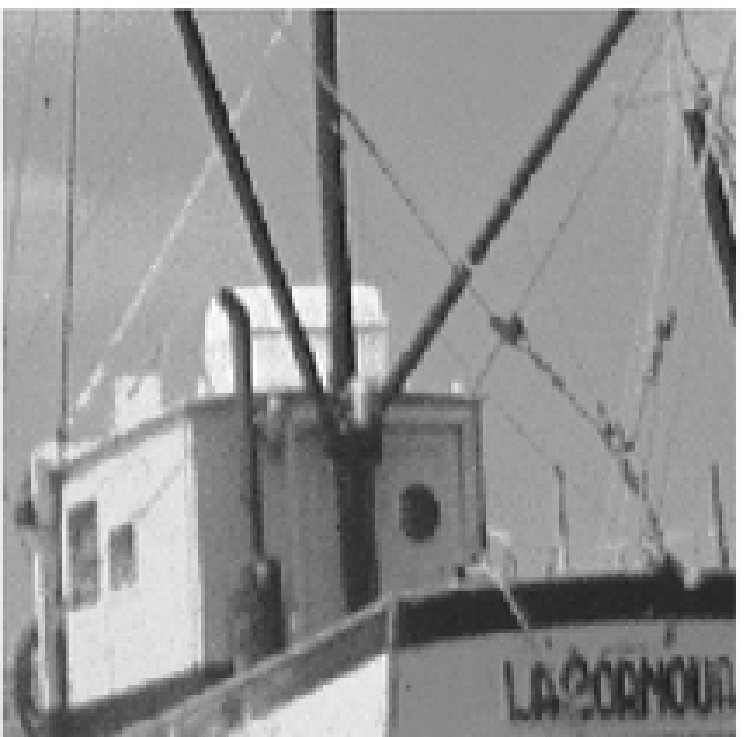}}
{\includegraphics[width=3cm,height=3cm]{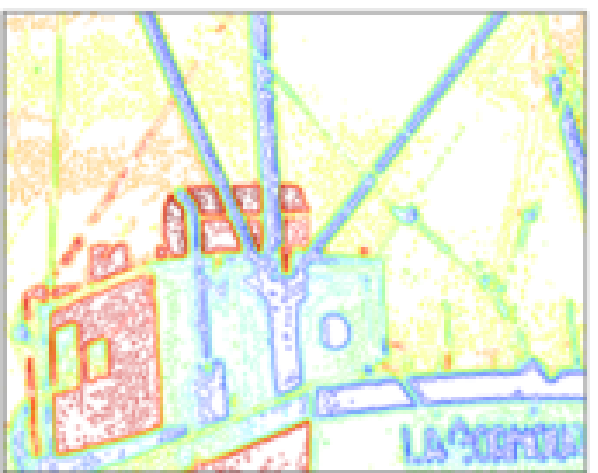}}
{\includegraphics[width=3cm,height=3cm]{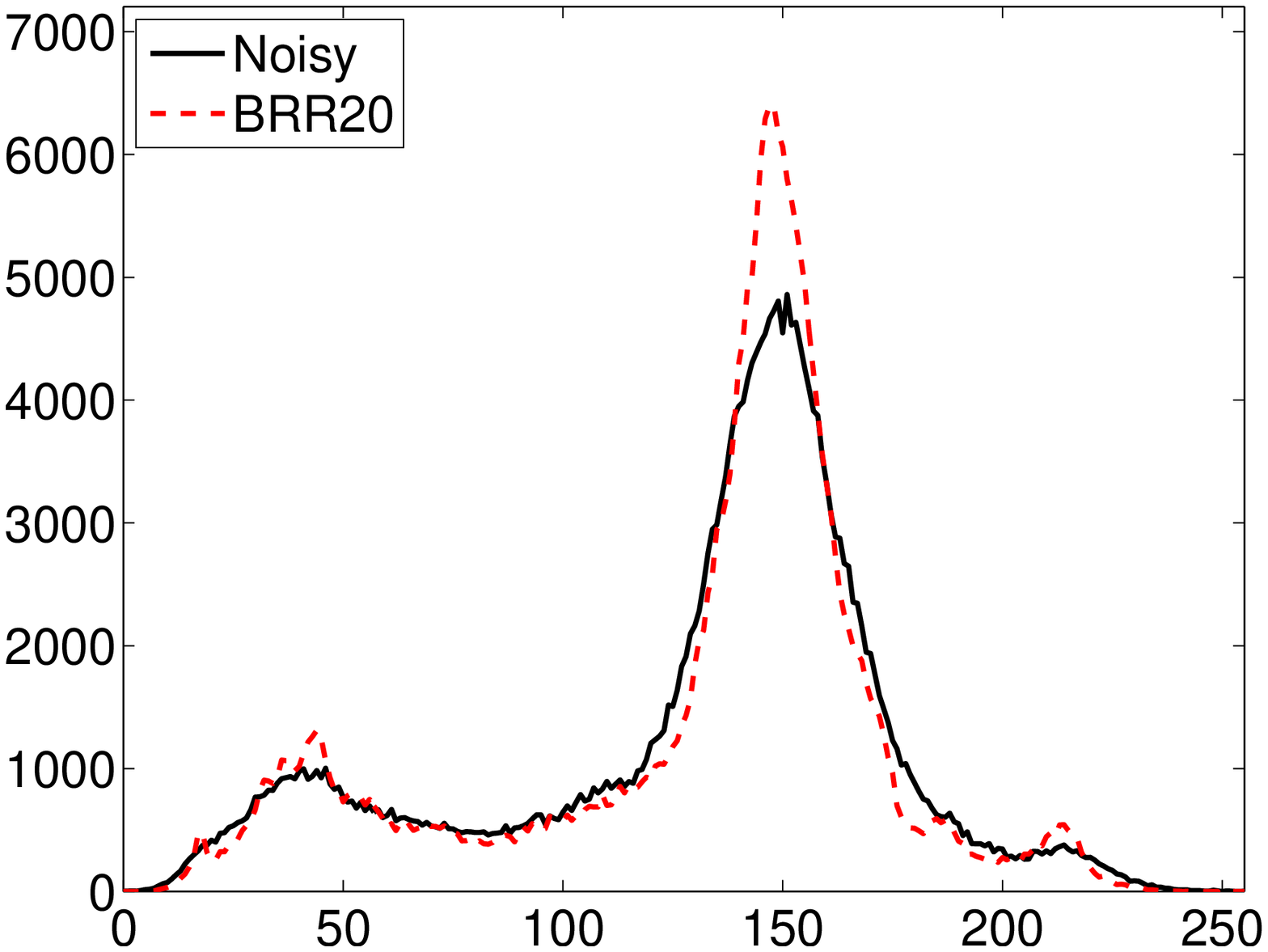}}\\
{\includegraphics[width=3cm,height=3cm]{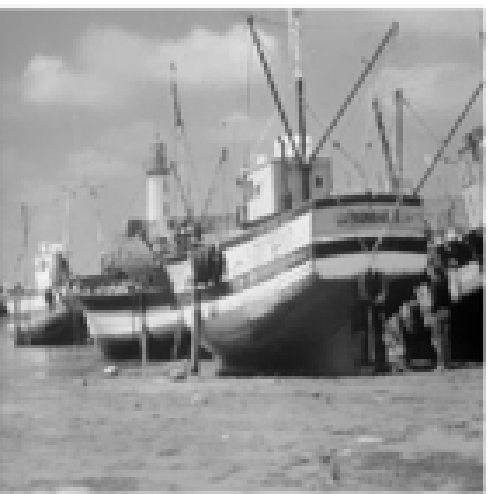}}
{\includegraphics[width=3cm,height=3cm]{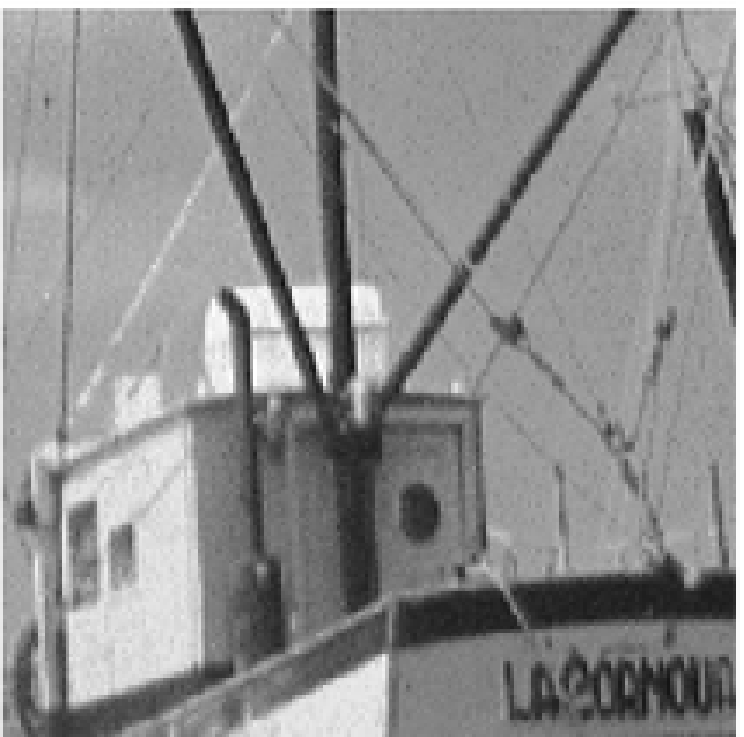}}
{\includegraphics[width=3cm,height=3cm]{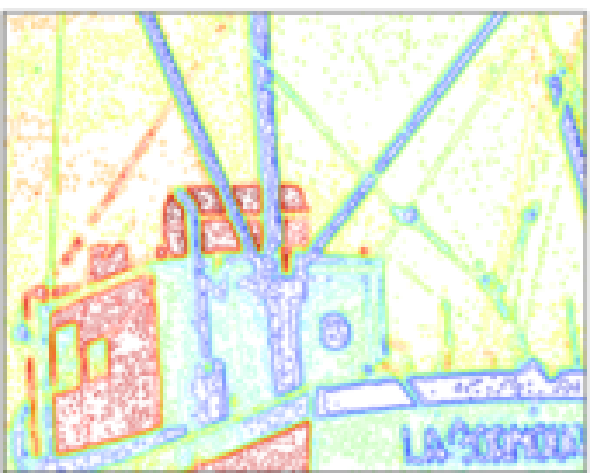}}
{\includegraphics[width=3cm,height=3cm]{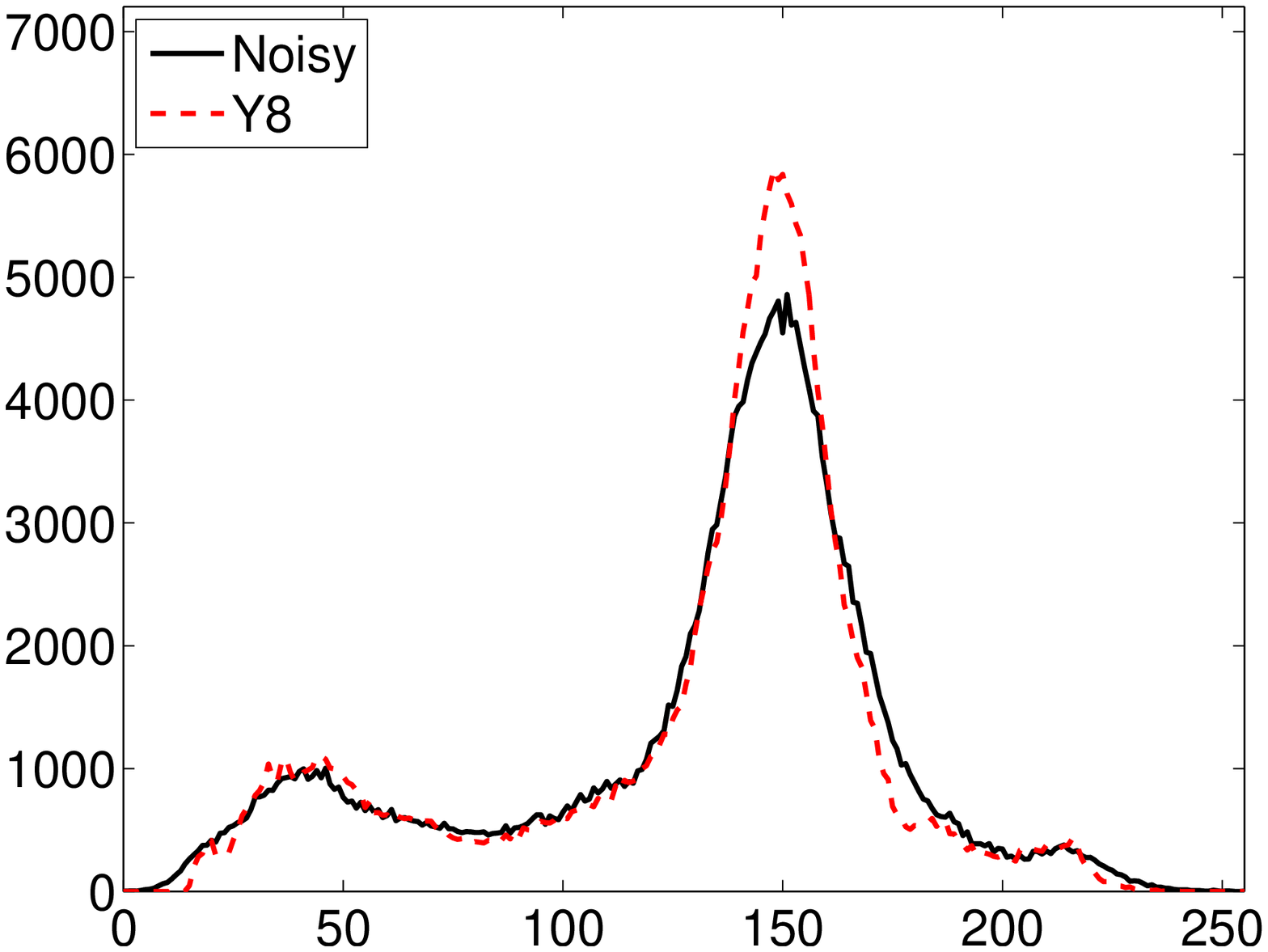}}\\
{\includegraphics[width=3cm,height=3cm]{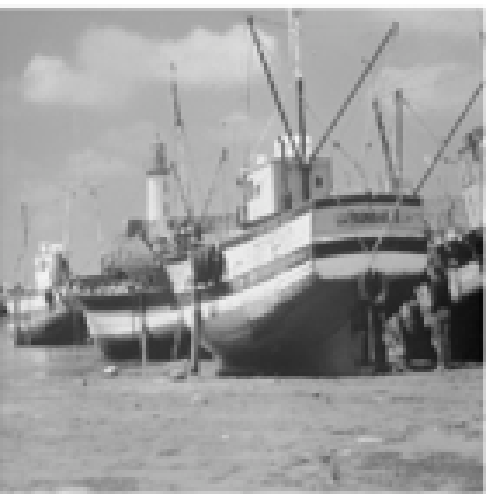}}
{\includegraphics[width=3cm,height=3cm]{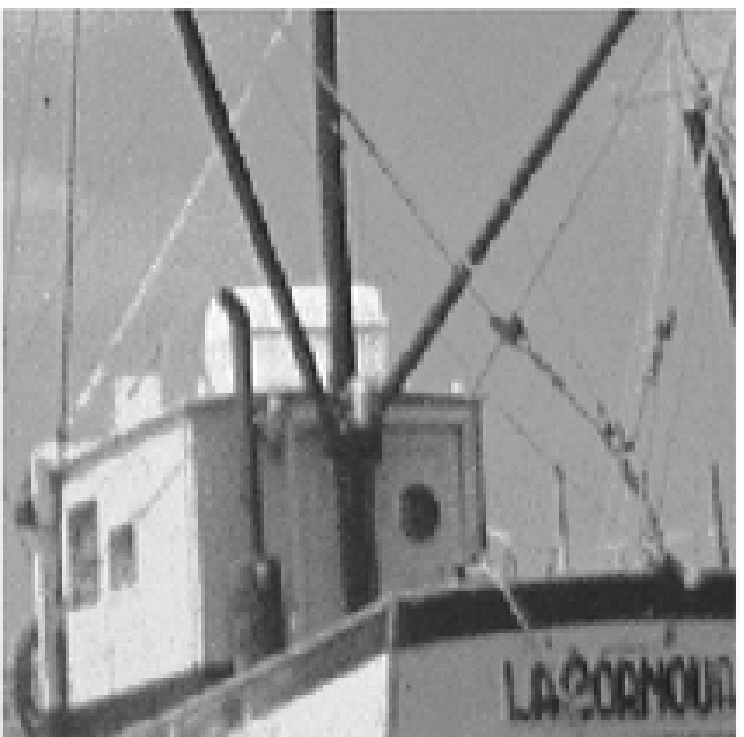}}
{\includegraphics[width=3cm,height=3cm]{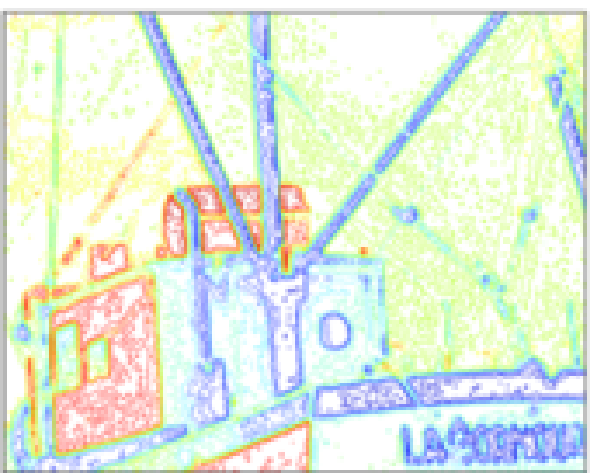}}
{\includegraphics[width=3cm,height=3cm]{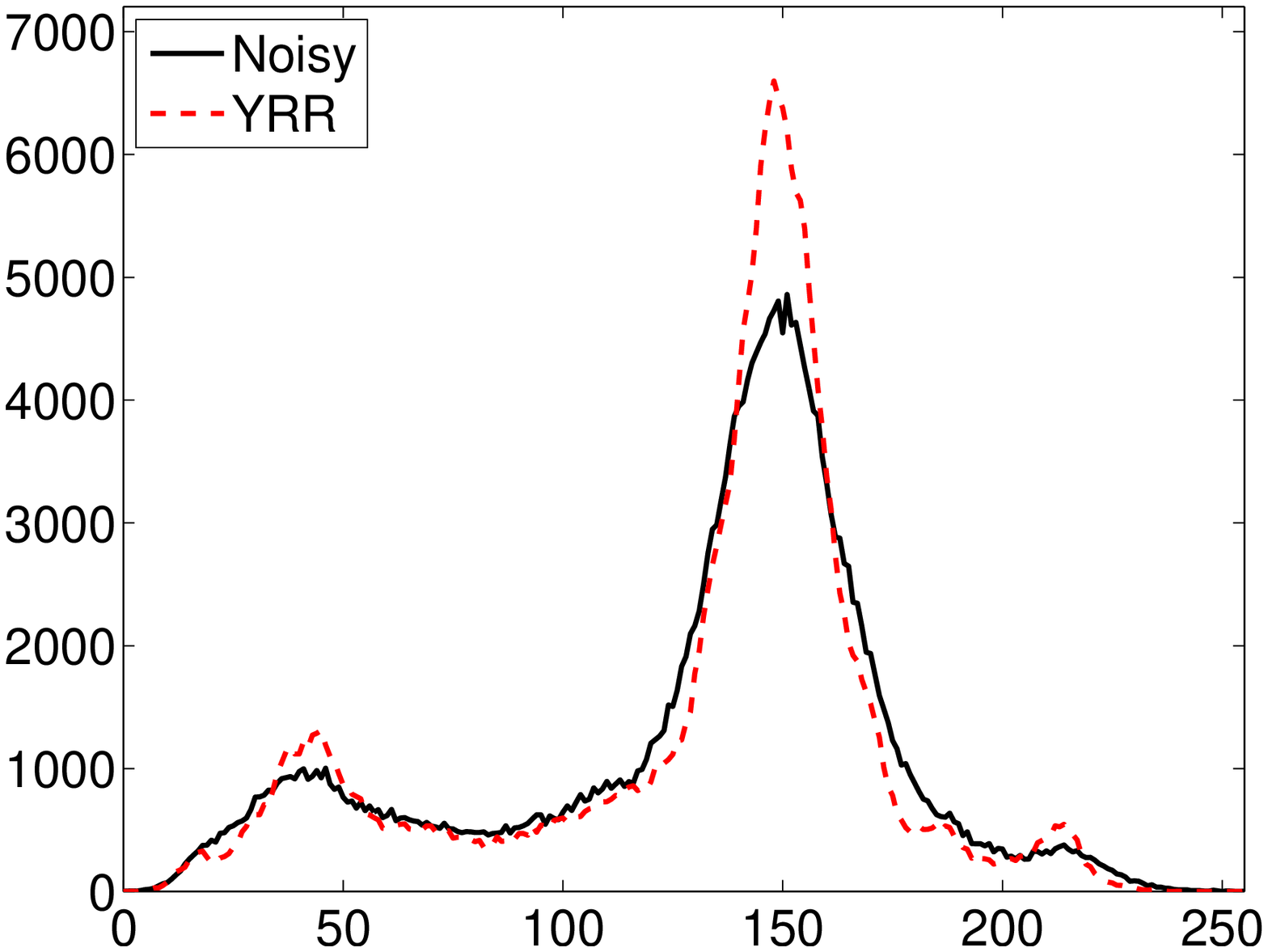}}\\
{\includegraphics[width=3cm,height=3cm]{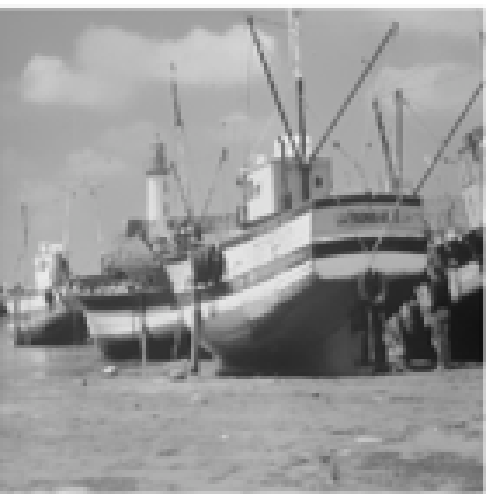}}
{\includegraphics[width=3cm,height=3cm]{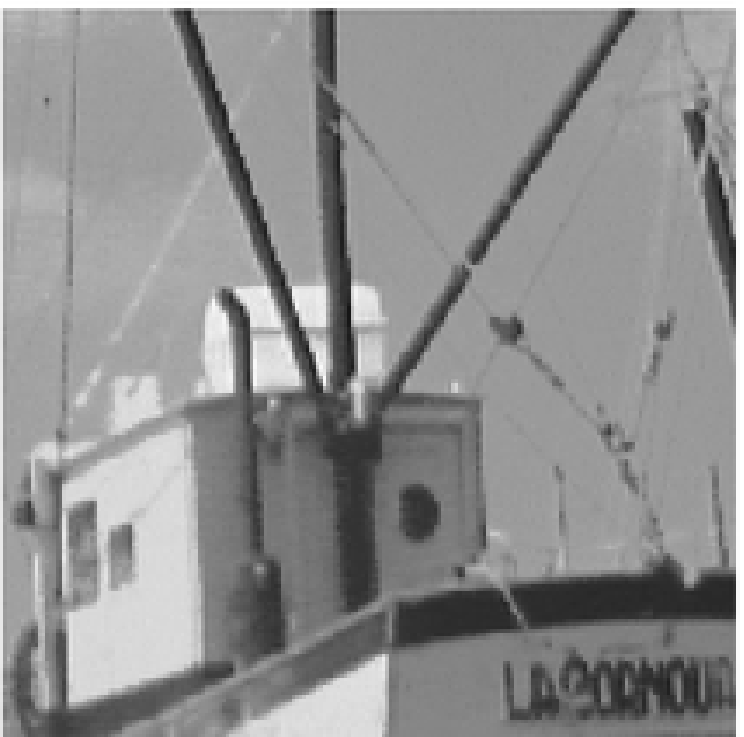}}
{\includegraphics[width=3cm,height=3cm]{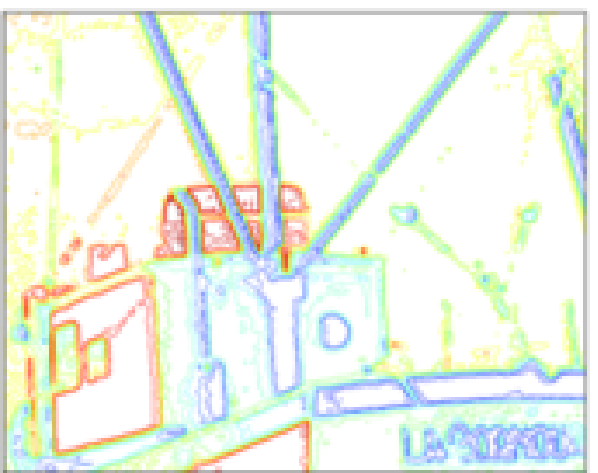}}
{\includegraphics[width=3cm,height=3cm]{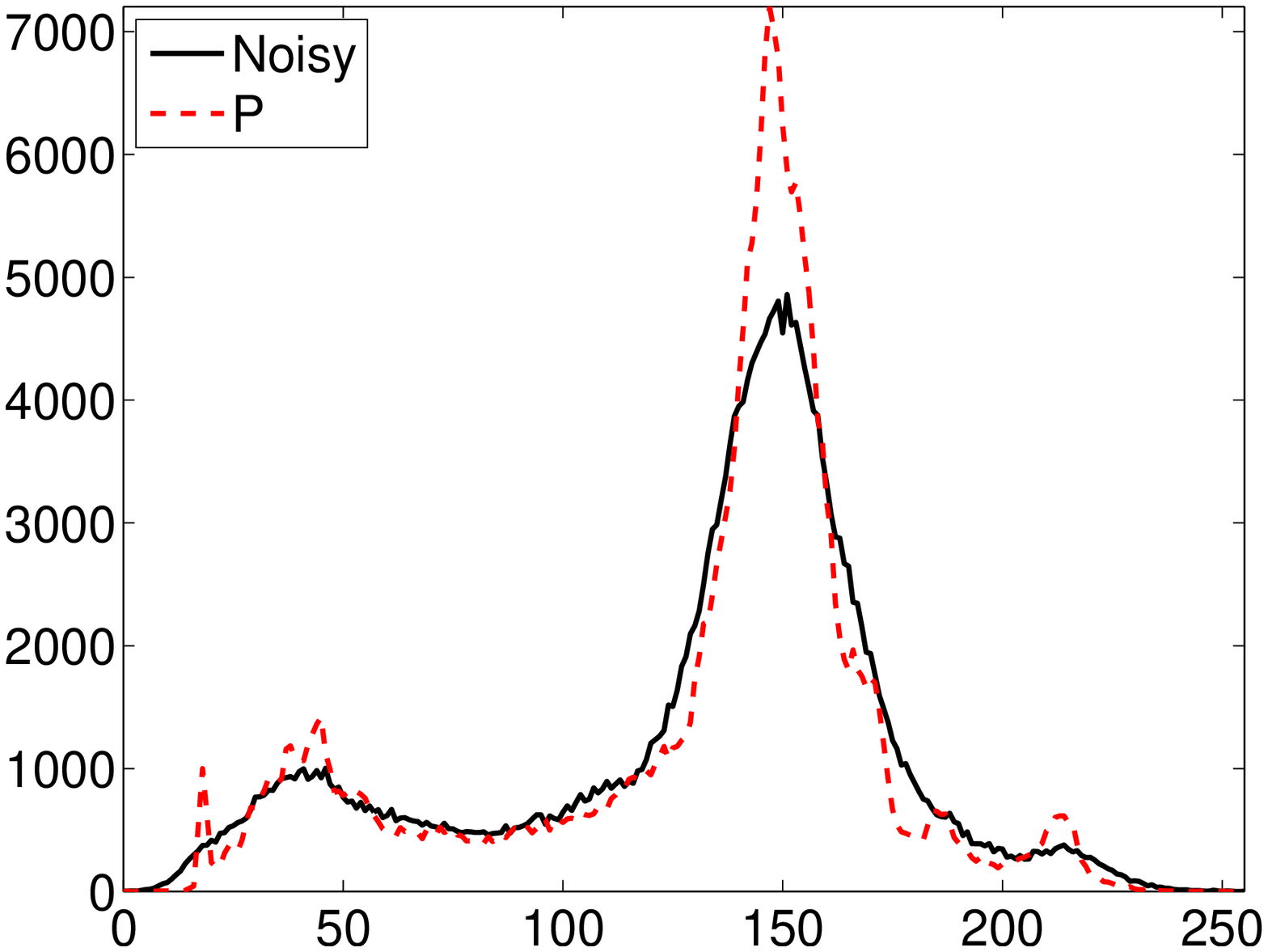}}\\
{\includegraphics[width=3cm,height=3cm]{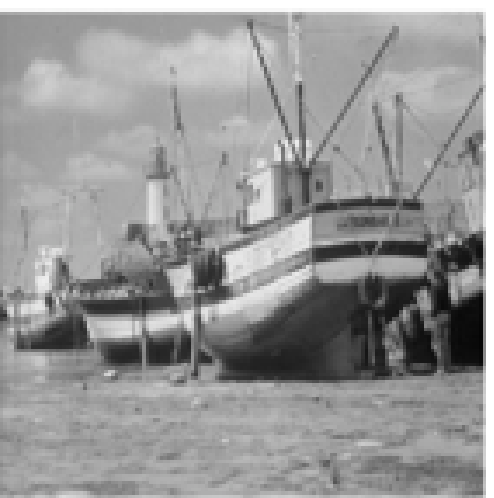}}
{\includegraphics[width=3cm,height=3cm]{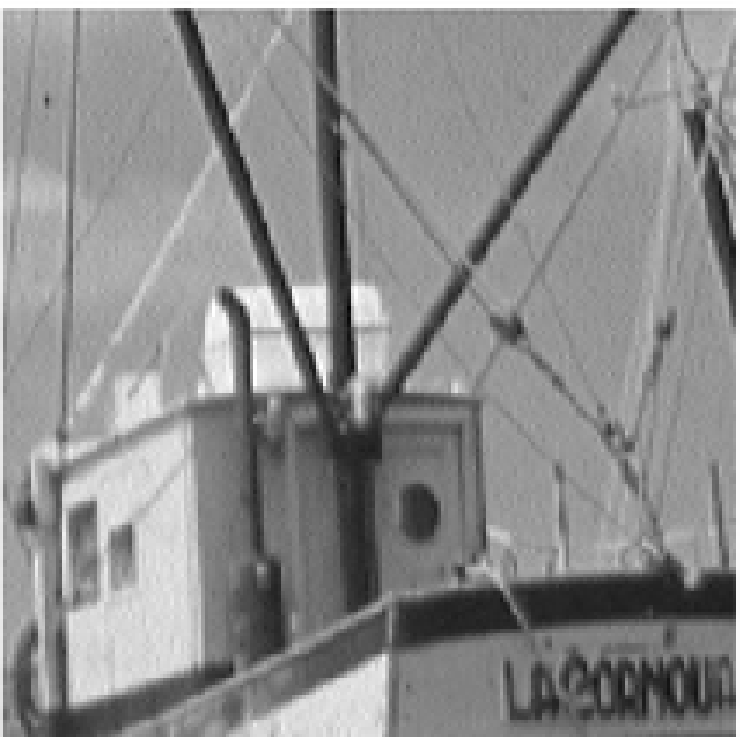}}
{\includegraphics[width=3cm,height=3cm]{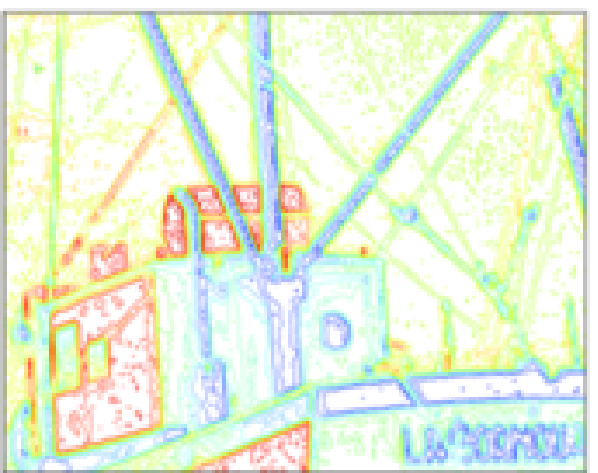}}
{\makebox[3cm]{}}\\
\caption{ Image \emph{Boat}, $h=16$. The columns are, from left to right, the 
denoised image, a detail of the image, the contour plot of the detail, and the histogram of the denoised image. The rows give, from above to below, the noisy image, 
the results of BPB, BRR20, Y8, YRR and P, and in the last row, the ground truth clean image. }
\label{fig_boat} 
\end{figure*}

\begin{figure*}[ht] 
\centering 
{\includegraphics[width=3cm,height=3cm]{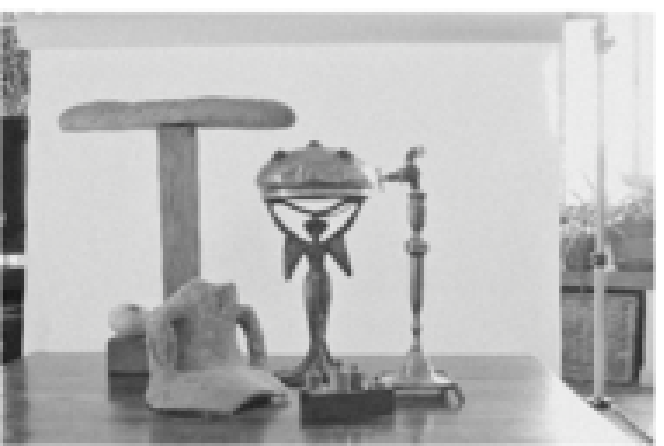}}
{\includegraphics[width=3cm,height=3cm]{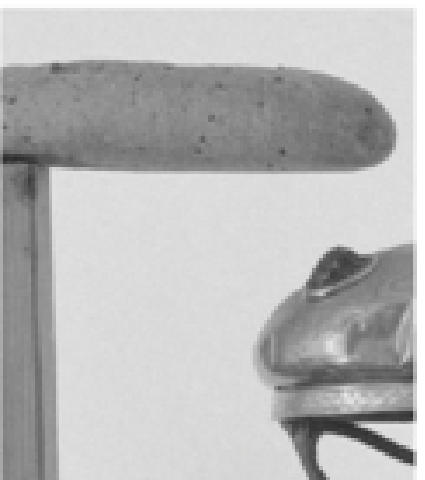}}
{\includegraphics[width=3cm,height=3cm]{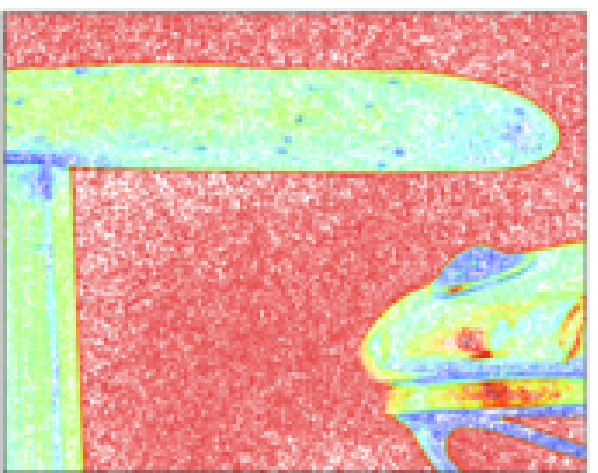}}
{\makebox[3cm]{}}\\
{\includegraphics[width=3cm,height=3cm]{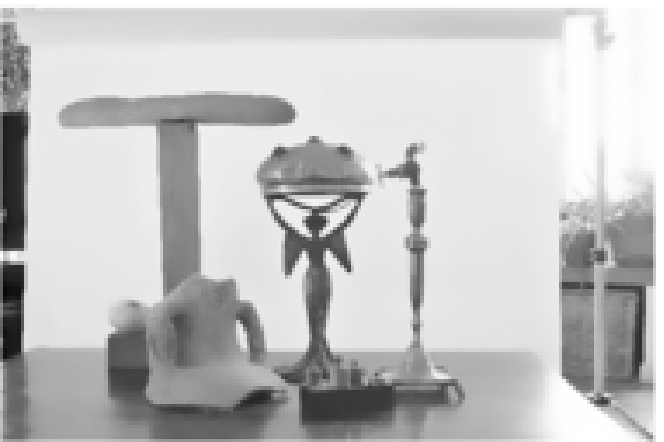}}
{\includegraphics[width=3cm,height=3cm]{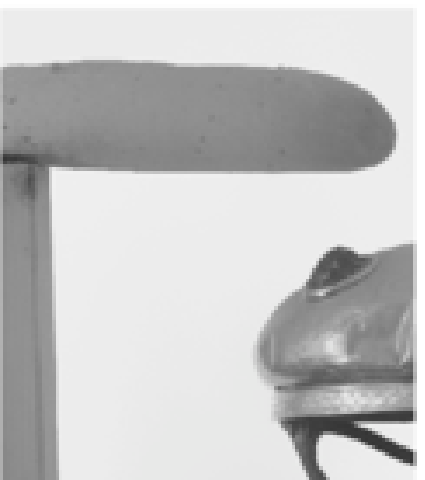}}
{\includegraphics[width=3cm,height=3cm]{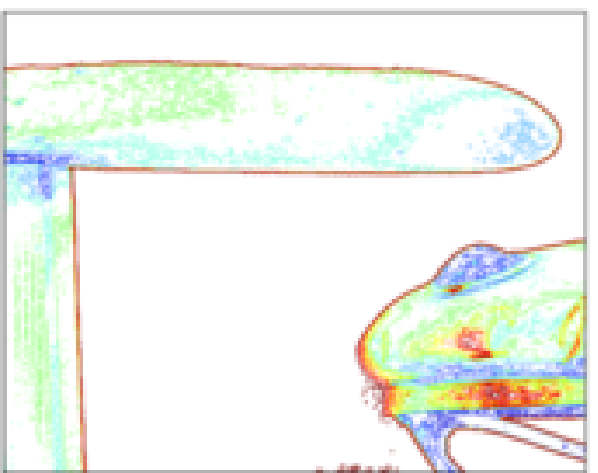}}
{\includegraphics[width=3cm,height=3cm]{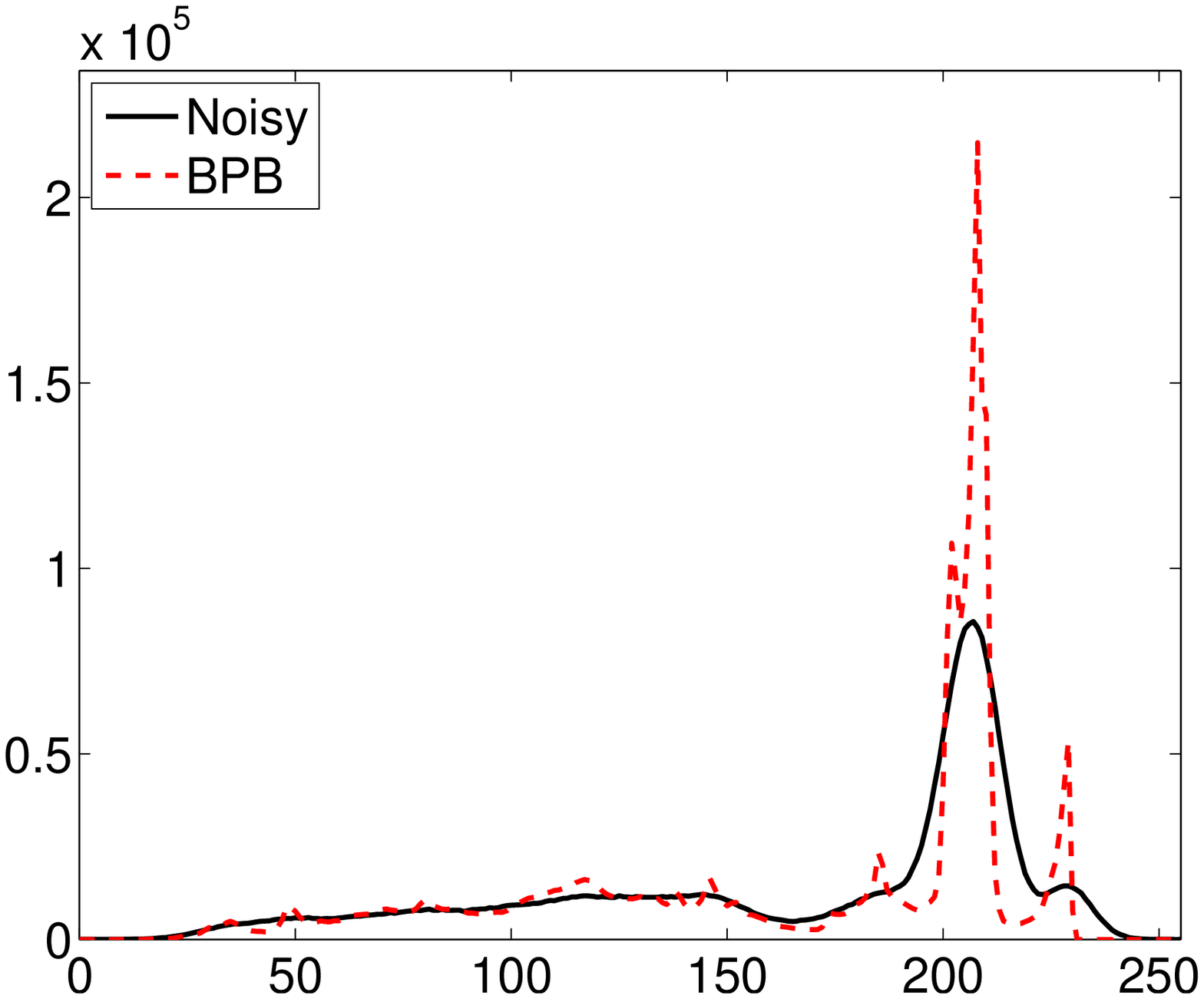}}\\
{\includegraphics[width=3cm,height=3cm]{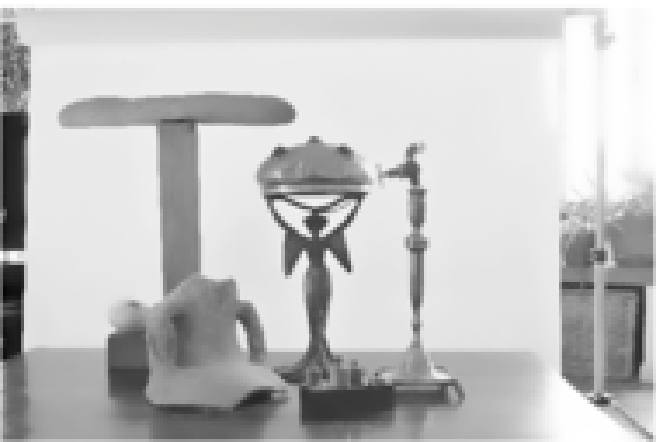}}
{\includegraphics[width=3cm,height=3cm]{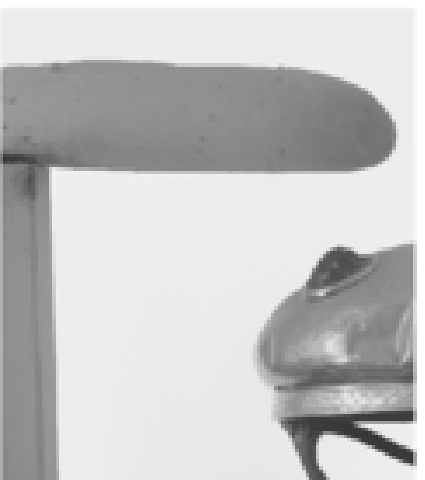}}
{\includegraphics[width=3cm,height=3cm]{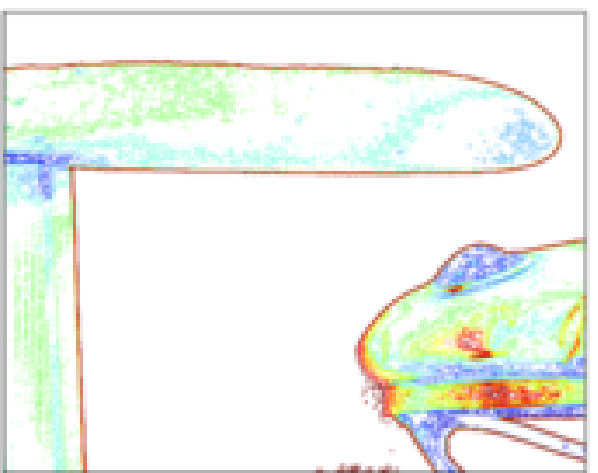}}
{\includegraphics[width=3cm,height=3cm]{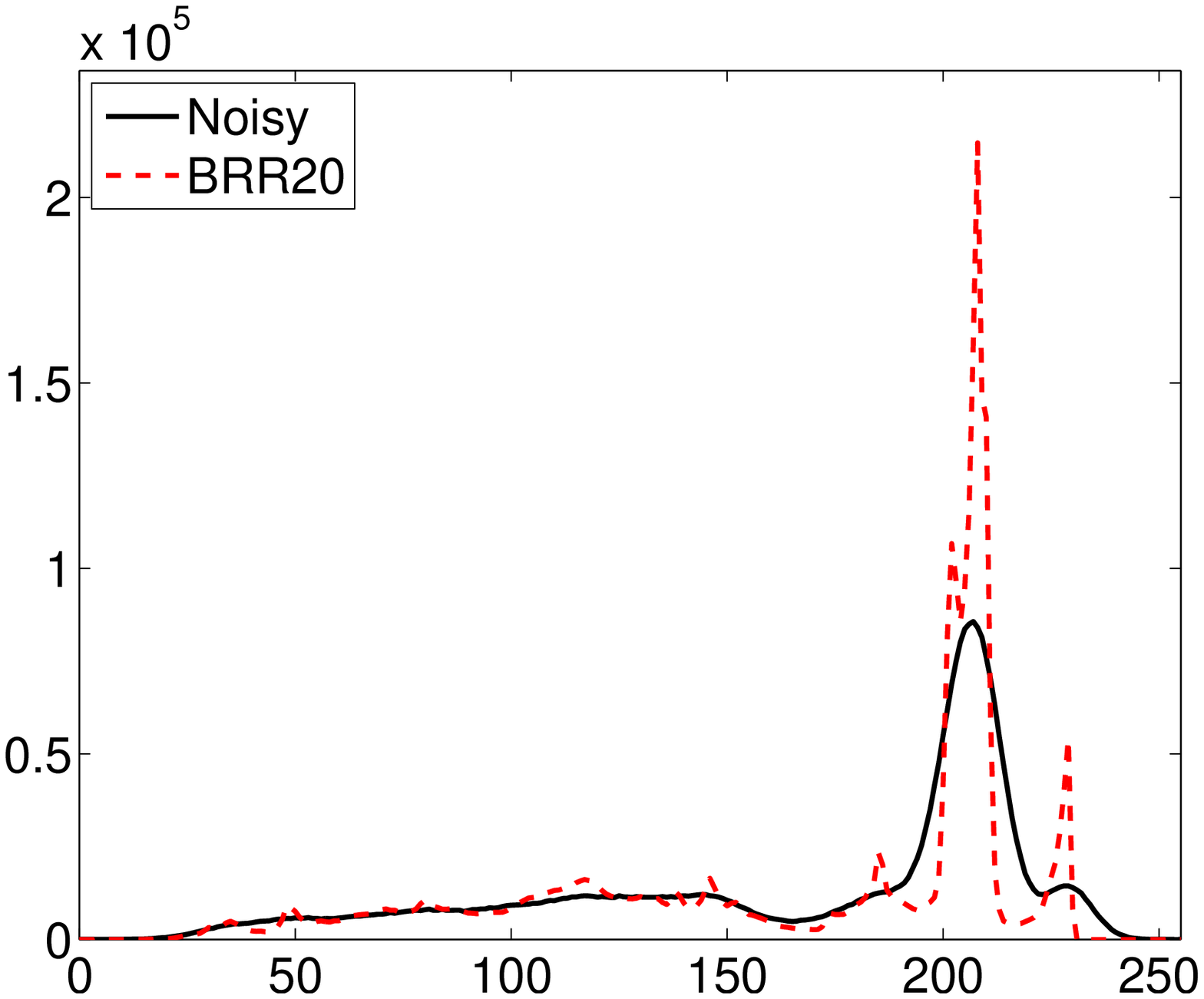}}\\
{\includegraphics[width=3cm,height=3cm]{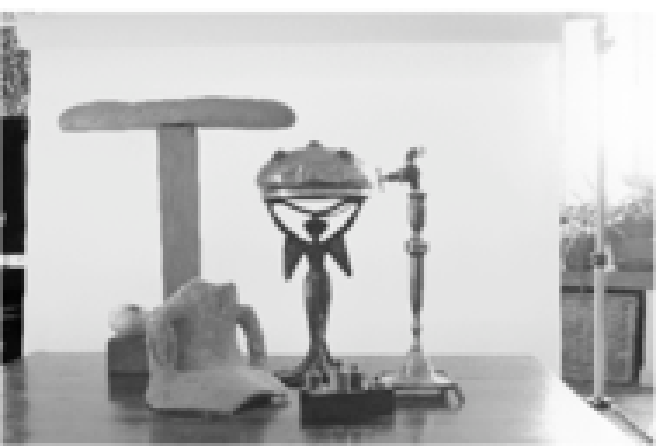}}
{\includegraphics[width=3cm,height=3cm]{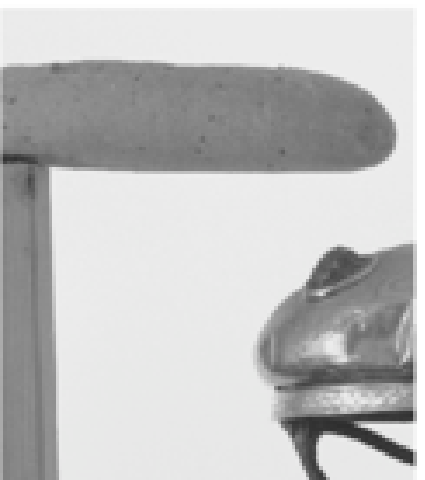}}
{\includegraphics[width=3cm,height=3cm]{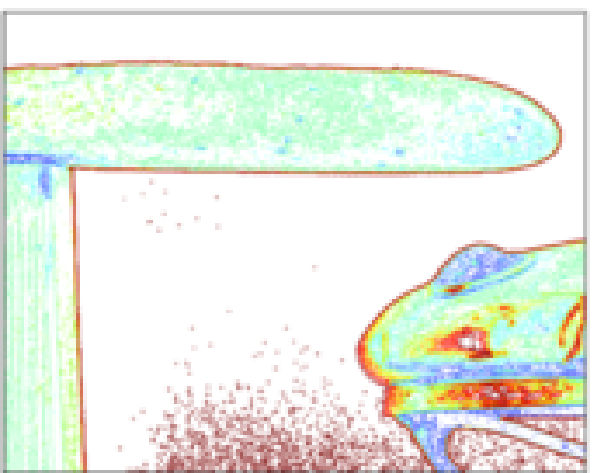}}
{\includegraphics[width=3cm,height=3cm]{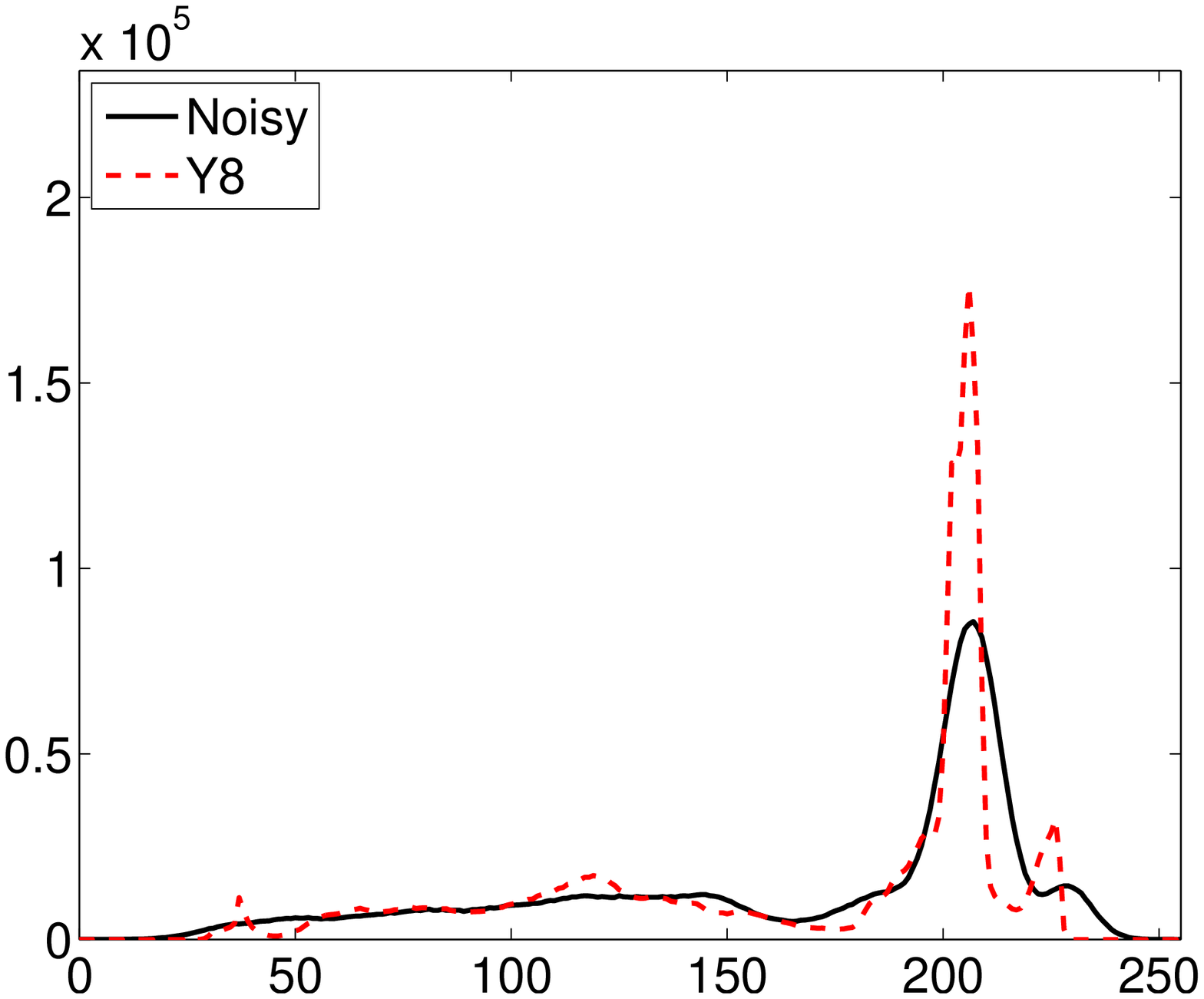}}\\
{\includegraphics[width=3cm,height=3cm]{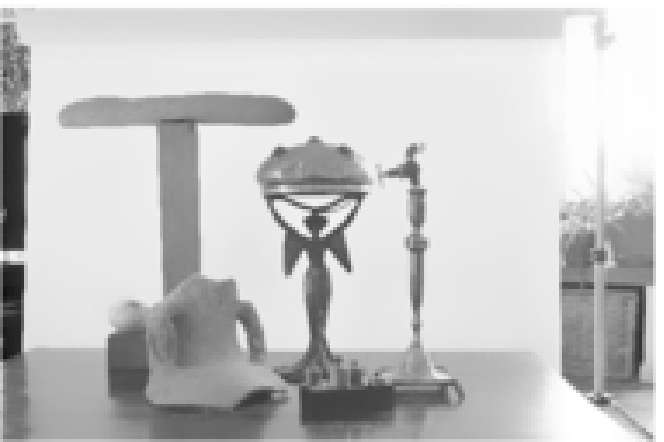}}
{\includegraphics[width=3cm,height=3cm]{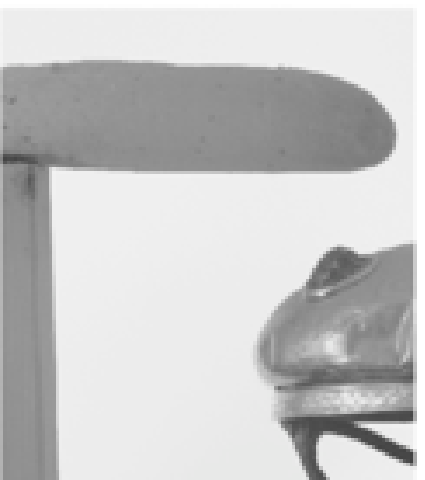}}
{\includegraphics[width=3cm,height=3cm]{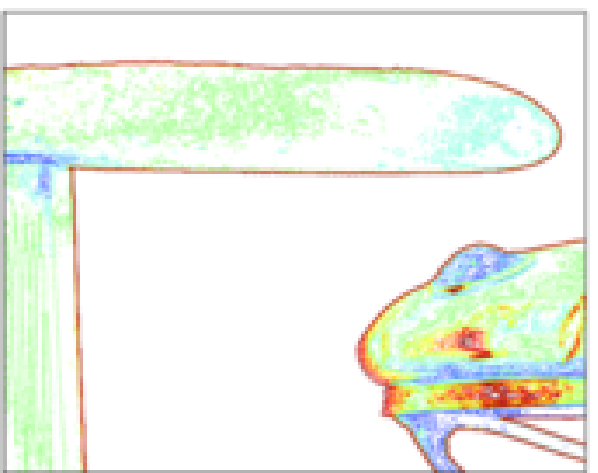}}
{\includegraphics[width=3cm,height=3cm]{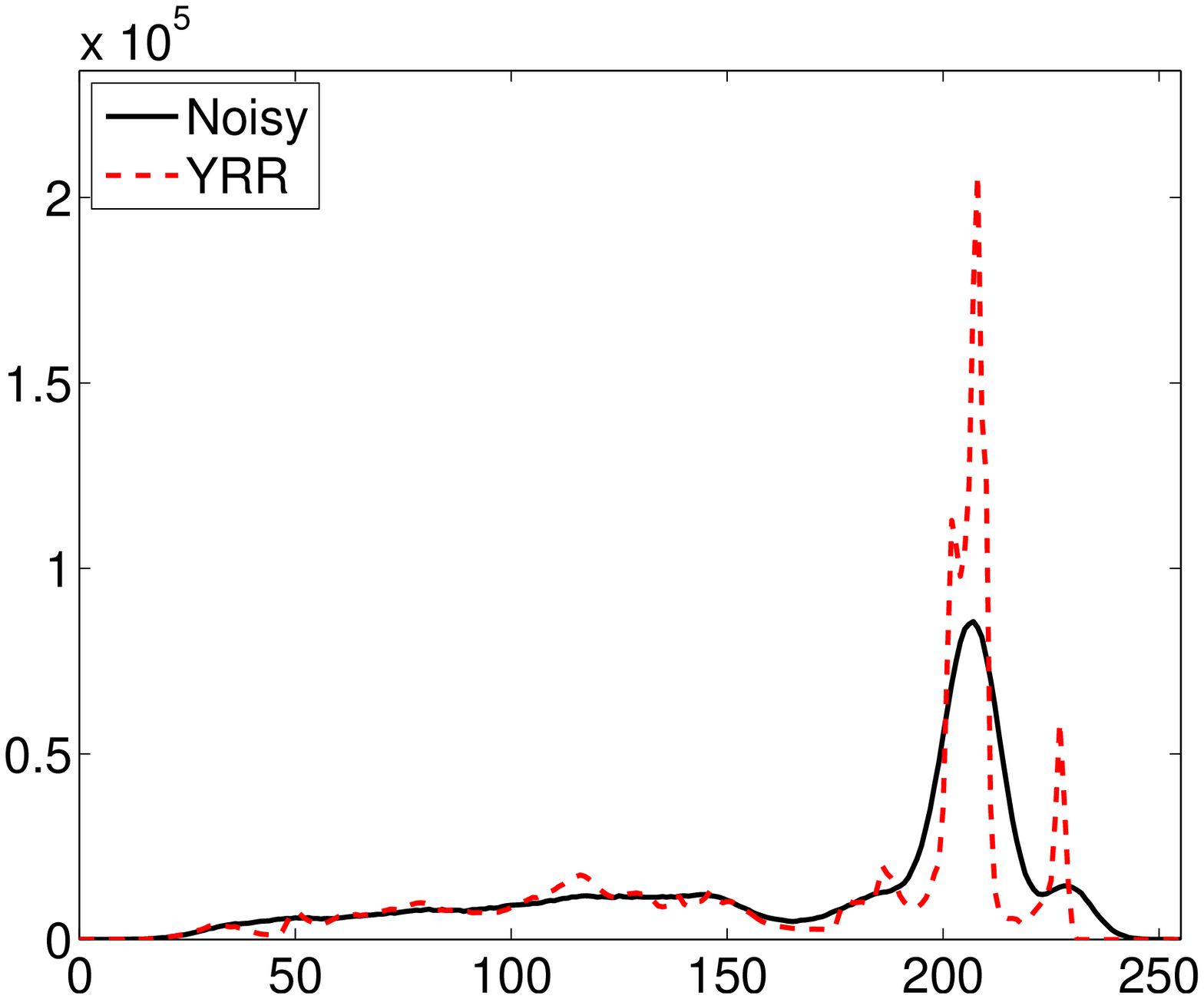}}\\
{\includegraphics[width=3cm,height=3cm]{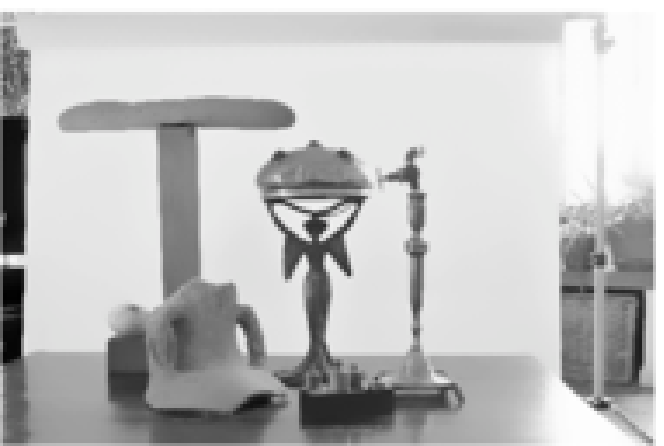}}
{\includegraphics[width=3cm,height=3cm]{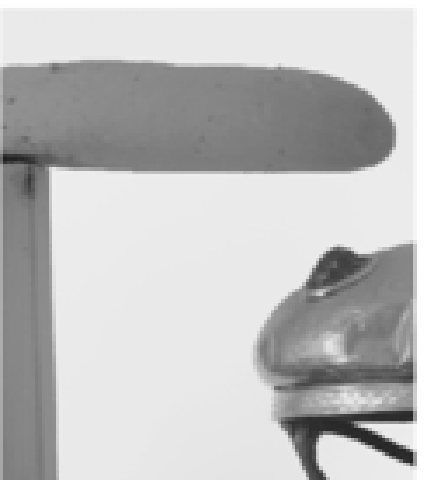}}
{\includegraphics[width=3cm,height=3cm]{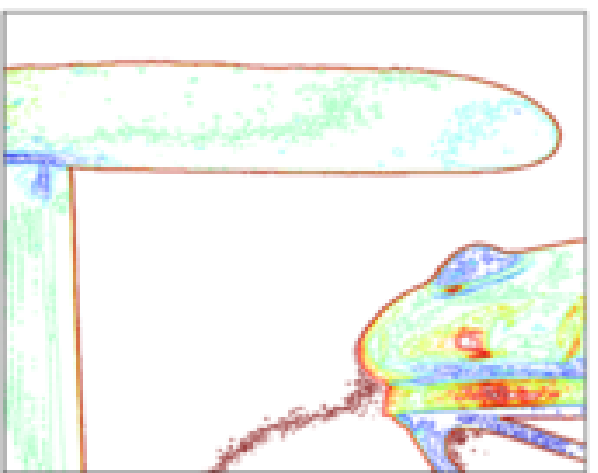}}
{\includegraphics[width=3cm,height=3cm]{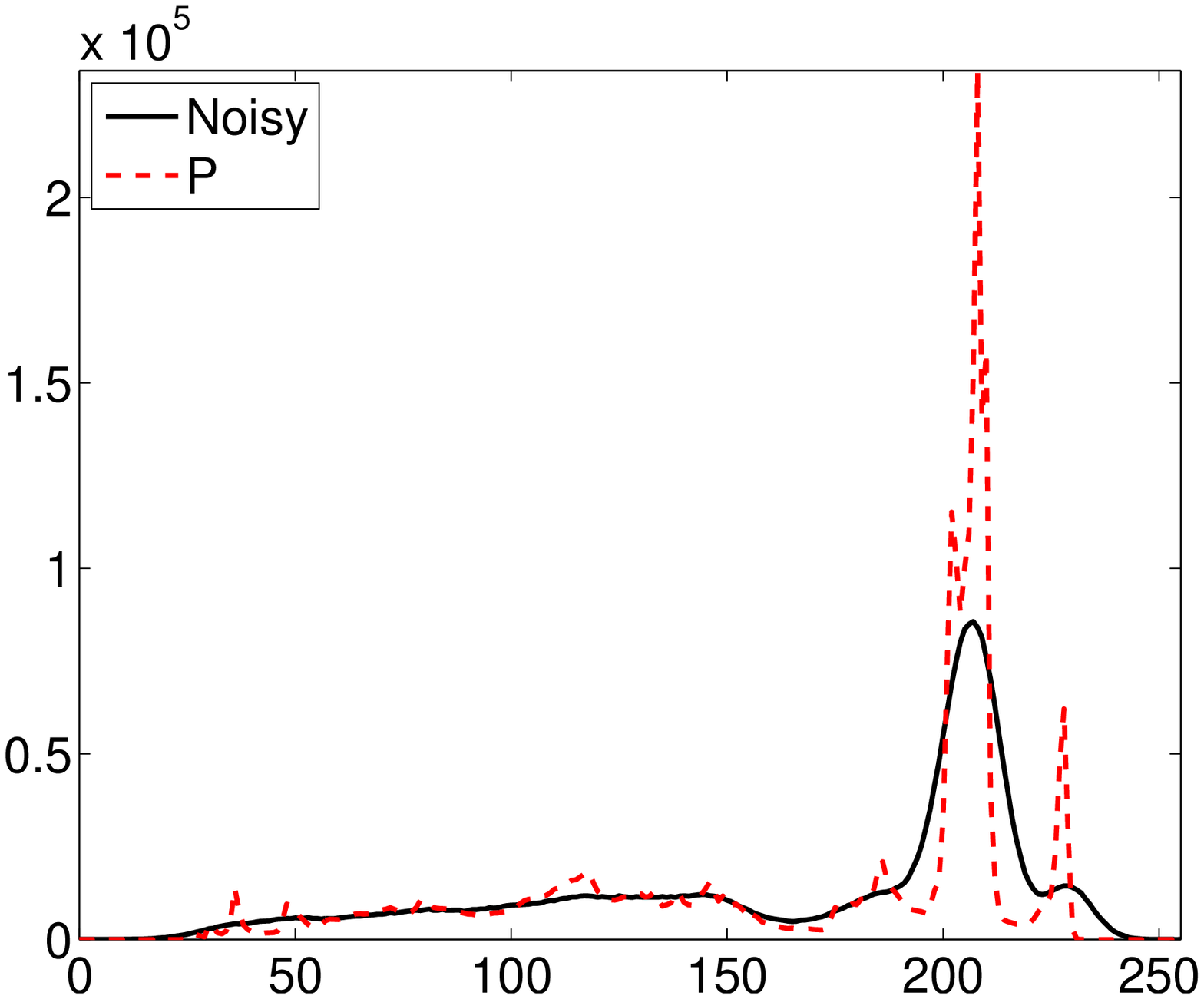}}\\
{\includegraphics[width=3cm,height=3cm]{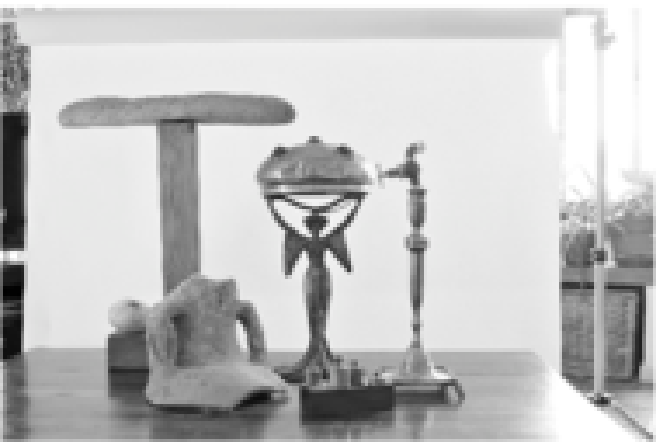}}
{\includegraphics[width=3cm,height=3cm]{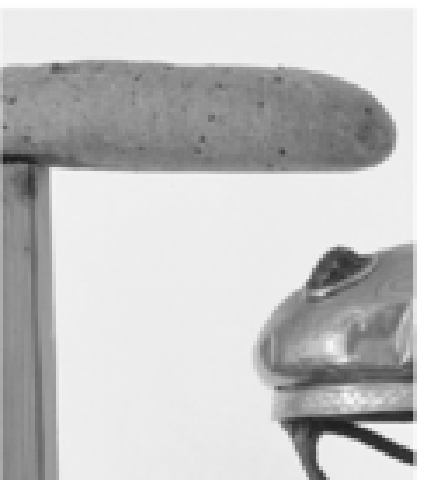}}
{\includegraphics[width=3cm,height=3cm]{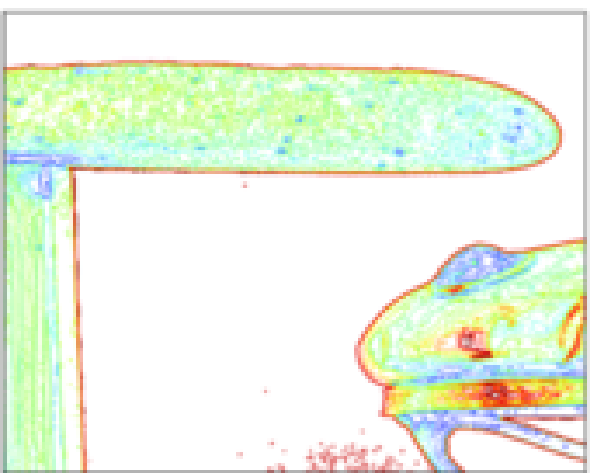}}
{\makebox[3cm]{}}\\
\caption{  Image \emph{Still life}, $h=32$. The columns are, from left to right, the 
denoised image, a detail of the image, the contour plot of the detail, and the histogram of the denoised image. The rows give, from above to below, the noisy image, 
the results of BPB, BRR20, Y8, YRR and P, and in the last row, the ground truth clean image. }
\label{fig_adam3} 
\end{figure*}

Finally, in Figures~\ref{fig_clock}, \ref{fig_boat} and \ref{fig_adam3} we show
some of the filters outcome. The columns give, from left to right, the 
denoised image, a detail of the image, the contour plot of the detail, and the histogram of the denoised image. The rows give, from above to below, the noisy image, 
the results of BPB, BRR20, Y8, YRR and P, and in the last row, the ground truth clean image.

It may be observed that, in general, the permutohedral algorithm performs a hardest 
denoising than the other algorithms. This is also reflected in the formation of 
picks in the histogram. The Y8 gives the softest denoising and performs some smoothing in the histogram of the denoised image, unlike the other algorithms. Finally, BPB and 
BRR20 are almost indistinguishable, while YRR is very close to them, although with a harder denoising effect. The staircasing effect due to the reduction of level sets is
present in all the algorithms, although not specially relevant in none of them.

\section{Summary}
In this paper we introduced the use of functional rearrangements 
to express bilateral type filters in terms of integral operators
in the one-dimensional space $[0,\abs{\O}]$.

We have proved the equivalence between the pixel-based formulation of the
original filter and its rearranged version. In addition, we proved the convergence of
discrete finite step-wise approximations of image and filter to their corresponding 
continuous limits. Although this is a property easy to obtain in the pixel-based 
formulation, it is far from trivial in its rearranged version.

In the case in which the spatial kernel, $w_\rho$, is homogeneous (e.g. the Neighborhood filter), it was proven in \cite{Galiano2014} that the level set structure of the image is left invariant 
through the filtering process, allowing to compute the filter jointly for all 
the pixels in each level set, instead of pixel-wise. 

However, if the spatial kernel is not homogeneous the invariance of the $u-$level sets 
through the filter is, in general, broken. Despite this fact, there still remains  an
important property of the rearranged version: the range kernel $\cK_h(u(\bx)-u(\by))$ is transformed into a pixel-independent kernel $\cK(q_k-q_i)$, implying 
a large gain in computational effort, as already observed for particular cases in \cite{Porikli2008}.
We have illustrated numerically this property.

The present state of the rearranging technique has an inherent restriction: the need 
of a relation of order in the range space. While this may be a minor issue for 
vector-valued color images, for which a map to some one-dimensional color space 
may be used, it is unclear how to extend the method to patch-based algorithms 
such as the Nonlocal Means. Whether or not the patch reordering techniques introduced,
among others, by Ram et al. \cite{Ram2013} may be adapted to the rearranging  approach will be  the focus of future research.

\section{Acknowledgments}

The authors are partially supported by the Spanish DGI Project MTM2013-43671.

The authors thank to the anonymous reviewers for their interesting comments 
and suggestions, that highly contributed to the improvement of our work.

%%%%%%%%%%%%%%%%%%

\newpage
\begin{table*} 
{\footnotesize
\centering
\begin{tabular}{|c|c||c|c||c|c||c|c||c|c|}
\hline
h  & \multicolumn{1}{|c||}{YPB} & \multicolumn{2}{|c||}{YRR} & \multicolumn{2}{|c||}{EYang256} & \multicolumn{2}{|c||}{EYang64}& \multicolumn{2}{|c|}{EYang8} \\
\hline 
\hline 
\multicolumn{3}{|c}{} & \multicolumn{4}{c}{\emph{Clock} ($256\times 256$)} & \multicolumn{3}{c|}{} \\
\hline 
h   & Time   & Time & PSNR & Time & PSNR &Time & PSNR &Time & PSNR \\
\hline 
  4 &   0.28  &    0.02 &   61.09 &   0.09 &  61.09&   0.08&  37.02&   0.10&  19.47 \tabularnewline \hline 
  8 &   1.02  &    0.02 &   59.93 &   0.12 &  59.93&   0.13&  42.92&   0.11&  25.01 \tabularnewline \hline 
 16 &   3.91  &    0.03 &   55.90 &   0.15 &  55.90&   0.15&  44.23&   0.14&  21.84 \tabularnewline \hline 
 32 &   15.40 &    0.06 &   49.73 &   0.10 &  49.73&   0.09&  43.37&   0.10&  21.36 \tabularnewline \hline 
\hline 
\multicolumn{3}{|c}{} & \multicolumn{4}{c}{\emph{Boat} ($512\times 512$)} & \multicolumn{3}{c|}{} \\
\hline 
h   & Time   & Time & PSNR & Time & PSNR &Time & PSNR &Time & PSNR \\
\hline 
  4 &  1.10   &    0.06 &   66.21 &   0.42 &  66.21&   0.37&  36.53&   0.38&  20.64 \tabularnewline \hline 
  8 &  4.23   &    0.08 &   61.53 &   0.44 &  61.53&   0.45&  41.23&   0.50&  21.60 \tabularnewline \hline 
 16 &  16.79  &    0.14 &   57.28 &   0.51 &  57.28&   0.40&  42.17&   0.40&  19.90 \tabularnewline \hline 
 32 &  71.15  &    0.22 &   49.74 &   0.48 &  49.74&   0.36&  43.45&   0.36&  21.82 \tabularnewline \hline 
\hline 
\multicolumn{3}{|c}{} & \multicolumn{4}{c}{\emph{Airport} ($1024\times 1024$)} & \multicolumn{3}{c|}{} \\
\hline 
h   & Time   & Time & PSNR & Time & PSNR &Time & PSNR &Time & PSNR \\
\hline 
  4 &  4.43  &    0.23 &   69.65 &   1.58 &  69.80&   1.44&  36.43&   1.61&  19.80 \tabularnewline \hline 
  8 &  17.19 &    0.30 &   64.02 &   1.80 &  64.02&   1.78&  41.58&   1.82&  22.60 \tabularnewline \hline 
 16 &  70.75 &    0.58 &   59.14 &   1.80 &  59.13&   1.70&  42.72&   1.75&  20.60 \tabularnewline \hline 
 32 &  291.4&    0.92 &   50.62 &   1.71 &  50.62&   1.44&  43.23&   1.54&  19.42 \tabularnewline \hline 
\hline 
\multicolumn{3}{|c}{} & \multicolumn{4}{c}{\emph{Still life} ($2144\times 1424$)} & \multicolumn{3}{c|}{} \\
\hline
h   & Time   & Time & PSNR & Time & PSNR &Time & PSNR &Time & PSNR \\
\hline 
4 &   14.07 &    0.68 &   64.30 &   3.82 &  64.29&   3.45&  38.08&   3.97&  19.58 \tabularnewline \hline 
  8 & 52.71 &    1.06 &   59.27 &   4.41 &  59.27&   3.95&  43.90&   4.00&  22.24 \tabularnewline \hline 
 16 & 204.0   &    1.71 &   54.59 &   5.20 &  54.59&   4.61&  45.01&   4.61&  21.28 \tabularnewline \hline 
 32 & 834.8 &    2.65 &   48.37 &   4.74 &  48.37&   4.73&  44.49&   4.62&  22.61 \tabularnewline \hline 
\end{tabular}
\caption{ Comparison between the pixel-based Yaroslavsky (YPB) filter algorithm, the rearranged version introduced in this article (YRR), and several instances of Yang's exact algorithm ($\tilde n =256,~64,~8$). For each algorithm, left column  gives
execution times (in sec.) while second column gives PSNR with respect to YPB. Image sizes enclosed in parentheses.} 
\label{table1}
}
\end{table*}

\begin{table*} 
{\footnotesize
\centering
\begin{tabular}{|c|c|c|c|c|c|c|c||c|c|c|c|c|c|}
\hline 
\multicolumn{8}{|c||}{Compared to ground truth} & \multicolumn{6}{|c|}{Compared to BPB} \\
\hline 
\hline 
\multicolumn{5}{|c}{} & \multicolumn{4}{c}{\emph{Clock} ($256\times 256$)} & \multicolumn{5}{c|}{} \\
\hline 
h  & BPB & BRRM & BRR20& Y256 & Y8 & P & YRR & BRRM & BRR20& Y256 & Y8 & P & YRR\\
\hline 
  4 &  3.62 &    3.62 &   3.62 &  3.32 & -1.24&  3.58 & 3.64 &  66.2  &  42.6&  26.9&  1.17 &  25.9&  29.7 \tabularnewline \hline 
  8 &  4.08 &    4.08 &   4.08 &  3.77 &  3.89&  4.34 & 4.03 &  62.2  &  43.1&  26.5&  21.3 &  19.2&  27.2 \tabularnewline \hline 
 16 &  4.40 &    4.40 &   4.40 &  4.04 &  4.10&  4.30 & 4.31 &  63.8  &  41.5&  23.2&  20.3 &  13.5&  21.1 \tabularnewline \hline 
 32 &  4.31 &    4.31 &   4.31 &  3.93 &  4.04&  3.31 & 3.55 &    57  &  38.4&  16.5&    16 &  7.17&  12.4 \tabularnewline \hline
\hline 
\multicolumn{5}{|c}{} & \multicolumn{4}{c}{\emph{Boat} ($512\times 512$)} & \multicolumn{5}{c|}{} \\
\hline 
h  & BPB & BRRM & BRR20& Y256 & Y8 & P & YRR & BRRM & BRR20& Y256 & Y8 & P & YRR\\
\hline 
  4 &  10.60 &   10.60 &  10.60 & 10.80 &  4.36&  10.60 & 10.60 &  69.20  &  42.70&  26.80&  5.26  &  25.80&  29.40 \tabularnewline \hline 
  8 &  11.00 &   11.00 &  11.00 & 11.00 &  10.50& 11.00 & 10.70 &  63.10  &  42.10&  26.30&  20.60 &  17.70&  26.00 \tabularnewline \hline 
 16 &  9.23 &    9.23 &   9.23 &  8.93 &   8.80&  7.46  & 8.69  &  62.90  &  40.70&  23.10&  18.30 &  12.40&  20.30 \tabularnewline \hline 
 32 &  4.83 &    4.83 &   4.83 &  4.20 &  4.22&   2.59  & 3.95  &  57.70  &  38.40&  17.20&  15.80 &   7.80&  14.20 \tabularnewline \hline 
\hline 
\multicolumn{5}{|c}{} & \multicolumn{4}{c}{\emph{Airport} ($1024\times 1024$)} & \multicolumn{5}{c|}{} \\
\hline 
h  & BPB & BRRM & BRR20& Y256 & Y8 & P & YRR & BRRM & BRR20& Y256 & Y8 & P & YRR\\
\hline 
  4 &  3.83 &    3.83 &   3.83 &    ND&  3.15&  3.81 & 3.83 &  71.00  &  42.60&    ND &  11.00 &  26.00&  29.40 \tabularnewline \hline 
  8 &  3.89 &    3.89 &   3.89 &    ND&  3.85&  3.78 & 3.87 &  67.00  &  42.20&    ND &  21.70 &  18.40&  26.70 \tabularnewline \hline 
 16 &  3.60 &    3.60 &   3.59 &    ND&  3.59&  2.94 & 3.59 &  64.30  &  41.10&    ND &  17.00 &  12.60&  21.70 \tabularnewline \hline 
 32 &  2.45 &    2.45 &   2.44 &    ND&  2.35&  1.07 & 2.38 &  58.20  &  38.70&    ND &  11.5 &    8.53&  15.50 \tabularnewline \hline 
\hline 
\multicolumn{5}{|c}{} & \multicolumn{4}{c}{\emph{Still life} ($2144\times 1424$)} & \multicolumn{5}{c|}{} \\
\hline
h  & BPB & BRRM & BRR20& Y256 & Y8 & P & YRR & BRRM & BRR20& Y256 & Y8 & P & YRR\\
\hline 
  4 &    -1 &      -1 &     -1 &    ND& -1.71& -0.99 &-0.99 &  66.60  &  43.30&    ND &  7.96 &  25.50&  30.20 \tabularnewline \hline 
  8 & -0.89 &   -0.89 &  -0.89 &    ND&  -1.3& -0.83 &-0.90 &  63.00  &  43.60&    ND &  19.10 &  19.90&  28.40 \tabularnewline \hline 
 16 & -0.92 &   -0.92 &  -0.92 &    ND& -1.39& -0.97 &-0.97 &  60.70  &  42.80&    ND &  15.40 &  16.90&    23.00 \tabularnewline \hline 
 32 & -1.31 &    -1.3 &   -1.3 &    ND& -1.94& -1.47 &-1.48 &  54.70  &  40.50&    ND &  10.50 &  12.30&    17.00 \tabularnewline \hline 

\end{tabular}
\caption{ Left box: PSNR between the ground truth image and the pixel-based Bilateral filter (BPB), its rearranged version for the maximum number of meaningful levels (BRRM), 
the rearranged version for 20 levels (BRR20), two instances of Yang's algorithm, (Y256 and Y8), the Permutohedral algorithm (P), and the Yaroslavsky filter in its rearranged version (YRR). 
Right box: PSNR between the pixel-based Bilateral filter (BPB) and the other algorithms. ``ND'' means ``no data available`` due to 
too huge Y256 memory requirements  for these images. Image sizes enclosed in parentheses.} 
\label{table2}
}
\end{table*}

\begin{table*} 
{\footnotesize
\centering
\begin{tabular}{|c|c|c|c|c|c|c|c||c|c|c|c|c|c|}
\hline 
\multicolumn{8}{|c||}{Execution time} & \multicolumn{6}{|c|}{Execution time relative to BPB} \\
\hline 
\hline
\multicolumn{5}{|c}{} & \multicolumn{4}{c}{\emph{Clock} ($256\times 256$)} & \multicolumn{5}{c|}{} \\
\hline 
h   & BPB    & BRRM    & BRR20  & Y256  & Y8    & P     & YRR   &    BRRM &  BRR20 & Y256  & Y8     & P      & YRR \\
\hline
  4 &   0.56 &    0.23 &   0.17 &  0.32 &  0.02 &  0.46 &  0.02 &    2.43 &   3.29 &  1.75 &     28 &  1.217 &    28 \tabularnewline \hline 
  8 &   2.06 &    0.52 &   0.25 &   0.3 &  0.02 &  0.19 &  0.02 &    3.96 &   8.24 &  6.87 &    103 &  10.84 &   103 \tabularnewline \hline 
 16 &   7.98 &    2.56 &   0.62 &   0.3 &  0.02 &  0.05 &  0.03 &    3.12 &   12.9 &  26.6 &    399 &  159.6 &   266 \tabularnewline \hline 
 32 &  31.41 &    11.4 &    1.8 &  0.32 &  0.02 &  0.03 &  0.06 &    2.76 &   17.4 &  98.2 &   1570 &   1047 &   524 \tabularnewline \hline 
\hline 
\multicolumn{5}{|c}{} & \multicolumn{4}{c}{\emph{Boat} ($512\times 512$)} & \multicolumn{5}{c|}{} \\
\hline 
h   & BPB    & BRRM    & BRR20  & Y256  & Y8    & P     & YRR   &    BRRM &  BRR20 & Y256  & Y8     & P      & YRR \\
\hline
  4 &   2.19 &     0.9 &   0.68 &  1.36 &   0.1 &  2.11 &  0.06 &    2.43 &   3.22 &  1.61 &   21.9 &  1.038 &  36.5 \tabularnewline \hline 
  8 &   8.33 &    2.17 &   1.12 &  1.24 &  0.06 &  1.04 &  0.08 &    3.84 &   7.44 &  6.72 &  138.8 &   8.01 &   104 \tabularnewline \hline 
 16 &  32.64 &    10.1 &   2.44 &   1.2 &  0.06 &  0.32 &  0.14 &    3.23 &   13.4 &  27.2 &    544 &    102 &   233 \tabularnewline \hline 
 32 &  132.1 &    47.2 &   7.27 &  1.16 &  0.06 &  0.14 &  0.22 &    2.79 &   18.2 &   114 &   2201 &  943.2 &   600 \tabularnewline \hline 
\hline 
\multicolumn{5}{|c}{} & \multicolumn{4}{c}{\emph{Airport} ($1024\times 1024$)} & \multicolumn{5}{c|}{} \\
\hline
h   & BPB    & BRRM    & BRR20  & Y256  & Y8    & P     & YRR   &    BRRM &  BRR20 & Y256  & Y8     & P      & YRR \\
\hline
  4 &   8.76 &    3.56 &   2.78 &     ND &  0.24 &  9.88 &  0.23 &    2.46 &   3.15 &   ND &   36.5 & 0.8866 &  38.1 \tabularnewline \hline 
  8 &  33.37 &    8.45 &   4.29 &     ND &  0.22 &  5.08 &   0.3 &    3.95 &   7.78 &   ND &  151.7 &  6.569 &   111 \tabularnewline \hline 
 16 &  132.8 &    39.6 &   9.36 &     ND &  0.22 &  1.37 &  0.58 &    3.35 &   14.2 &   ND &  603.8 &  96.96 &   229 \tabularnewline \hline 
 32 &  529.5 &     185 &   28.1 &     ND &   0.2 &  0.58 &  0.92 &    2.86 &   18.8 &   ND &   2648 &  912.9 &   576 \tabularnewline \hline 
\hline 
\multicolumn{5}{|c}{} & \multicolumn{4}{c}{\emph{Still life} ($2144\times 1424$)} & \multicolumn{5}{c|}{} \\
\hline
h   & BPB    & BRRM    & BRR20  & Y256  & Y8    & P     & YRR   &    BRRM &  BRR20 & Y256  & Y8     & P      & YRR \\
\hline
  4 &   26.5 &    10.3 &   8.02 &     ND &  0.64 &  31.4 &  0.68 &    2.58 &    3.3 &   ND &  41.41 & 0.8429 &    39 \tabularnewline \hline 
  8 &  99.07 &    24.7 &   12.2 &     ND &  0.62 &  9.35 &  1.06 &    4.01 &   8.09 &   ND &  159.8 &   10.6 &  93.5 \tabularnewline \hline 
 16 &  384.8 &     111 &   26.8 &     ND &  0.66 &  2.58 &  1.71 &    3.47 &   14.4 &   ND &    583 &  149.1 &   225 \tabularnewline \hline 
 32 &   1536 &     511 &   80.6 &     ND &  0.62 &  1.52 &  2.65 &       3 &   19.1 &   ND &   2478 &   1011 &   580 \tabularnewline \hline 

\end{tabular}
\caption{ Execution times for the images used in the experiments (sizes enclosed in parentheses). Left box: Execution times ot the pixel-based Bilateral filter (BPB), its rearranged version for the maximum number of meaningful levels (BRRM), 
the rearranged version for 20 levels (BRR20), two instances of Yang's algorithm, (Y256 and Y8), the Permutohedral algorithm (P), and the Yaroslavsky filter in its rearranged version (YRR). 
Right box: Execution times relative to BPB, that is, number of times the other algorithms are faster than the BPB algorithm. ``ND'' means ``no data available`` due to 
too huge Y256 memory requirements  for these images.} 
\label{table3}
}
\end{table*}

\newpage

\end{document}